\documentclass[journal]{IEEEtran}
\usepackage{amsmath,amsfonts}
\usepackage{algorithmic}
\usepackage{algorithm}
\usepackage{array}
\usepackage[caption=false,font=normalsize,labelfont=sf,textfont=sf]{subfig}
\usepackage{textcomp}
\usepackage{stfloats}
\usepackage{url}
\usepackage{verbatim}
\usepackage{graphicx}
\usepackage{comment}
\usepackage{adjustbox}
\usepackage{multicol}
\usepackage{multirow}
\usepackage{makecell}
\usepackage{booktabs}
\usepackage{float}
\usepackage{nicematrix}
\usepackage{hyperref}
\usepackage[switch]{lineno}
\usepackage{cleveref}
\Crefname{figure}{Fig.}{Figs.}
\Crefname{table}{Tab.}{Tabs.}
\Crefname{equation}{Eq.}{Eqs.}
\Crefname{section}{Sec.}{Secs.}

\usepackage{array}

\usepackage[sorting=none]{biblatex}
\usepackage{xcolor}
\newcommand{\quotes}[1]{``#1''}

\begin{document}

\newcommand{\papertitle}{Illuminant-Adaptive 3D Lookup Tables for Camera Color Correction}

\title{\papertitle}

\author{Claudio Rota, Luca Cogo, Simone Bianco, Raimondo Schettini\\
Department of Informatics, Systems and Communication\\
University of Milano-Bicocca
\\Milan, Italy\\
\{name.surname\}@unimib.it\\

}

\maketitle

\begin{abstract}

Color correction is a key component of camera image signal processing (ISP) pipelines, encompassing illuminant discounting and colorimetric mapping of device-dependent sensor responses to device-independent color spaces, such as CIE~XYZ. Despite extensive research, accurate color correction remains challenging due to the non-linear relationship between camera sensor responses and CIE~XYZ color space, as well as to the increasing presence of highly chromatic and spectrally complex LED illuminants.
We propose a color correction framework based on illuminant-adaptive three-dimensional lookup tables (LUTs), which we call Color~Correction~LUT~(C$^2$LUT). Our method combines a chromaticity-aware illuminant representation with a non-linear color transformation, enabling accurate correction under illuminants spanning a wide range of chromaticities and spectral complexities. We employ Tucker tensor decomposition to represent the LUTs, ensuring that computational requirements remain sufficiently low for deployment in camera ISPs. In addition, we introduce a large-scale illuminants dataset comprising 1,473 spectral power distributions, with different chromaticities and spectral profiles.
Experiments across multiple cameras, illuminants, reflectance datasets, and real captured images demonstrate consistent improvements over existing methods for color correction, reducing CIE~$\boldsymbol{\Delta E_{00}}$ by up to 20\% and angular error by up to 18\% while remaining compatible with modern camera hardware constraints. Code and datasets are available at \url{https://github.com/claudiom4sir/C2LUT}.
\end{abstract}


\section{Introduction}
Modern digital cameras employ a sequence of in-camera image processing operations to transform raw sensor measurements into visually accurate output images. Within this image signal processing (ISP) pipeline, the colorimetric mapping stage plays a fundamental role by compensating for scene illumination and converting device-dependent sensor RGB responses into a device-independent color space, such as CIE~XYZ~\cite{CIE19}. The accuracy of this transformation strongly influences the overall color fidelity and visual consistency of captured images under varying illumination conditions~\cite{fairchild2013color}.

Conventional in-camera color correction pipelines were originally designed under the assumption that captured scenes were illuminated by standard light sources, such as daylight and incandescent lamps~\cite{green2023fundamentals}. The spectral characteristics of these illuminants can be reasonably approximated using blackbody radiation models, whose spectral  distributions (SPDs) are determined by temperature according to the Planck’s law~\cite{planck1901law}. In the CIE~xy-chromaticity diagram, the chromaticities of ideal blackbody radiators at increasing temperatures define the Planckian locus. 
Under this assumption, scene illumination can be compactly represented using correlated color temperature (CCT), which provides a one-dimensional descriptor of the illuminant chromaticity. Consequently, many commercial camera pipelines estimate the scene CCT and derive the corresponding color space transformation (CST) through interpolation between pre-calibrated color correction matrices (CCMs)~\cite{dng}.

\begin{figure}
    \centering
    \includegraphics[width=\columnwidth]{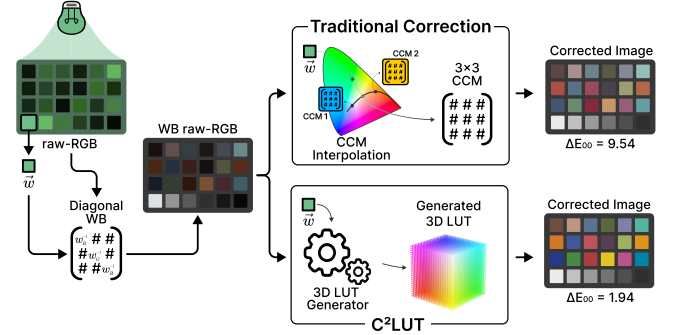}
    \caption{Overview of the proposed color correction framework (C$^2$LUT). Traditional color correction methods are based on a CCM obtained via interpolation between two reference CCMs calibrated under D65 and Illuminant A, weighted by the estimated CCT of the scene. Our framework instead employs a LUT generator that produces an illuminant-adaptive 3D~LUT for color correction, leading to more colorimetrically accurate results.
    }
    \label{fig:graphical_abstract}
\end{figure}

The use of interpolated CCMs based on illuminant CCT presents two important limitations. First, since CCMs perform linear transformations, their color correction accuracy relies on the Luther condition~\cite{von1927gebiet}, which requires the camera spectral sensitivities to be a linear transformation of the CIE~XYZ color matching functions (CMFs). Since real camera sensors generally do not satisfy this condition~\cite{finlayson2020designing}, a CCM-based correction cannot reproduce accurate color responses under arbitrary combinations of illuminants and surface reflectances. Second, modern illumination environments increasingly include LED light sources characterized by highly non-smooth and narrow-band SPDs that cannot be approximated by Planck's law. As a result, their chromaticity coordinates may substantially deviate from the Planckian locus, making CCT an insufficient descriptor of the illuminant. In such cases, interpolating CCMs based solely on illuminant CCT may lead to inaccurate color transformations~\cite{tedla2025off}.

\begin{figure*}
    \centering
    \includegraphics[width=1\linewidth]{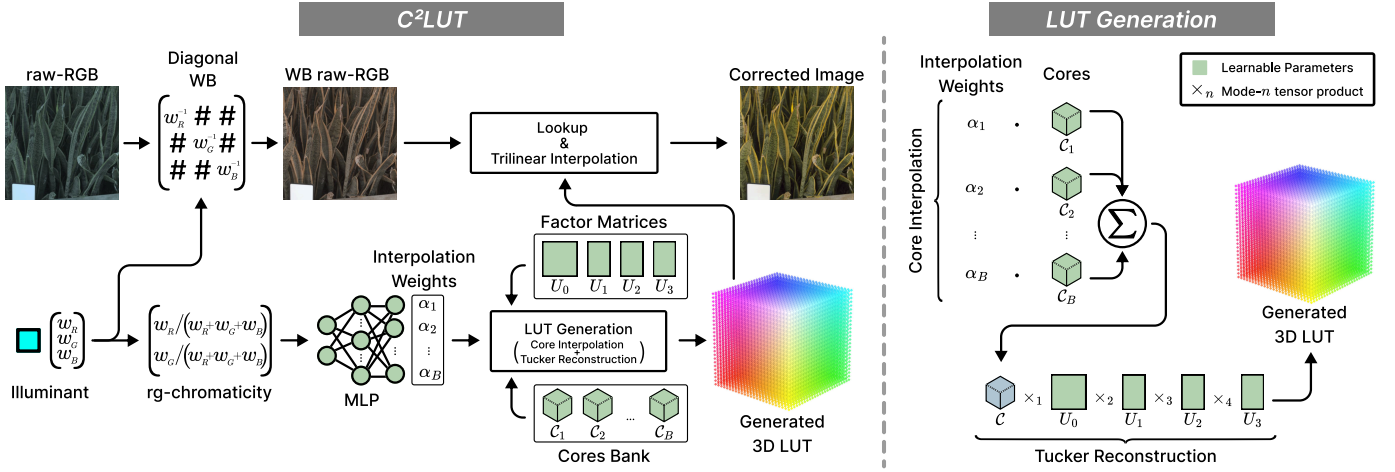}
    \caption{\textit{Left}: Overview of C$^2$LUT. 
    The raw-RGB image is first white-balanced using the illuminant white point, and then color-corrected using an illuminant-dependent 3D~LUT. Using the illuminant rg-chromaticity, an MLP predicts the coefficients for linearly interpolating a set of basis cores, which are then assembled via Tucker tensor reconstruction into the final 3D~LUT for color correction. \textit{Right}: Detailed illustration of the basis core interpolation and Tucker reconstruction process. The coefficients $\{\alpha_b\}^B_{b=1}$ are employed to linearly combine a set of basis cores $\{\mathcal{C}_b\}^B_{b=1}$, producing an interpolated core $\mathcal{C}$. It is subsequently refined and upsampled through the factor matrices $U_0, U_1, U_2, U_3$ to produce the final 3D~LUT.
    \vspace{-4mm}
    }
    \label{fig:proposed_method}
\end{figure*}

In the literature, several works have been proposed to tackle these problems. On one hand, to overcome the limitations of CCM-based linear transformations, several works have explored non-linear color correction strategies, including polynomial regression models~\cite{hong2001study, finlayson2015color} and neural networks~\cite{macdonald2021camera, kucuk2022exposure}, which provide more expressive mappings between sensor responses and target color spaces. However, these methods still use CCT as illuminant representation, limiting their robustness under modern non-Planckian lighting conditions. On the other hand, to overcome the limitations of CCT as illuminant descriptor, some works~\cite{tedla2025off} proposed replacing the one-dimensional CCT representation with a two-dimensional chromaticity-based formulation for CCM estimation: instead of representing illumination solely through CCT, they model illuminants directly in the CIE~xy-chromaticity space and predict a chromaticity-dependent CCM. However, this solution still relies on a CCM to perform color correction and cannot model the non-linear relationship between camera sensor responses and the CIE~XYZ CMFs~\cite{finlayson2020designing}.

In this work, we jointly address both limitations by combining a chromaticity-aware illuminant representation with a non-linear color correction framework.
While non-linear corrections can in principle be implemented via neural networks or polynomial mappings, 3D lookup tables (LUTs) offer a compelling trade-off between expressiveness and computational efficiency, making them particularly suitable for deployment in camera ISPs~\cite{selan2005using}. The proposed framework, which we call Color~Correction~LUT~(C$^2$LUT), is summarized in \Cref{fig:graphical_abstract}. It captures the non-linear relationship between camera raw responses and reference XYZ values while avoiding the limitations of CCT-based illuminant modeling under spectrally complex and non-Planckian light sources.
Starting from public datasets~\cite{barnard2002data, brendel_2020_4051012, lspdd_database} containing SPD measurements from different light sources, including standard, fluorescent and LED illuminants, we construct a new dataset comprising 1,473 illuminants with different spectral properties and covering a broad region of the CIE~xy-chromaticity gamut. We combine these illuminants with spectral reflectance datasets~\cite{li2021multispectral, zhang2016metamer, mccamy1976color} and multiple real-camera spectral sensitivity functions (SSFs)~\cite{imageengineering} to generate physically accurate RAW-XYZ correspondences under diverse lighting conditions. 
Using these data, we optimize C$^2$LUT to produce illuminant-dependent LUTs. As shown in \Cref{fig:proposed_method}, a set of basis LUTs is interpolated using weights predicted by a lightweight multi-layer perceptron (MLP), which depend on the illuminant chromaticity. Combining the basis LUTs according to these weights yields a color correction LUT adapted to the specific illuminant, allowing the method to generalize to arbitrary lighting conditions. We parameterize the LUT representation of C$^2$LUT using Tucker tensor decomposition~\cite{tucker1966some}, which considerably reduces the number of optimizable parameters while preserving representational capacity. Experimental results 
show that C$^2$LUT consistently achieves lower CIE~$\Delta E_{00}$~\cite{sharma2005ciede2000} and angular error compared to existing approaches for in-camera color correction, demonstrating superior color reproduction accuracy under varying illumination conditions while remaining compatible with the computational constraints of modern camera pipelines.

The main contributions of this work can be summarized as follows:
\begin{itemize}

\item We propose C$^2$LUT, an illuminant-adaptive 3D~LUT framework for in-camera color correction, in which multiple basis LUTs, represented via Tucker tensor decomposition, are interpolated using weights predicted by a lightweight MLP from the scene illuminant chromaticity.

\item We construct a large-scale illuminants dataset comprising 1,473 SPDs with different spectral properties covering a broad region of the CIE~xy-chromaticity gamut.

\item We conduct extensive experiments on three camera devices, three reflectance datasets, and 295 test illuminants, demonstrating consistent improvements over representative in-camera color correction baselines.

\end{itemize}

\begin{figure}[t]
    \centering
    \begin{adjustbox}{width=\columnwidth}
    \setlength{\tabcolsep}{-10pt}
    \begin{tabular}{cc}
    \includegraphics[width=\columnwidth]{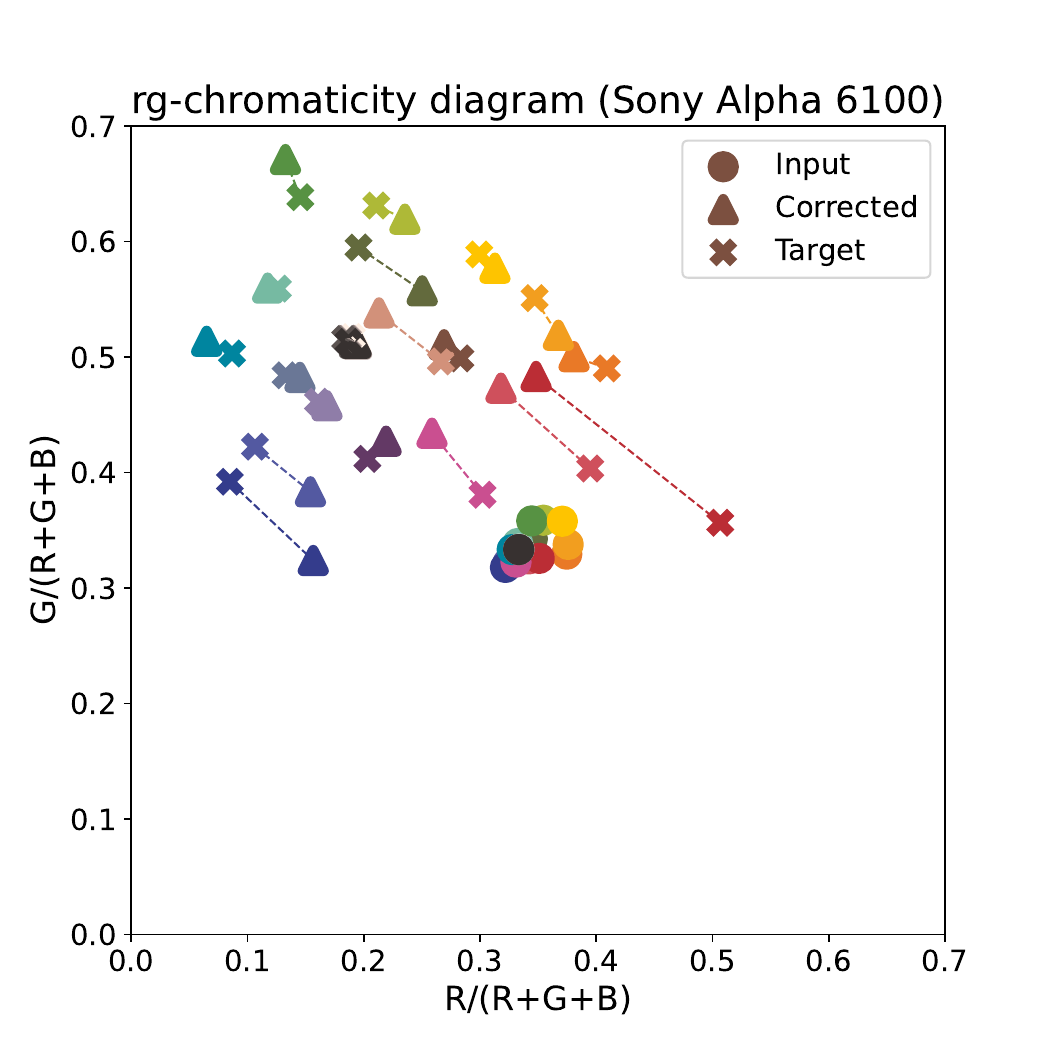} & \includegraphics[width=\columnwidth]{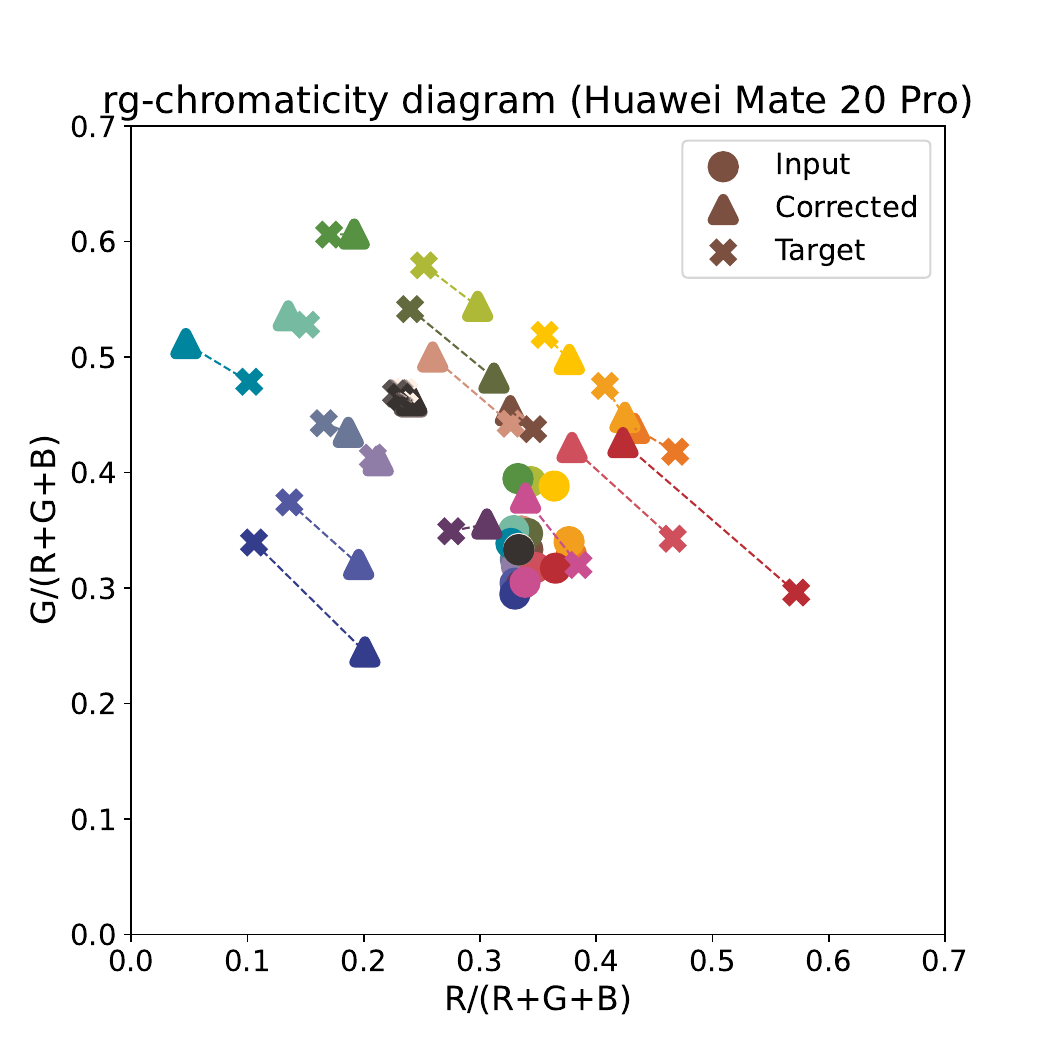} 
    \end{tabular}
    \end{adjustbox}
    \caption{Input: rg-chromaticity coordinates of the Macbeth Color Checker (MCC) patches under an LED illuminant with a blue-shifted chromaticity. 
    Corrected: rg-chromaticity coordinates obtained after applying the optimal CCM computed via least-squares regression. Target: expected rg-chromaticity coordinates. The corrected coordinates remain substantially displaced from the targets, indicating that a linear model such as the CCM cannot fully compensate for this illuminant. The non-overlapping distributions of triangles and crosses confirm that the correct mapping is inherently non-linear.
    }
    \label{fig:motivation_nonlinear}
\end{figure}

\section{Motivation and related works}

The in-camera color correction pipeline aims to transform the sensor raw image into a device-independent color space under a canonical illuminant. In modern imaging pipelines, this process is typically decomposed into three stages: illuminant estimation, illuminant discounting, and color space transform (CST). 

The first stage estimates the color of the scene illuminant, which is responsible for the color cast observed in the raw image. A substantial body of research has investigated this problem over the years. Early methods relied on statistical assumptions about scene reflectance distributions and image statistics~\cite{van2007edge}, while more recent approaches leverage deep neural networks to infer the illuminant based on the image semantics~\cite{bianco2015color, hu2017fc4}.

Once the illuminant has been estimated, illuminant discounting is commonly performed through a diagonal 3$\times$3 scaling matrix, following the von Kries adaptation model~\cite{kries1905dic}. This operation applies a channel-wise normalization intended to remove the global color cast introduced by the scene illumination. However, this step only guarantees the correct mapping of neutral colors. 
A subsequent CST stage further refines the correction and maps colors into a target color space (e.g., CIE~XYZ~\cite{CIE19}) under a reference illuminant (e.g., D65). In practice, camera pipelines usually implement this step through a linear 3$\times$3 color correction matrix (CCM) whose coefficients are determined according to the estimated illuminant CCT, interpolating between two pre-computed calibration matrices associated with tungsten ($\sim$2500K) and daylight ($\sim$6500K) illuminants~\cite{dng}.

Despite its simplicity and efficiency, this conventional pipeline presents several important limitations. First, the interpolation strategy parameterizes illuminants solely through their CCT, effectively projecting the illuminant chromaticity onto a one-dimensional representation along the Planckian locus. While this approximation is reasonable for blackbody-like illuminants, it becomes inaccurate for many modern artificial light sources, particularly LEDs and mixed illuminants, whose chromaticities can lie far away from the Planckian locus. Second, the entire pipeline relies on linear operations at every stage. This assumption neglects the intrinsic non-linearities of the image formation and color correction process. In particular, the mapping from raw sensor responses to perceptually accurate tristimulus values is inherently non-linear because real camera sensors typically violate the Luther condition~\cite{von1927gebiet}. Furthermore, as shown in \Cref{fig:motivation_nonlinear}, relighting operations that transform colors from the scene illuminant to a canonical illuminant are themselves non-linear in camera space. Consequently, purely linear pipelines often produce residual color errors, especially under challenging illuminants or for highly saturated colors, highlighting the need for more expressive correction models.

Several works have attempted to address these limitations. A first line of research focuses on improving the estimation of the CCMs. Karaimer and Brown~\cite{karaimer2018improving} showed that increasing the number of pre-calibrated matrices used during interpolation improves color accuracy compared to the standard two-matrix formulation. Similarly, Tedla~et~al.~\cite{tedla2025off} replaced the conventional CCT-based interpolation with a learned mapping conditioned directly on the illuminant CIE xy-chromaticity, enabling a more flexible adaptation to illuminants that deviate from the Planckian locus. However, these approaches still rely on global linear transformations and therefore remain fundamentally limited in their ability to model non-linear color mappings.
Another line of work instead focuses on overcoming the limitations of linear transformations. Polynomial and root-polynomial methods have been proposed to better approximate the non-linear mapping from camera raw space to reference color spaces~\cite{hong2001study,finlayson2015color}, while more recent neural-network-based approaches~\cite{macdonald2021camera, kucuk2022exposure} further increase the expressive power of the transformation. However, many of these methods either ignore the dependency of the correction on the scene illuminant or still rely on illuminant representations based solely on CCT. Additionally, the dense pixel-wise nature of these approaches often results in spatially inconsistent corrections~\cite{tedla2025off} and limited scalability to high-resolution images.
In this work, we propose a framework that jointly addresses these limitations by combining a non-linear color transformation with an illuminant representation based directly on illuminant chromaticity.

\section{Methodology}

\subsection{Problem formulation}

The response of a digital camera sensor in channel $k \in \{R, G, B\}$ 
to a scene surface with spectral reflectance $R(\lambda)$ under an 
illuminant with SPD $E(\lambda)$ is modeled as:
\begin{equation}
    \rho_k = \int_\lambda E(\lambda)\, R(\lambda)\, S_k(\lambda)\, d\lambda, \quad k \in \{R, G, B\},
    \label{eq:image_formation}
\end{equation}
where $S_k(\lambda)$ denotes the spectral sensitivity of the $k$-th camera 
channel~\cite{fairchild2013color}. The goal of color correction is to map the raw camera responses $\rho = [\rho_R, \rho_G, \rho_B]^\top$ to a device-independent color space (e.g., CIE~XYZ~\cite{CIE19}) under a canonical reference illuminant (e.g., D65). Specifically, given a surface reflectance $R(\lambda)$, the target XYZ values $t = [t_X, t_Y, t_Z]^\top$ under the reference illuminant $E_{\text{ref}}(\lambda)$ are defined as:
\begin{equation}
    t_k = \int_\lambda E_{\text{ref}}(\lambda)\, R(\lambda)\, C_{k}(\lambda)\, d\lambda,
    \quad k \in \{X, Y, Z\},
    \label{eq:xyz_target}
\end{equation}
where $C_{k}(\lambda)$ denotes the CIE~XYZ CMFs. The color correction problem thus consists of finding the mapping $f: \rho \mapsto t$ that converts the camera raw image captured under the scene illuminant into device-independent XYZ tristimulus values as seen under the canonical reference illuminant D65.
The methods considered in this work assume a spatially uniform scene illuminant and therefore estimate a single global mapping $f$ that is applied to the entire image.

\begin{figure}[t]
    \centering
    \begin{adjustbox}{width=\columnwidth}
    \setlength{\tabcolsep}{-1pt}
    \begin{tabular}{cc}
    \includegraphics[width=0.99\columnwidth]{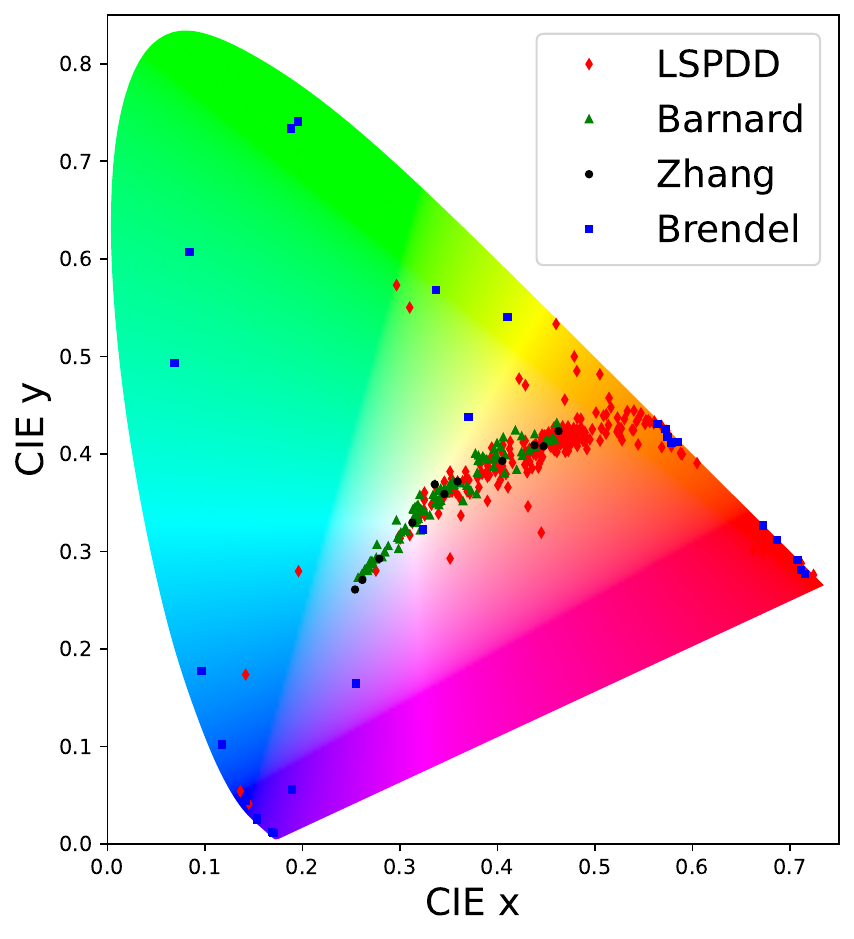} & \includegraphics[width=0.99\columnwidth]{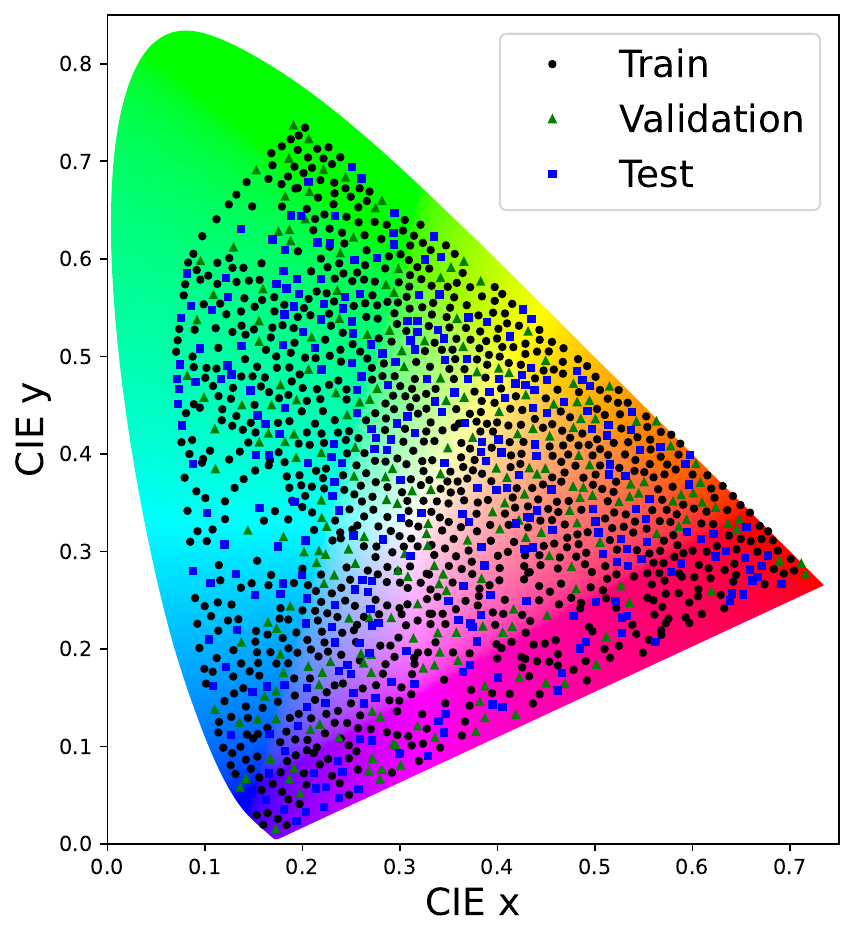} 
    \end{tabular}
    \end{adjustbox}
    \caption{CIE~xy-chromaticity distribution of the illuminants used in this work. \textit{Left}: measured SPDs collected from the source datasets. \textit{Right}: final set of 1,473 illuminants obtained after SPD interpolation and chromaticity-space sampling, shown together with the train/validation/test split. The dataset provides broad coverage of conventional and highly chromatic illuminants.
    }
    \label{fig:dataset_chromaticities}
\end{figure}

\subsection{Illuminants dataset}
\label{sec:dataset}

We collect SPDs from multiple publicly available sources, covering a wide range of lighting technologies and chromaticities. Specifically, we include the Brendel dataset~\cite{brendel_2020_4051012}, 
which contains highly chromatic LED illuminants spanning a broad region of 
the CIE~xy-chromaticity space; the dataset of Zhang et al.~\cite{zhang2016metamer}, comprising 
standard, fluorescent, and LED illuminants; the LPSDD dataset, which provides 
a diverse collection of real-world light sources~\cite{lspdd_database}; and the Barnard et al. 
dataset~\cite{barnard2002data}, which provides a set of measured natural and fluorescent illuminant SPDs captured in both indoor and outdoor scenarios. 
By aggregating these complementary sources, we obtain 438 illuminants, covering both conventional and modern illumination conditions, including Planckian, fluorescent, and highly chromatic and spectrally complex LED illuminants. The chromaticity of these illuminants in the CIE~xy color space is shown in \Cref{fig:dataset_chromaticities} (\textit{left}).

We synthesize additional SPDs through convex linear interpolation between pairs of measured illuminants to increase the density of illuminant samples and improve the uniformity of chromaticity space coverage. Given two SPDs $A(\lambda)$ and $B(\lambda)$, intermediate illuminant SPDs $C(\lambda)$ are generated as $C(\lambda) = \alpha A(\lambda) + (1 - \alpha)B(\lambda)$, with $\alpha \in \{0.1, 0.2, \ldots, 0.9\}$, producing over 700,000 physically plausible illuminants~\cite{punnappurath2025improved}. We project the resulting SPDs into the CIE~xy-chromaticity space, and select a representative subset to ensure uniform spatial coverage: illuminants are retained only if their Euclidean distance in the CIE~xy-chromaticity space to all previously selected samples exceeds 0.01. This filtering reduces redundancy while maintaining broad coverage of the CIE xy gamut. The final dataset comprises 1,473 illuminants spanning both Planckian and highly non-Planckian lighting conditions, as illustrated in \Cref{fig:dataset_chromaticities} (\textit{right}). We randomly partition the illuminants into training, validation, and test sets by allocating 80\% to training (942 illuminants) and 20\% to testing (295 illuminants), with 20\% of the training set further reserved for validation (236 illuminants).

\subsection{RAW-XYZ pair generation}
\label{sec:data_generation}

We generate RAW-XYZ pairs by combining the illuminants 
dataset described in \Cref{sec:dataset} with a set of surface reflectance spectra and camera SSFs. For each combination of illuminant $E(\lambda)$, reflectance $R(\lambda)$, and camera spectral sensitivities $S_{\{R, G, B\}}(\lambda)$, we compute the raw camera response $\rho$ according to \Cref{eq:image_formation}, and the corresponding target XYZ values $t$ under the D65 reference illuminant according to \Cref{eq:xyz_target}. The illuminant 
white point $w$ is computed as the camera response to a perfect 
white reflector $R(\lambda) = \vec{1}$ under the scene illuminant:
\begin{equation}
    w_k = \int_\lambda E(\lambda)\, S_k(\lambda)\, d\lambda, \quad k \in \{R, G, B\}.
    \label{eq:white_point}
\end{equation}
Each sample therefore consists of a raw response $\rho$, the corresponding illuminant white point $w$, 
and the target XYZ tristimulus values $t$.

\subsection{Proposed method for color correction}
\label{sec:method}

Given the raw camera sensor response $\rho$ and the white point $w = [w_R, w_G, w_B]^\top$ of the scene illuminant, the goal of the proposed framework is to map $\rho$ to the output value $t$ in the CIE~XYZ~\cite{CIE19} color space as seen under the D65 reference illuminant. An overview is illustrated in \Cref{fig:proposed_method}.

As a first step, we white-balance the raw signal using the illuminant white point to determine a diagonal $3 \times 3$ matrix:
\begin{equation}
    \tilde{\rho} = \begin{bmatrix}
w_R^{-1} & 0 & 0 \\
0 & w_G^{-1} & 0 \\
0 & 0 & w_B^{-1}
\end{bmatrix} \rho
\end{equation}

producing a white-balanced pixel $\tilde{\rho}=[\tilde{\rho}_R, \tilde{\rho}_G, \tilde{\rho}_B]^\top$ with the 
global illuminant color cast removed.

We implement the color correction mapping via a 3D~LUT~$\in \mathbb{R}^{3 \times N \times N \times N}$, where $N$ is the LUT resolution. Using Tucker tensor decomposition~\cite{tucker1966some,hitchcock1927expression}, we represent the LUT by a set of basis cores $\{\mathcal{C}_b\}_{b=1}^{B} \in \mathbb{R}^{3 \times R \times R \times R}$, where $R$ is the Tucker rank, and four shared learned factor matrices $U_0 \in \mathbb{R}^{3 \times 3}$ and $U_1, U_2,U_3 \in \mathbb{R}^{N \times R}$.

We represent the scene illuminant by its rg-chromaticity, computed from 
the white point $w$ as:
\begin{equation}
    (r, g) = \left( \frac{w_R}{w_R + w_G + w_B},\ 
    \frac{w_G}{w_R + w_G + w_B} \right),
    \label{eq:chromaticity}
\end{equation}
producing a compact 2D descriptor that captures the chromatic properties 
of the illuminant.
We then feed the chromaticity $(r, g)$ into a lightweight MLP that 
predicts a set of interpolation weights 
$\vec{\alpha} = [\alpha_1, \ldots, \alpha_B]^\top$, with 
$\alpha_b \geq 0$ and $\sum_{b=1}^{B} \alpha_b = 1$, which are used to combine the $B$ basis cores 
into a single illuminant-adaptive core:
\begin{equation}
    \mathcal{C} = \sum_{b=1}^{B} \alpha_b  \mathcal{C}_b.
    \label{eq:core_combination}
\end{equation}

We then upsample the interpolated core $\mathcal{C}$ to a full-resolution 
LUT $\mathcal{T}$ of shape $3 \times N \times N \times N$ via Tucker  tensor reconstruction using the shared factor matrices:
\begin{equation}
    \mathcal{T} = \mathcal{C} \times_1 U_0 \times_2 U_1 
    \times_3 U_2 \times_4 U_3,
    \label{eq:tucker}
\end{equation}
where $\times_n$ denotes the mode-$n$ tensor product. The spatial factor 
matrices $U_1, U_2, U_3$ upsample the core $C$
from $R\times R \times R$ to $N \times N \times N$, while the color factor matrix $U_0$ allows 
learned mixing across color channels.

Once we obtain the adaptive LUT $\mathcal{T}$, we apply it to $\tilde{\rho}$ via trilinear 
interpolation. Given a white-balanced pixel 
$\tilde{p}$ with 
values in $[0, 1]$, we compute the integer grid indices of the enclosing 
cell as:
\begin{equation}
    i = \lfloor \tilde{p}_R (N-1) \rfloor, \quad 
    j = \lfloor \tilde{p}_G (N-1) \rfloor, \quad 
    k = \lfloor \tilde{p}_B (N-1) \rfloor,
    \label{eq:grid_indices}
\end{equation}
and the fractional offsets within the cell as:
\begin{equation}
    \delta_r = \tilde{p}_R  (N-1) - i, \, 
    \delta_g = \tilde{p}_G (N-1) - j, \,
    \delta_b = \tilde{p}_B (N-1) - k.
    \label{eq:fractional_offsets}
\end{equation}
We then compute the output tristimulus value $\hat{t}$ as a weighted sum over the 
8 vertices of the enclosing cell:
\begin{equation}
    \hat{t} = \sum_{u=0}^{1} \sum_{v=0}^{1} \sum_{z=0}^{1} 
    w_{uvz}\, \mathcal{T}[i{+}u,\, j{+}v,\, k{+}z],
    \label{eq:trilinear}
\end{equation}
where the interpolation weights are:
\begin{equation}
    w_{uvz} = \delta_r^u (1-\delta_r)^{1-u}\, \delta_g^v (1-\delta_g)^{1-v}\, 
    \delta_b^z (1-\delta_b)^{1-z}.
    \label{eq:trilinear_weights}
\end{equation}
The resulting output $\hat{t} = [\hat{t}_X, \hat{t}_Y, \hat{t}_Z]^\top$ is the estimated XYZ
tristimulus value of the pixel under the D65 reference illuminant in CIE~XYZ.

We jointly optimize the MLP parameters, the basis cores $\{C_b\}_{b=1}^{B}$, and the factor matrices $U_0, U_1, U_2, U_3$ end-to-end via gradient descent by minimizing a combination of two loss terms. The first is the CIE~$\Delta E_{76}$ color difference~\cite{robertson1977cie} between the predicted and target tristimulus values. Given a predicted output $\hat{t}$ and a target $t$, it is defined as:
\begin{equation}
    \mathcal{L}_{\Delta E_{76}} = \left\| \hat{t}^* - t^* \right\|_2,
    \label{eq:loss}
\end{equation}
where $\hat{t}^*$ and $t^*$ denote the CIE 
$L^*a^*b^*$~\cite{standard2007colorimetry} representation of $\hat{t}$ and $t$, respectively. The second is a smoothness regularization term that penalizes large 
gradients of the LUT $\mathcal{T}$ along each axis of the RGB input grid, encouraging smooth color transformations across the input domain:
\begin{equation}
    \mathcal{L}_{\text{smooth}} = \sum_{d \in \{R, G, B\}} 
    \left\| \nabla_d \mathcal{T} \right\|_2,
    \label{eq:smooth}
\end{equation}
where $\nabla_d$ denotes the gradient operator along dimension 
$d$. 
The total training objective is:
\begin{equation}
    \mathcal{L} = \mathcal{L}_{\Delta E_{76}} + \lambda\, \mathcal{L}_{\text{smooth}},
    \label{eq:total_loss}
\end{equation}
where $\lambda$ is a weighting factor. 

\begin{table*}[t]
    \centering
    \caption{Quantitative comparison of color correction methods on the 41M, KAUST and MCC datasets for three camera sensors (Canon~EOS~40D, Sony~Alpha~6100, Huawei~Mate~20~Pro). Results are reported in terms of CIE~$\Delta E_{00}$ and angular error ($^\circ$), including mean, standard deviation, and percentile statistics (25th, 50th, 95th). All methods are evaluated using reference illuminant white points. Best results are highlighted in \textbf{bold}.}
    \label{tab:overall_results}
    \begin{adjustbox}{width=\textwidth}
    \begin{NiceTabular}{ccccccccccccccccc}
         \toprule
         \Block{3-1}{Dataset} & \Block{3-1}{Method} & \multicolumn{15}{c}{$\Delta E_{00} \downarrow$}\\\cmidrule{3-17}
         & & \multicolumn{5}{c}{Canon} & \multicolumn{5}{c}{Sony} & \multicolumn{5}{c}{Huawei}\\\cmidrule(lr){3-7}\cmidrule(lr){8-12}\cmidrule(lr){13-17}
         & &  Mean & Std & 25th & 50th & 95th & Mean & Std & 25th & 50th & 95th & Mean & Std & 25th & 50th & 95th \\\midrule

         \Block{5-1}{41M} & CCM-CCT & 7.11 & 3.49 & 4.05 & 6.56 & 12.93 & 7.12 & 3.54 & 4.13 & 6.23 & 13.66 & 7.53 & 3.41 & 4.45 & 7.53 & 12.95\\
         & CST-MLP & 5.18 & 2.70 & 3.06 & 4.46 & 9.84 & 5.37 & 2.94 & 3.14 & 4.41 & 10.97 & 5.60 & 3.11 & 3.13 & 4.69 & 11.02\\
         & RP-MLP & 4.91 & 2.69 & 2.86 & 4.15 & 9.60 & 5.16 & 2.90 & 2.89 & 4.23 & 10.47 & 5.36 & 3.02 & 3.00 & 4.33 & 10.80\\
         & CCCNN & 4.91 & 2.50 & 3.02 & 4.07 & 9.45 & 5.18 & 2.76 & 3.07 & 4.26 & 10.22 & 5.31 & 3.02 & 3.03 & 4.29 & 11.29\\
         & C$^2$LUT (Ours) & \textbf{4.32} & \textbf{2.27} & \textbf{2.46} & \textbf{3.62} & \textbf{8.75} & \textbf{4.54} & \textbf{2.53} & \textbf{2.61} & \textbf{3.69} & \textbf{9.26} & \textbf{4.72} & \textbf{2.74} & \textbf{2.61} & \textbf{3.75} & \textbf{9.84}\\
         \cmidrule(lr){1-17}

         \Block{5-1}{KAUST} & CCM-CCT & 6.35 & 3.01 & 3.83 & 5.93 & 11.51 & 6.35 & 3.14 & 3.78 & 5.65 & 11.77 & 7.03 & 3.42 & 3.99 & 6.93 & 13.43\\
         & CST-MLP & 4.80 & 2.51 & 2.82 & 4.18 & 9.04 & 4.83 & 2.75 & 2.69 & 4.17 & 9.44 & 5.17 & 3.03 & 2.84 & 4.29 & 10.84\\
         & RP-MLP & 4.56 & 2.51 & 2.59 & 3.81 & 9.04 & 4.71 & 2.71 & 2.60 & 3.82 & 9.51 & 5.00 & 2.98 & 2.78 & 4.11 & 10.99\\
         & CCCNN & 4.62 & 2.34 & 2.72 & 3.94 & 8.94 & 4.78 & 2.63 & 2.75 & 3.96 & 9.48 & 4.97 & 2.97 & 2.74 & 3.99 & 11.30 \\
         & C$^2$LUT (Ours) & \textbf{4.36} & \textbf{2.20} & \textbf{2.66} & \textbf{3.65} & \textbf{8.57} & \textbf{4.50} & \textbf{2.40} & \textbf{2.66} & \textbf{3.69} & \textbf{9.17} & \textbf{4.76} & \textbf{2.77} & \textbf{2.70} & \textbf{3.74} & \textbf{10.19}\\
         \cmidrule(lr){1-17}
         
         \Block{5-1}{MCC} & CCM-CCT & 7.97 & 4.28 & 4.41 & 7.25 & 15.58 & 8.31 & 4.56 & 4.43 & 7.27 & 16.79 & 8.79 & 4.41 & 4.91 & 8.37 & 16.70\\
         & CST-MLP & 5.99 & 3.57 & 3.17 & 4.98 & 12.55 & 6.31 & 3.95 & 3.32 & 5.04 & 14.01 & 6.53 & 3.98 & 3.33 & 5.29 & 13.84\\
         & RP-MLP & 5.57 & 3.62 & 2.68 & 4.45 & 12.05 & 6.00 & 4.03 & 2.87 & 4.71 & 14.16 & 6.26 & 4.00 & 3.15 & 4.89 & 13.53\\
         & CCCNN & 5.32 & 3.30 & 2.80 & 4.07 & 11.96 & 5.99 & 3.81 & 3.00 & 4.58 & 13.82 & 5.96 & 3.90 & 2.84 & 4.70 & 13.73\\
         & C$^2$LUT (Ours) & \textbf{4.47} & \textbf{2.97} & \textbf{2.14} & \textbf{3.24} & \textbf{10.60} & \textbf{4.78} & \textbf{3.37} & \textbf{2.26} & \textbf{3.47} & \textbf{12.32} & \textbf{4.93} & \textbf{3.54} & \textbf{2.26} & \textbf{3.54} & \textbf{12.46}\\

         \midrule
         & & \multicolumn{15}{c}{Angular Error $(^\circ)\downarrow$}\\\cmidrule{3-17}

         \Block{5-1}{41M} & CCM-CCT & 4.51 & 2.05 & 2.67 & 4.46 & 8.11 & 4.63 & 2.19 & 2.65 & 4.47 & 8.68 & 5.14 & 2.38 & 2.98 & 5.07 & 9.10\\
         & CST-MLP & 3.16 & 1.49 & 1.95 & 2.95 & 5.56 & 3.30 & 1.58 & 2.01 & 3.06 & 5.88 & 3.35 & 1.71 & 1.97 & 3.14 & 6.23\\
         & RP-MLP & 2.99 & 1.47 & 1.78 & 2.78 & 5.31 & 3.13 & 1.56 & 1.87 & 2.89 & 5.67 & 3.22 & 1.70 & 1.80 & 3.09 & 6.25\\
         & CCCNN & 2.94 & 1.34 & 1.86 & 2.70 & 5.20 & 3.10 & 1.43 & 1.93 & 2.85 & 5.61 & 3.15 & 1.58 & 1.88 & 2.89 & 6.13\\
         & C$^2$LUT (Ours) & \textbf{2.64} & \textbf{1.27} & \textbf{1.61} & \textbf{2.45} & \textbf{4.86} & \textbf{2.78} & \textbf{1.37} & \textbf{1.67} & \textbf{2.55} & \textbf{5.10} & \textbf{2.85} & \textbf{1.52} & \textbf{1.64} & \textbf{2.54} & \textbf{5.42}\\
         \cmidrule(lr){1-17}

         \Block{5-1}{KAUST} & CCM-CCT & 4.16 & 1.89 & 2.45 & 3.92 & 7.73 & 4.21 & 2.03 & 2.39 & 3.99 & 8.03 & 4.74 & 2.30 & 2.69 & 4.46 & 8.65\\
         & CST-MLP & 3.10 & 1.45 & 1.97 & 2.83 & 5.59 & 3.10 & 1.55 & 1.90 & 2.89 & 5.71 & 3.18 & 1.67 & 1.89 & 2.91 & 6.01\\
         & RP-MLP & 2.95 & 1.46 & \textbf{1.83} & 2.68 & 5.56 & 2.99 & 1.55 & 1.82 & 2.73 & 5.33 & 3.11 & 1.69 & 1.82 & 2.80 & 6.24\\
         & CCCNN & 2.93 & 1.39 & 1.88 & 2.54 & \textbf{5.35} & 3.01 & 1.49 & 1.86 & 2.69 & 5.46 & 3.07 & 1.60 & 1.86 & 2.67 & 6.09\\
         & C$^2$LUT (Ours) & \textbf{2.80} & \textbf{1.32} & \textbf{1.83} & \textbf{2.44} & 5.46 & \textbf{2.85} & \textbf{1.40} & \textbf{1.80} & \textbf{2.56} & \textbf{5.25} & \textbf{2.94} & \textbf{1.55} & \textbf{1.80} & \textbf{2.56} & \textbf{5.70}\\
         \cmidrule(lr){1-17}

         \Block{5-1}{MCC} & CCM-CCT & 5.05 & 2.62 & 2.78 & 4.94 & 9.85 & 5.38 & 2.86 & 2.87 & 5.21 & 10.67 & 5.83 & 2.85 & 3.17 & 5.94 & 10.72\\
         & CST-MLP & 3.51 & 1.93 & 1.95 & 3.13 & 6.87 & 3.80 & 2.12 & 2.12 & 3.28 & 7.47 & 3.90 & 2.21 & 2.07 & 3.45 & 7.85\\
         & RP-MLP & 3.27 & 1.90 & 1.68 & 2.84 & 6.57 & 3.58 & 2.08 & 1.88 & 3.09 & 7.24 & 3.76 & 2.20 & 2.00 & 3.35 & 7.60\\
         & CCCNN & 3.15 & 1.71 & 1.73 & 2.62 & 6.25 & 3.47 & 1.91 & 1.86 & 2.84 & 6.90 & 3.46 & 1.99 & 1.84 & 2.92 & 7.17\\
         & C$^2$LUT (Ours) & \textbf{2.62} & \textbf{1.48} & \textbf{1.48} & \textbf{2.20} & \textbf{5.42} & \textbf{2.85} & \textbf{1.65} & \textbf{1.60} & \textbf{2.39} & \textbf{5.98} & \textbf{2.91} & \textbf{1.83} & \textbf{1.53} & \textbf{2.31} & \textbf{6.57}\\
         \bottomrule
         
    \end{NiceTabular}
    \end{adjustbox}
\end{table*}
\section{Experiments}

\begin{figure*}[t]
    \centering
    \setlength{\tabcolsep}{1pt}
    \begin{tabular}{ccccccccc}
    
    {\small Illuminant} &
    {\small RAW Input} &
    &
    {\small CCM-CCT} &
    {\small CST-MLP} &
    {\small RP-MLP} &
    {\small CCCNN} &
    {\small C$^2$LUT (Ours)} &
    {\small Reference} \\
    
    \includegraphics[width=0.15\textwidth]{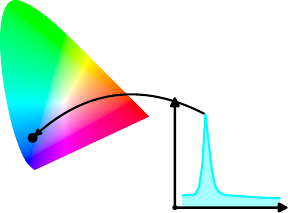} &
    \includegraphics[width=0.11\textwidth]{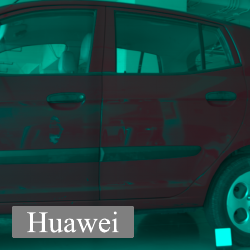} &
    \includegraphics[width=0.0089\textwidth]{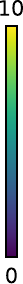} &
    \includegraphics[width=0.11\textwidth]{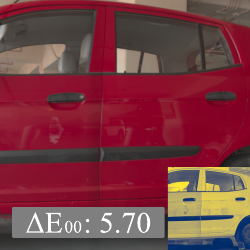} &
    \includegraphics[width=0.11\textwidth]{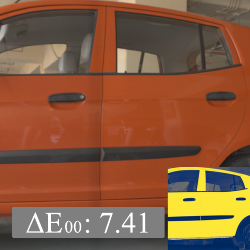} &
    \includegraphics[width=0.11\textwidth]{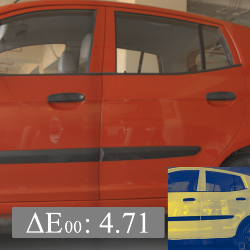} &
    \includegraphics[width=0.11\textwidth]{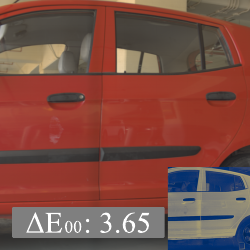} &
    \includegraphics[width=0.11\textwidth]{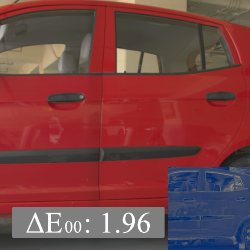} &
    \includegraphics[width=0.11\textwidth]{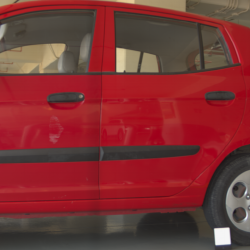}
    \\
    \includegraphics[width=0.15\textwidth]{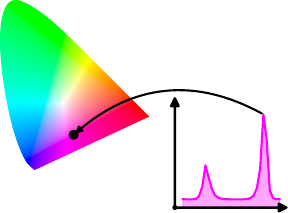} &
    \includegraphics[width=0.11\textwidth]{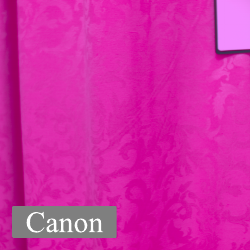} &
    \includegraphics[width=0.0089\textwidth]{Figures/KAUST/cbar.pdf} &
    \includegraphics[width=0.11\textwidth]{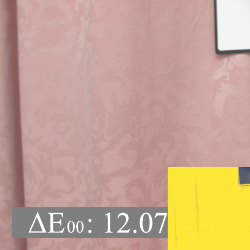} &
    \includegraphics[width=0.11\textwidth]{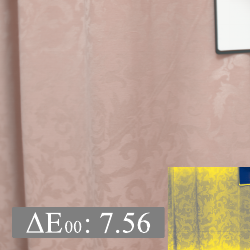} &
    \includegraphics[width=0.11\textwidth]{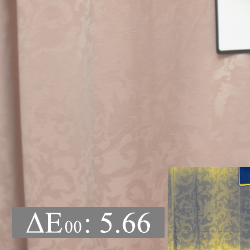} &
    \includegraphics[width=0.11\textwidth]{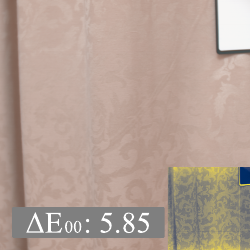} &
    \includegraphics[width=0.11\textwidth]{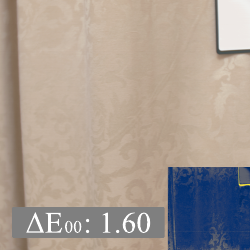} &
    \includegraphics[width=0.11\textwidth]{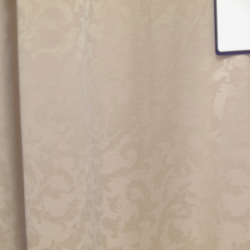}
    \\
    \includegraphics[width=0.15\textwidth]{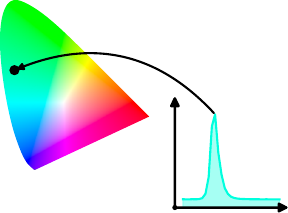} &
    \includegraphics[width=0.11\textwidth]{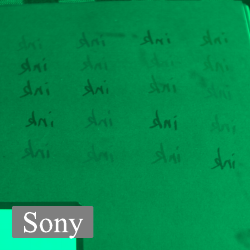} &
    \includegraphics[width=0.0089\textwidth]{Figures/KAUST/cbar.pdf} &
    \includegraphics[width=0.11\textwidth]{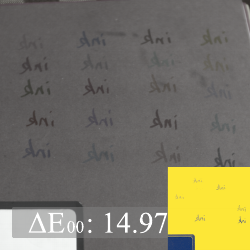} &
    \includegraphics[width=0.11\textwidth]{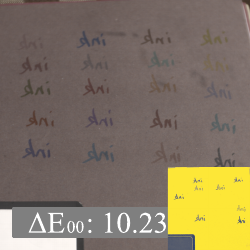} &
    \includegraphics[width=0.11\textwidth]{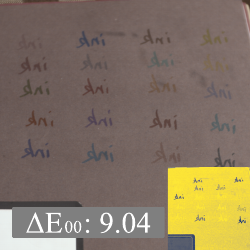} &
    \includegraphics[width=0.11\textwidth]{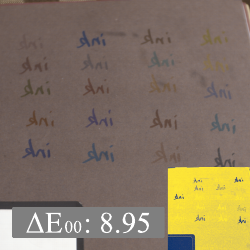} &
    \includegraphics[width=0.11\textwidth]{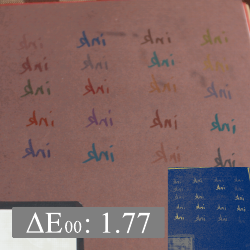} &
    \includegraphics[width=0.11\textwidth]{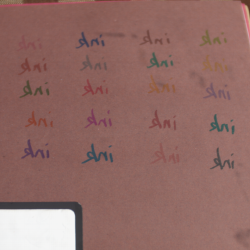}
    
    \end{tabular}
    \caption{Qualitative comparison of color correction results under challenging test illuminants. Results are shown for the compared methods on images rendered using three different camera SSFs. Each corrected image is associated with a per-pixel $\Delta E_{00}$ map in range [0-10], which helps visualize the spatial distribution of color correction errors across the image.
    The proposed C$^2$LUT framework produces colors that are visually closer to the reference and achieves the lowest CIE~$\Delta E_{00}$ across all camera models.}
    \label{fig:qualitative_kaust}
\end{figure*}

\subsection{Setup}
\label{sec:setup}

We need three elements to obtain RAW-XYZ pairs: (1) camera SSFs and CIE~XYZ~\cite{CIE19} CMFs, (2) illuminant SPDs and (3) surface reflectance spectra.
The camera SSFs are taken from the Image Engineering dataset~\cite{imageengineering} and correspond to two DSLR/mirrorless cameras (Canon~EOS~40D and Sony~Alpha~6100) and one smartphone camera (Huawei~Mate~20~Pro). For the illuminants, we use the dataset described in \Cref{sec:dataset}. For surface reflectances, we use three datasets. 
The first is the 41 Million (41M) reflectances dataset of Zhang et al.~\cite{zhang2016metamer} which we randomly partition into train, validation, and test splits. Since this dataset consists of individual reflectance spectra rather than images, we randomly arrange reflectances into $256 \times 256$ images for processing. The second is the KAUST dataset~\cite{li2021multispectral}, which comprises 409 hyperspectral images with shape $512 \times 512$. The third is the Macbeth Color Checker (MCC)~\cite{mccamy1976color}, a standard 24-patche colorimetric target. 
All spectral data are sampled over the wavelengths range 400--700\,nm 
with steps of 10\,nm, resulting in 31 spectral bands. 
Training is performed using the 41M train split, while KAUST and MCC are used exclusively for testing, making our evaluation cross-dataset. Evaluation is also performed on 41M test split. For each combination of image and illuminant, we render the raw camera response $\rho$ and the corresponding XYZ 
target $t$ via spectral rendering as described in \Cref{sec:data_generation}.

We evaluate color correction accuracy using two metrics. 
The first is the CIE~$\Delta E_{00}$~\cite{sharma2005ciede2000}, which measures perceptual color differences in the CIE~$L^*a^*b^*$ color space and accounts for the non-uniformity of human color perception through dedicated weighting functions for lightness, chroma, and hue. The second is the angular error (AE), defined as the angle between the predicted and target XYZ tristimulus vectors and measures chromatic dissimilarity independently of luminance. For both metrics, lower values indicate better performance. Note that we optimize the models using the CIE~$\Delta E_{76}$ color difference as a proxy for CIE~$\Delta E_{00}$, as the latter exhibits numerical instability when used as the loss function.

We compare our framework against four baselines. CCM-CCT~\cite{dng} 
is the standard in-camera CST, which interpolates between two pre-calibrated $3\times3$ CCMs associated with the D65 and 
CIE Standard Illuminant A, using the estimated CCT of the scene illuminant. 
CST-MLP~\cite{tedla2025off} is an MLP that predicts a $3\times3$ CCM from the illuminant chromaticity. 
Finlayson et al.~\cite{finlayson2015color} is a root-polynomial (RP) 
color correction model, which constructs a 22-dimensional polynomial expansion of the raw-RGB values and applies 
a $3\times 22$ correction matrix $M$ to predict the 
XYZ output. We extend the original formulation by using an MLP to predict $M$ from the illuminant chromaticity, as done in CST-MLP, thus we call this method RP-MLP. CCCNN~\cite{macdonald2021camera} is a pixel-wise MLP that takes as input the raw-RGB values of a pixel and the illuminant CCT, and directly predicts the 
corresponding XYZ value. Following~\cite{tedla2025off}, we replace CCT with illuminant chromaticity in CCCNN and adopt this chromaticity-conditioned variant in our experiments.

\begin{figure*}[t]
    \centering
    \setlength{\tabcolsep}{1pt}
    \begin{tabular}{ccccccc}
    
    {\small Illuminant} &
    {\small RAW Input} &
    {\small CCM-CCT} &
    {\small CST-MLP} &
    {\small RP-MLP} &
    {\small CCCNN} &
    {\small C$^2$LUT (Ours)} \vspace{-4mm}\\

    \includegraphics[width=0.15\textwidth]{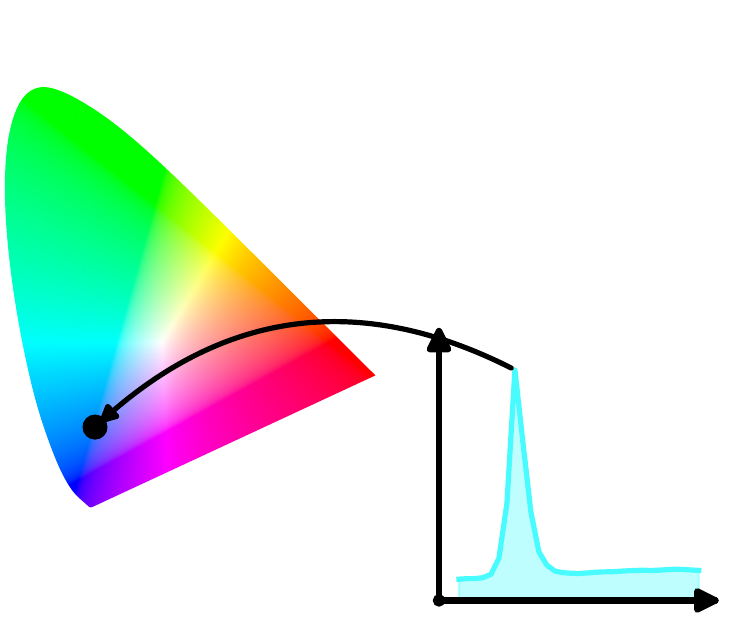} &
    \includegraphics[width=0.13\textwidth]{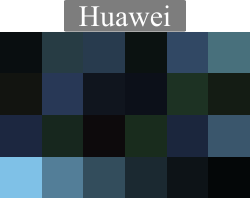} &
    \includegraphics[width=0.13\textwidth]{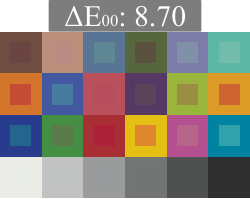} &
    \includegraphics[width=0.13\textwidth]{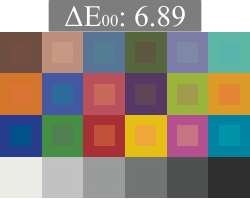} &
    \includegraphics[width=0.13\textwidth]{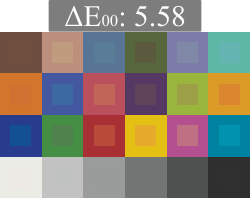} &
    \includegraphics[width=0.13\textwidth]{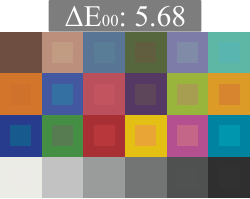} &
    \includegraphics[width=0.13\textwidth]{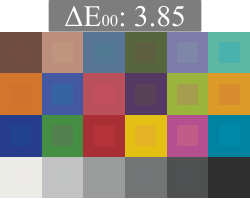}
    \vspace{-2mm}
    \\ 
    \includegraphics[width=0.15\textwidth]{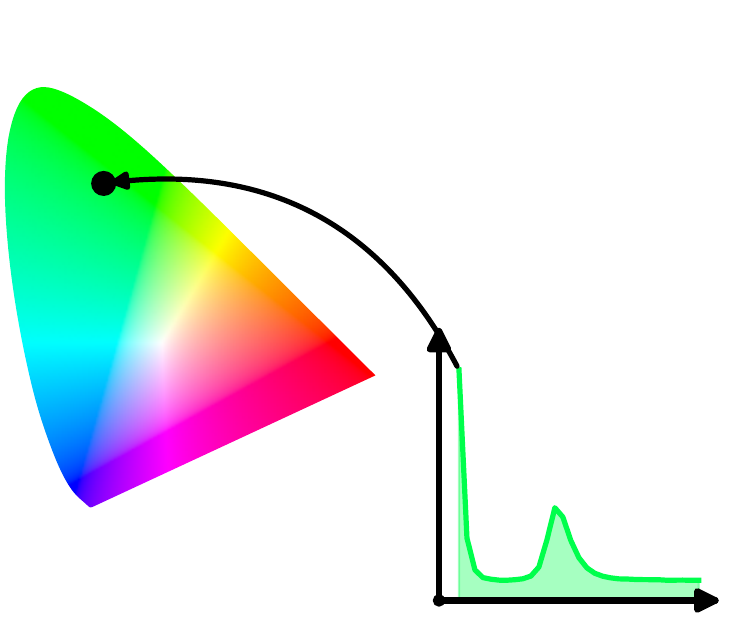} &
    \includegraphics[width=0.13\textwidth]{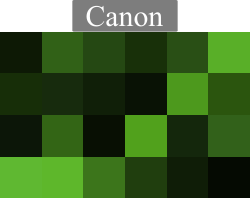} &
    \includegraphics[width=0.13\textwidth]{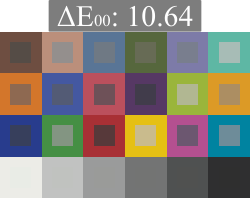} &
    \includegraphics[width=0.13\textwidth]{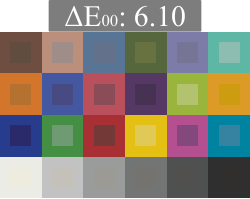} &
    \includegraphics[width=0.13\textwidth]{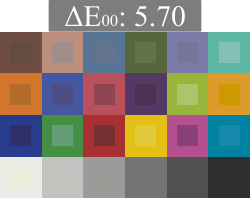} &
    \includegraphics[width=0.13\textwidth]{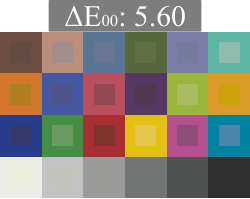} &
    \includegraphics[width=0.13\textwidth]{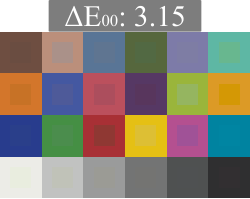}
    \vspace{-2mm}
    \\
    \includegraphics[width=0.15\textwidth]{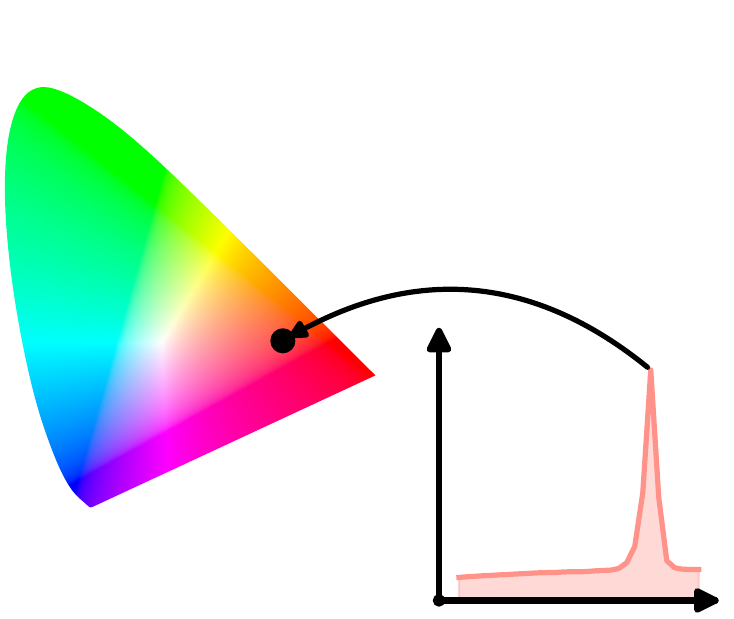} &
    \includegraphics[width=0.13\textwidth]{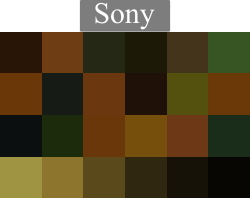} &
    \includegraphics[width=0.13\textwidth]{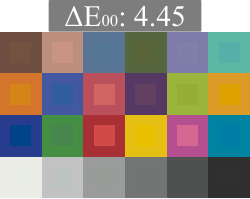} &
    \includegraphics[width=0.13\textwidth]{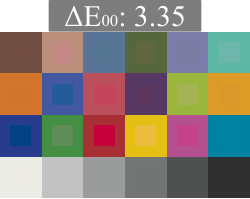} &
    \includegraphics[width=0.13\textwidth]{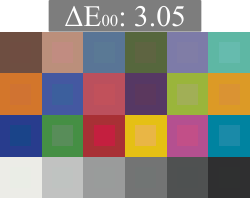} &
    \includegraphics[width=0.13\textwidth]{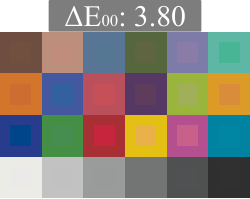} &
    \includegraphics[width=0.13\textwidth]{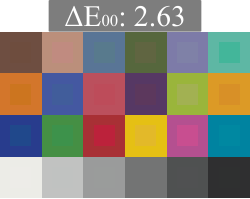}
    \end{tabular}
    \caption{Qualitative comparison of color correction methods on Macbeth Color Checker patches rendered under challenging test illuminants. Each corrected patch includes an inset square representing the reference color for direct comparison. For each camera, the proposed C$^2$LUT method produces colors that more closely match the target, resulting in the lowest CIE~$\Delta E_{00}$ among all compared approaches.
    \vspace{-2mm}
    }
    \label{fig:qualitative_mcc}
\end{figure*}

\subsection{Implementation details}

All models are trained on the training split of the 41M~\cite{zhang2016metamer} 
dataset, with the exception of CCM-CCT~\cite{dng}, which relies on a fixed interpolation 
scheme. All learning-based methods share the same MLP architecture, with the input and output dimensions adapted to the underlying correction strategy.
Specifically, CST-MLP~\cite{tedla2025off} predicts a $3\times3$ CCM (9 outputs), 
RP-MLP~\cite{finlayson2015color} predicts a $3\times 22$ CCM (66 outputs), and our C$^2$LUT predicts 
$B$ interpolation weights for the basis cores ($B$ outputs). These models take only the illuminant chromaticity as input (2 inputs). Unlike the others, CCCNN~\cite{macdonald2021camera} takes both the raw pixel value and the illuminant chromaticity as input (5 inputs) and directly predicts the XYZ output (3 outputs). All these models have a single 32-dimension hidden layer.

The basis cores and factor matrices of C$^2$LUT are initialized by performing the Tucker tensor decomposition of the identity LUT, so that the model starts from a neutral color correction at the beginning of training. Since the interpolation weights are equal to $\frac{1}{B}$ at initialization, all $B$ cores 
start identical. We empirically set the Tucker rank to $R=17$, the number of cores $B=12$, and the LUT resolution to $N=65$ (see ablation study in \Cref{sec:ablations}).

We train all the models 
until convergence
using a batch size of 16, where each batch element consists of a $256\times256$ image of randomly sampled reflectances rendered under a randomly sampled illuminant. We use the Adam optimizer with an initial learning rate of $10^{-3}$, decayed to $10^{-6}$ following a cosine annealing scheme. For the loss function $\mathcal{L}$ in \Cref{eq:total_loss}, we empirically set $\lambda$ to $10^2$ for C$^2$LUT to compensate for the different orders of magnitude of the two loss components, thereby balancing their contributions during training. For all other models, which do not incorporate a LUT, we set $\lambda=0$, such that only $\mathcal{L}_{\Delta E_{76}}$ is optimized. A separate model is trained for each of the three cameras considered in this work.

\subsection{Color correction results}
\label{subsec:colorcorrectionknownilluminants}
\begin{table*}[t]
    \centering
    \caption{Quantitative comparison of color correction methods on the 41M dataset, where test illuminants are divided into three subsets according to their CIE~xy-chromaticity distance from the reference D65 illuminant (Close, Mid, Far). Results are reported in terms of CIE~$\Delta E_{00}$ and angular error ($^\circ$), including mean, standard deviation, and percentile statistics (25th, 50th, 95th). All methods are evaluated across three camera sensors (Canon~EOS~40D, Sony~Alpha~6100, Huawei~Mate~20~Pro). Best results are highlighted in \textbf{bold}.}
    \label{tab:results_distancefromD65}
    \begin{adjustbox}{width=\textwidth}
    \begin{NiceTabular}{ccccccccccccccccc}
         \toprule
         \Block{3-1}{Illuminant} & \Block{3-1}{Method} & \multicolumn{15}{c}{$\Delta E_{00} \downarrow$}\\\cmidrule{3-17}
         & & \multicolumn{5}{c}{Canon} & \multicolumn{5}{c}{Sony} & \multicolumn{5}{c}{Huawei}\\\cmidrule(lr){3-7}\cmidrule(lr){8-12}\cmidrule(lr){13-17}
         & &  Mean & Std & 25th & 50th & 95th & Mean & Std & 25th & 50th & 95th & Mean & Std & 25th & 50th & 95th \\\midrule

         \Block{5-1}{Close} & CCM-CCT & 4.37 & 2.12 & 3.11 & 3.89 & 8.56 & 4.26 & 1.83 & 3.28 & 3.96 & 8.25 & 4.58 & 2.18 & 3.15 & 4.08 & 9.16\\
         & CST-MLP & 3.35 & 1.94 & 2.11 & 3.06 & 6.98 & 3.38 & 1.90 & 2.31 & 3.06 & 6.78 & 3.39 & 2.00 & 2.23 & 2.99 & 7.10\\
         & RP-MLP & 3.11 & 1.92 & 1.98 & 2.66 & 6.51 & 3.24 & 1.86 & 2.07 & 2.79 & 6.54 & 3.28 & 1.95 & 2.08 & 2.92 & 6.70\\
         & CCCNN & 3.22 & 1.65 & 2.15 & 2.93 & 6.33 & 3.25 & 1.64 & 2.23 & 3.02 & 6.35 & 3.27 & 1.86 & 2.21 & 2.90 & 6.11\\
         & C$^2$LUT (Ours) & \textbf{2.84} & \textbf{1.63} & \textbf{1.83} & \textbf{2.39} & \textbf{5.99} & \textbf{2.90} & \textbf{1.61} & \textbf{1.96} & \textbf{2.50} & \textbf{5.81} & \textbf{3.00} & \textbf{1.85} & \textbf{1.91} & \textbf{2.54} & \textbf{5.80}\\\cmidrule(lr){1-17}

         \Block{5-1}{Mid} & CCM-CCT & 7.50 & 2.61 & 5.55 & 7.47 & 11.62 & 7.21 & 2.40 & 5.34 & 7.20 & 11.49 & 8.09 & 2.43 & 6.24 & 8.49 & 11.58\\
         & CST-MLP & 5.36 & 2.39 & 3.82 & 4.73 & 9.58 & 5.31 & 2.54 & 3.48 & 4.95 & 9.40 & 5.57 & 2.60 & 3.54 & 5.16 & 9.47\\
         & RP-MLP & 5.07 & 2.42 & 3.37 & 4.48 & 9.43 & 5.11 & 2.59 & 3.28 & 4.46 & 9.02 & 5.32 & 2.58 & 3.36 & 4.70 & 9.37\\
         & CCCNN & 4.85 & 2.06 & 3.40 & 4.41 & 8.43 & 4.98 & 2.28 & 3.50 & 4.47 & 8.84 & 5.02 & 2.46 & 3.32 & 4.36 & 8.67\\
         & C$^2$LUT (Ours) & \textbf{4.27} & \textbf{1.91} & \textbf{2.80} & \textbf{3.78} & \textbf{7.57} & \textbf{4.38} & \textbf{2.07} & \textbf{2.85} & \textbf{3.88} & \textbf{7.94} & \textbf{4.49} & \textbf{2.23} & \textbf{2.92} & \textbf{3.93} & \textbf{8.03}\\\cmidrule(lr){1-17}

         \Block{5-1}{Far} & CCM-CCT & 9.38 & 3.47 & 6.48 & 9.83 & 14.23 & 9.78 & 3.57 & 6.62 & 10.01 & 15.50 & 9.88 & 3.12 & 7.84 & 10.39 & 14.18\\
         & CST-MLP & 6.79 & 2.55 & 4.53 & 7.37 & 10.35 & 7.35 & 2.79 & 4.97 & 7.84 & 11.46 & 7.77 & 2.92 & 5.47 & 8.44 & 12.01\\
         & RP-MLP & 6.53 & 2.51 & 4.27 & 7.15 & 9.83 & 7.06 & 2.77 & 4.46 & 7.57 & 11.17 & 7.43 & 2.88 & 5.05 & 8.03 & 11.99\\
         & CCCNN & 6.60 & 2.43 & 4.43 & 7.17 & 10.02 & 7.26 & 2.57 & 4.91 & 7.76 & 11.07 & 7.55 & 2.87 & 4.90 & 7.62 & 11.98\\
         & C$^2$LUT (Ours) & \textbf{5.81} & \textbf{2.18} & \textbf{3.71} & \textbf{6.25} & \textbf{9.01} & \textbf{6.30} & \textbf{2.53} & \textbf{3.94} & \textbf{6.36} & \textbf{10.04} & \textbf{6.61} & \textbf{2.73} & \textbf{4.42} & \textbf{6.64} & \textbf{10.74}\\

         \midrule
         & & \multicolumn{15}{c}{Angular Error $(^\circ)\downarrow$}\\\cmidrule{3-17}
        
         \Block{5-1}{Close} & CCM-CCT & 2.98 & 1.32 & 2.18 & 2.71 & 5.25 & 2.96 & 1.29 & 2.10 & 2.72 & 5.09 & 3.24 & 1.55 & 2.18 & 2.96 & 6.17\\
         & CST-MLP & 2.24 & 1.21 & 1.39 & 2.05 & 4.41 & 2.28 & 1.23 & 1.39 & 2.07 & 4.48 & 2.18 & 1.25 & 1.30 & 2.01 & 4.50\\
         & RP-MLP & 2.08 & 1.19 & 1.25 & 1.94 & 4.14 & 2.17 & 1.22 & 1.27 & 1.94 & 4.21 & 2.10 & 1.23 & 1.21 & 1.85 & 4.27\\
         & CCCNN & 2.09 & \textbf{1.03} & 1.37 & 1.89 & 3.87 & 2.14 & \textbf{1.06} & 1.44 & 1.92 & 4.01 & 2.11 & \textbf{1.11} & 1.32 & 1.87 & 4.18\\
         & C$^2$LUT (Ours) & \textbf{1.87} & \textbf{1.03} &\textbf{ 1.20} & \textbf{1.66} & \textbf{3.55} & \textbf{1.93} & \textbf{1.06} & \textbf{1.23} & \textbf{1.66} & \textbf{3.84} & \textbf{1.92} & 1.12 & \textbf{1.19} & \textbf{1.65} & \textbf{3.98}\\\cmidrule(lr){1-17}

         \Block{5-1}{Mid} & CCM-CCT & 5.28 & 1.68 & 4.46 & 5.48 & 8.02 & 5.34 & 1.82 & 4.17 & 5.63 & 8.02 & 5.95 & 1.85 & 4.93 & 6.35 & 8.53\\
         & CST-MLP & 3.49 & 1.32 & 2.47 & 3.61 & 5.49 & 3.54 & 1.43 & 2.36 & 3.63 & 6.01 & 3.53 & 1.44 & 2.44 & 3.64 & 6.12\\
         & RP-MLP & 3.29 & 1.30 & 2.28 & 3.33 & 5.13 & 3.35 & 1.41 & 2.26 & 3.43 & 5.69 & 3.39 & 1.45 & 2.24 & 3.38 & 5.97\\
         & CCCNN & 3.11 & 1.15 & 2.14 & 3.07 & 4.97 & 3.24 & 1.26 & 2.19 & 3.25 & 5.30 & 3.17 & 1.28 & 2.11 & 3.15 & 5.42\\
         & C$^2$LUT (Ours) & \textbf{2.80} & \textbf{1.09} & \textbf{1.90} & \textbf{2.76} & \textbf{4.95} & \textbf{2.88} & \textbf{1.20} & \textbf{1.92} & \textbf{2.75} & \textbf{5.07} & \textbf{2.88} & \textbf{1.23} & \textbf{1.95} & \textbf{2.79} & \textbf{5.34}\\\cmidrule(lr){1-17}

         \Block{5-1}{Far} & CCM-CCT & 5.27 & 2.14 & 3.71 & 5.40 & 8.68 & 5.56 & 2.29 & 3.56 & 5.66 & 9.03 & 6.22 & 2.42 & 4.74 & 6.31 & 9.47\\
         & CST-MLP & 3.73 & 1.49 & 2.52 & 4.00 & 5.67 & 4.05 & 1.51 & 3.01 & 4.24 & 5.80 & 4.31 & 1.68 & 3.14 & 4.60 & 6.68\\
         & RP-MLP & 3.59 & 1.48 & 2.34 & 3.80 & 5.51 & 3.85 & 1.52 & 2.71 & 4.13 & 5.83 & 4.14 & 1.72 & 3.12 & 4.29 & 6.46\\
         & CCCNN & 3.61 & 1.35 & 2.46 & 3.79 & 5.41 & 3.89 & \textbf{1.37} & 2.83 & 4.16 & 5.76 & 4.14 & \textbf{1.58} & 2.97 & 4.34 & \textbf{6.32}\\
         & C$^2$LUT (Ours) & \textbf{3.24} & \textbf{1.27} & \textbf{2.02} & \textbf{3.31} & \textbf{4.99} & \textbf{3.49} & \textbf{1.37} & \textbf{2.43} & \textbf{3.74} & \textbf{5.60} & \textbf{3.74} & 1.59 & \textbf{2.54} & \textbf{3.88} & 6.36\\
         \bottomrule
         
    \end{NiceTabular}
    \end{adjustbox}
    \vspace{-2mm}
\end{table*}

\begin{table*}[t]
    \centering
    \caption{Quantitative comparison of color correction methods on the 41M dataset
    , where test illuminants are categorized according to spectral bandwidth (broad-band, mid-band, narrow-band) based on the area under the normalized SPD. Results are reported in terms of CIE~$\Delta E_{00}$ and angular error ($^\circ$), including mean, standard deviation, and percentile statistics (25th, 50th, 95th). All methods are evaluated across three camera sensors (Canon~EOS~40D, Sony~Alpha~6100, Huawei~Mate~20~Pro). Best results are highlighted in \textbf{bold}.}
    \label{tab:results_AUC}
    \begin{adjustbox}{width=\textwidth}
    \begin{NiceTabular}{ccccccccccccccccc}
         \toprule
         \Block{3-1}{Illuminant} & \Block{3-1}{Method} & \multicolumn{15}{c}{$\Delta E_{00} \downarrow$}\\\cmidrule{3-17}
         & & \multicolumn{5}{c}{Canon} & \multicolumn{5}{c}{Sony} & \multicolumn{5}{c}{Huawei}\\\cmidrule(lr){3-7}\cmidrule(lr){8-12}\cmidrule(lr){13-17}
         & &  Mean & Std & 25th & 50th & 95th & Mean & Std & 25th & 50th & 95th & Mean & Std & 25th & 50th & 95th \\\midrule

         \Block{5-1}{Broad-band} & CCM-CCT & 4.52 & 1.98 & 3.35 & 4.17 & 7.67 & 4.48 & 1.85 & 3.43 & 4.12 & 8.08 & 5.01 & 2.67 & 3.48 & 4.22 & 9.97\\
         & CST-MLP & 3.56 & 1.76 & 2.31 & 3.16 & 7.51 & 3.54 & 1.70 & 2.33 & 3.19 & 7.42 & 3.69 & 2.04 & 2.25 & 3.21 & 7.69\\
         & RP-MLP & 3.31 & 1.82 & 2.07 & 2.86 & 7.49 & 3.38 & 1.68 & 2.14 & 2.86 & 7.17 & 3.54 & 1.98 & 2.14 & 3.00 & 7.53\\
         & CCCNN & 3.37 & \textbf{1.54} & 2.27 & 3.09 & 6.89 & 3.40 & 1.59 & 2.36 & 3.03 & 7.36 & 3.47 & 1.85 & 2.26 & 3.00 & 7.27\\
         & C$^2$LUT (Ours) & \textbf{2.99} & 1.60 &\textbf{ 1.90} & \textbf{2.46} & \textbf{6.64} & \textbf{3.01} & \textbf{1.51} & \textbf{2.00} & \textbf{2.56} & \textbf{6.58} & \textbf{3.13} & \textbf{1.73} & \textbf{1.99} & \textbf{2.56} & \textbf{6.85}\\\cmidrule(lr){1-17}

         \Block{5-1}{Mid-band} & CCM-CCT & 7.66 & 2.78 & 5.51 & 8.35 & 11.77 & 7.38 & 2.71 & 5.29 & 7.47 & 11.91 & 7.97 & 2.59 & 6.10 & 8.43 & 11.94 \\
         & CST-MLP & 5.28 & 2.44 & 3.38 & 4.80 & 9.28 & 5.27 & 2.52 & 3.32 & 4.64 & 9.73 & 5.31 & 2.51 & 3.46 & 4.59 & 9.44\\
         & RP-MLP & 4.94 & 2.42 & 3.08 & 4.29 & 9.43 & 5.02 & 2.53 & 2.99 & 4.26 & 9.14 & 5.08 & 2.46 & 3.16 & 4.18 & 9.16\\
         & CCCNN & 4.87 & 2.14 & 3.11 & 4.43 & 8.39 & 4.96 & 2.25 & 3.27 & 4.37 & 9.33 & 4.87 & 2.31 & 3.11 & 4.20 & 9.10\\
         & C$^2$LUT (Ours) & \textbf{4.20} & \textbf{1.95} & \textbf{2.69} & \textbf{3.70} & \textbf{7.81} & \textbf{4.34} & \textbf{2.11} & \textbf{2.73} & \textbf{3.90} & \textbf{8.44} & \textbf{4.29} & \textbf{2.04} & \textbf{2.68} & \textbf{3.70} & \textbf{8.04}\\\cmidrule(lr){1-17}

         \Block{5-1}{Narrow-band} & CCM-CCT & 9.17 & 3.74 & 5.59 & 9.57 & 14.55 & 9.50 & 3.80 & 5.73 & 10.00 & 15.70 & 9.60 & 3.22 & 7.10 & 10.07 & 14.14\\
         & CST-MLP & 6.71 & 2.83 & 4.48 & 6.42 & 11.24 & 7.30 & 3.13 & 4.82 & 7.12 & 12.68 & 7.77 & 3.21 & 5.18 & 8.29 & 13.33\\
         & RP-MLP & 6.48 & 2.76 & 4.01 & 6.15 & 11.13 & 7.07 & 3.07 & 4.46 & 7.16 & 12.43 & 7.44 & 3.13 & 4.86 & 7.83 & 13.04\\
         & CCCNN & 6.50 & 2.65 & 4.06 & 5.99 & 10.69 & 7.17 & 2.86 & 4.83 & 7.39 & 11.85 & 7.55 & 3.19 & 4.90 & 7.57 & 12.61\\
         & C$^2$LUT (Ours) & \textbf{5.74} & \textbf{2.34} & \textbf{3.68} & \textbf{5.47} & \textbf{9.12} & \textbf{6.28} & \textbf{2.70} & \textbf{3.84} & \textbf{6.32} & \textbf{11.13} & \textbf{6.71} & \textbf{2.97} & \textbf{4.28} & \textbf{6.30} & \textbf{11.66}\\

         \midrule
         & & \multicolumn{15}{c}{Angular Error $(^\circ)\downarrow$}\\\cmidrule{3-17}
        
         \Block{5-1}{Broad-band} & CCM-CCT & 3.16 & 1.51 & 2.22 & 2.76 & 5.94 & 3.09 & 1.39 & 2.16 & 2.76 & 5.45 & 3.71 & 2.20 & 2.14 & 2.99 & 8.39\\
         & CST-MLP & 2.31 & 1.16 & 1.46 & 2.08 & 4.13 & 2.35 & 1.19 & 1.53 & 2.12 & 4.30 & 2.38 & 1.48 & 1.37 & 2.09 & 5.09\\
         & RP-MLP & 2.15 & 1.15 & 1.28 & 1.97 & 4.11 & 2.21 & 1.13 & 1.32 & 1.94 & 4.15 & 2.26 & 1.44 & 1.29 & 1.92 & 4.78\\
         & CCCNN & 2.15 & 1.02 & 1.46 & 1.91 & 3.86 & 2.19 & 1.02 & 1.43 & 2.02 & 4.03 & 2.25 & 1.30 & 1.35 & 1.91 & 4.70\\
         & C$^2$LUT (Ours) & \textbf{1.93} & \textbf{0.99} & \textbf{1.21} & \textbf{1.73} & \textbf{3.72} & \textbf{1.98} & \textbf{1.01} & \textbf{1.23} & \textbf{1.73} & \textbf{3.59} & \textbf{2.04} & \textbf{1.23} & \textbf{1.23} & \textbf{1.66} & \textbf{4.22}\\\cmidrule(lr){1-17}

         \Block{5-1}{Mid-band} & CCM-CCT & 5.19 & 1.91 & 3.72 & 5.52 & 8.04 & 5.28 & 2.04 & 3.45 & 5.65 & 8.67 & 5.76 & 2.06 & 4.26 & 6.13 & 8.80\\
         & CST-MLP & 3.40 & 1.44 & 2.37 & 3.36 & 5.57 & 3.46 & 1.47 & 2.20 & 3.33 & 6.06 & 3.40 & 1.50 & 2.29 & 3.21 & 5.94\\
         & RP-MLP & 3.17 & 1.38 & 2.18 & 3.16 & 5.17 & 3.25 & 1.43 & 2.09 & 3.09 & 5.70 & 3.26 & 1.49 & 2.20 & 3.09 & 5.70\\
         & CCCNN & 3.05 & 1.24 & 2.09 & 2.93 & 5.31 & 3.18 & 1.28 & 2.11 & 3.06 & 5.38 & 3.07 & 1.27 & 2.02 & 2.94 & 5.09\\
         & C$^2$LUT (Ours) & \textbf{2.70} & \textbf{1.13} & \textbf{1.81} & \textbf{2.42} & \textbf{4.61} & \textbf{2.80} & \textbf{1.20} & \textbf{1.80} & \textbf{2.59} & \textbf{5.04} & \textbf{2.73} & \textbf{1.20} & \textbf{1.81} & \textbf{2.40} & \textbf{5.14}\\\cmidrule(lr){1-17}

         \Block{5-1}{Narrow-band} & CCM-CCT & 5.18 & 2.00 & 3.80 & 5.28 & 8.51 & 5.52 & 2.19 & 4.01 & 5.61 & 9.12 & 5.94 & 2.23 & 4.44 & 6.13 & 9.46\\
         & CST-MLP & 3.75 & 1.48 & 2.55 & 3.85 & 5.98 & 4.07 & 1.57 & 2.88 & 4.16 & 6.21 & 4.27 & 1.60 & 3.19 & 4.37 & 7.13\\
         & RP-MLP & 3.63 & 1.48 & 2.41 & 3.72 & 5.80 & 3.92 & 1.59 & 2.71 & 3.87 & \textbf{6.05} & 4.12 & 1.64 & 3.12 & 4.10 & 6.55\\
         & CCCNN & 3.62 & 1.33 & 2.51 & 3.51 & 5.78 & 3.92 & \textbf{1.42} & 2.79 & 3.74 & 6.14 & 4.13 & \textbf{1.56} & 3.01 & 4.24 & 6.77\\
         & C$^2$LUT (Ours) & \textbf{3.28} & \textbf{1.30} & \textbf{2.11} & \textbf{3.23} & \textbf{5.24} & \textbf{3.54} & \textbf{1.42} & \textbf{2.52} & \textbf{3.70} & 6.21 & \textbf{3.78} & 1.57 & \textbf{2.62} & \textbf{3.84} & \textbf{6.51}\\
         \bottomrule
         
    \end{NiceTabular}
    \end{adjustbox}
\end{table*}

The experimental results are shown in \Cref{tab:overall_results}, which reports $\Delta E_{00}$~\cite{sharma2005ciede2000} and AE for all methods across three cameras and three reflectance datasets. We observe that C$^2$LUT consistently achieves the lowest mean error across all cameras and datasets for both metrics, demonstrating the benefit of combining a non-linear color correction framework based on 3D~LUTs with a chromaticity-aware illuminant representation.

C$^2$LUT consistently outperforms all competing methods across all 
datasets, with improvements over CCCNN~\cite{macdonald2021camera} 
ranging from 4--6\% on KAUST~\cite{li2021multispectral} to 11--12\% on 41M~\cite{zhang2016metamer} and up to 20\% on 
MCC~\cite{mccamy1976color} in terms of mean $\Delta E_{00}$, and 
from 2--7\% on KAUST to 10\% on 41M and up to 18\% on MCC in terms 
of mean angular error. C$^2$LUT achieves the lowest standard deviation in all configurations, indicating better robustness to varying illumination conditions compared to all competing methods. In addition, the improvement is consistent across the full error distribution: C$^2$LUT has the lowest error also when measuring the 25th and 50th percentiles, confirming that the advantage is systematic. Finally, C$^2$LUT achieves the best 95th percentile in most configurations, demonstrating that the proposed framework maintains its advantage even under the most challenging illumination conditions in the test set. It is worth mentioning that, since KAUST and MCC are used exclusively for testing, the results demonstrate strong cross-dataset generalization of the proposed approach.

\Cref{fig:qualitative_kaust,fig:qualitative_mcc} show qualitative comparisons of color correction results on the KAUST and MCC datasets, respectively. Additional visualizations are provided in the supplementary material. Each row corresponds to a different camera and shows results for illuminants from distinct chromaticity regions: reddish, bluish, and greenish. The visual results produced by the proposed C$^2$LUT are consistently closer to the reference, while the other methods exhibit noticeable color shifts across image regions and multiple patches. This is further confirmed by the lower per-pixel $\Delta E_{00}$ values in the error maps of \Cref{fig:qualitative_kaust}, the reduced visual discrepancy between the corrected patches and their inset reference colors in \Cref{fig:qualitative_mcc}, and the average $\Delta E_{00}$ scores for each image.

\subsection{Performance across different illuminant properties}
We better analyze the behavior of the color correction methods under different illumination conditions by dividing the test illuminants into separate classes according to two properties: their chromaticity distance from the canonical illuminant D65 in the CIE~xy space, and the area under their normalized SPD curve (AUC), which characterizes the spectral bandwidth of the illuminant. For each property, illuminants are sorted and divided into three equal-width classes. For the chromaticity distance, the three classes 
correspond to illuminants that are close, mid, and far from D65. For the AUC, the three classes correspond to broad-band, mid-band, and narrow-band illuminants, where a higher AUC indicates a broader SPD. AUC correlates with the type of illuminant, with broad-band SPDs being associated with natural and incandescent illuminants, while narrow-band SPDs can be associated with highly chromatic LEDs.

\Cref{tab:results_distancefromD65,tab:results_AUC} report 
$\Delta E_{00}$ and AE for each class on the 41M dataset. Results on KAUST and MCC are reported in the supplementary material. As expected, errors increase consistently with the distance from D65 for all methods, confirming that illuminants far from the target one represent a more challenging correction scenario. C$^2$LUT consistently outperforms all competing methods across all three distance classes, with improvements over the best competitor ranging from 9\% to 12\% in both $\Delta E_{00}$ and AE. Notably, the relative improvement of C$^2$LUT over competing methods is larger for the Mid and Far classes compared to the Close class, suggesting that the proposed framework is particularly effective under illuminants whose chromaticity significantly deviates from D65. A similar trend is observed when illuminants are categorized by their spectral bandwidth. Errors consistently increase as illuminants become more narrow-band, confirming that spectrally complex light sources with peaky SPDs are inherently harder to correct. Once again, C$^2$LUT outperforms all competitors across all three bandwidth classes, with improvements over the best 
competitor of 10--14\% in $\Delta E_{00}$ and 9--12\% in AE. The largest improvements are observed for mid-band and narrow-band illuminants, demonstrating the ability of the proposed framework to 
handle spectrally complex light sources whose narrow-band and peaky SPDs represent a more challenging correction scenario 
compared to smooth broad-band illuminants.

\begin{table*}[t]
    \centering
    \caption{Quantitative comparison of color correction methods under illuminant estimation errors. True illuminants are perturbed according to an inverse Gaussian distribution fitted to FC$^4$ based illuminant estimation error statistics. Results are reported on the 41M, KAUST and MCC datasets for three camera sensors (Canon~EOS~40D, Sony~Alpha~6100, Huawei~Mate~20~Pro). Metrics include CIE~$\Delta E_{00}$ and angular error ($^\circ$), including mean, standard deviation, and percentile statistics (25th, 50th, 95th). All methods are evaluated under noisy illuminant conditions. Best results are highlighted in \textbf{bold}.}
    \label{tab:results_illuminantestimation}
    \begin{adjustbox}{width=\textwidth}
    \begin{NiceTabular}{ccccccccccccccccc}
         \toprule
         \Block{3-1}{Dataset} & \Block{3-1}{Method} & \multicolumn{15}{c}{$\Delta E_{00} \downarrow$}\\\cmidrule{3-17}
         & & \multicolumn{5}{c}{Canon} & \multicolumn{5}{c}{Sony} & \multicolumn{5}{c}{Huawei}\\\cmidrule(lr){3-7}\cmidrule(lr){8-12}\cmidrule(lr){13-17}
         & &  Mean & Std & 25th & 50th & 95th & Mean & Std & 25th & 50th & 95th & Mean & Std & 25th & 50th & 95th \\\midrule

         \Block{5-1}{41M} & CCM-CCT & 10.68 & 4.95 & 6.41 & 9.75 & 20.12 & 10.56 & 4.22 & 6.82 & 10.01 & 18.12 & 10.66 & 4.16 & 7.11 & 10.34 & 18.30 \\
         & CST-MLP & 8.11 & 2.92 & 5.41 & 7.94 & 12.79 & 8.30 & 2.72 & 5.84 & 8.25 & 12.77 & 8.37 & 2.96 & 5.75 & 8.16 & 13.44\\
         & RP-MLP & 7.80 & 2.84 & 5.21 & 7.58 & 12.55 & 8.01 & 2.67 & 5.66 & 7.88 & 12.43& 8.11 & 2.91 & 5.50 & 7.75 & 13.00 \\
         & CCCNN & 7.61 & 2.75 & 5.11 & 7.37 & 12.10 & 7.77 & 2.60 & 5.39 & 7.50 & 12.06 & 7.92 & 2.79 & 5.44 & 7.53 & 12.77\\
         & C$^2$LUT (Ours) & \textbf{7.11} & \textbf{2.69} & \textbf{4.64} & \textbf{6.93} & \textbf{11.56} & \textbf{7.29} & \textbf{2.55} & \textbf{4.97} & \textbf{6.89} & \textbf{11.67} & \textbf{7.40} & \textbf{2.72} & \textbf{5.06} & \textbf{7.11} & \textbf{12.09}\\
         \cmidrule(lr){1-17}

         \Block{5-1}{KAUST} & CCM-CCT & 10.11 & 4.52 & 6.37 & 9.18 & 18.80 & 9.96 & 3.81 & 6.58 & 9.34 & 17.22 & 10.28 & 3.94 & 6.97 & 9.85 & 17.37\\
         & CST-MLP & 7.85 & 2.65 & 5.43 & 7.65 & 12.35 & 8.00 & 2.50 & 5.82 & 7.72 & 12.38 & 8.10 & 2.74 & 5.88 & 7.88 & 12.77\\
         & RP-MLP & 7.46 & 2.60 & 5.14 & 7.24 & 11.97 & 7.63 & 2.45 & 5.45 & 7.29 & 11.97 & 7.72 & 2.69 & 5.44 & 7.37 & 12.24\\
         & CCCNN & 7.27 & 2.51 & 5.09 & 6.79 & 11.43 & 7.31 & 2.36 & 5.28 & 6.93 & 11.48 & 7.58 & 2.65 & 5.37 & 7.16 & 12.33\\
         & C$^2$LUT (Ours) & \textbf{6.97} & \textbf{2.42} & \textbf{4.73} & \textbf{6.58} & \textbf{11.09} & \textbf{7.09} & \textbf{2.28} & \textbf{5.09} & \textbf{6.68} & \textbf{10.92} & \textbf{7.28} & \textbf{2.55} & \textbf{5.17} & \textbf{6.95} & \textbf{11.67}\\
         \cmidrule(lr){1-17}

         \Block{5-1}{MCC} & CCM-CCT & 11.45 & 6.86 & 6.07 & 9.69 & 25.74 & 11.90 & 7.35 & 6.15 & 10.11 & 26.65 & 12.13 & 7.19 & 6.65 & 10.61 & 25.96 \\
         & CST-MLP & 9.77 & 4.77 & 5.84 & 9.20 & 17.65 & 9.92 & 4.55 & 5.95 & 9.48 & 17.85 & 10.01 & 4.95 & 5.69 & 9.33 & 18.28\\
         & RP-MLP & 9.31 & 4.63 & 5.41 & 8.69 & 17.28 & 9.64 & 4.56 & 5.79 & 9.10 & 17.62 & 9.76 & 4.95 & 5.60 & 9.16 & 17.93 \\
         & CCCNN & 9.04 & \textbf{4.56} & 5.17 & 8.12 & 17.23 & 9.10 & \textbf{4.46} & 5.36 & 8.19 & \textbf{16.85} & 9.39 & \textbf{4.88} & 5.26 & 8.48 & 17.92\\
         & C$^2$LUT (Ours) & \textbf{8.27} & 4.77 & \textbf{4.21} & \textbf{7.13} & \textbf{16.98} & \textbf{8.44} & 4.63 & \textbf{4.50} & \textbf{7.35} & 16.88 & \textbf{8.67} & 5.03 & \textbf{4.35} & \textbf{7.46} & \textbf{17.06}\\

         \midrule
         & & \multicolumn{15}{c}{Angular Error $(^\circ)\downarrow$}\\\cmidrule{3-17}

         \Block{5-1}{41M} & CCM-CCT &  6.30 & 2.76 & 4.04 & 5.94 & 11.50 & 6.27 & 2.52 & 4.20 & 5.90 & 10.91 & 6.76 & 2.68 & 4.59 & 6.62 & 11.52\\
         & CST-MLP & 4.91 & 1.98 & 3.22 & 4.83 & 8.42 & 5.11 & 1.91 & 3.55 & 5.00 & 8.49 & 5.23 & 2.10 & 3.64 & 4.92 & 9.04\\
         & RP-MLP & 4.78 & 1.94 & 3.18 & 4.67 & 8.38 & 4.99 & 1.86 & 3.48 & 4.90 & 8.34 & 5.11 & 2.06 & 3.55 & 4.82 & 8.84 \\
         & CCCNN & 4.70 & 1.87 & 3.11 & 4.57 & 8.02 & 4.87 & 1.77 & 3.49 & 4.75 & 8.05 & 5.04 & 1.94 & 3.54 & 4.83 & 8.52\\
         & C$^2$LUT (Ours) & \textbf{4.44} & \textbf{1.82} & \textbf{2.87} & \textbf{4.25} & \textbf{7.66} & \textbf{4.63} & \textbf{1.75} & \textbf{3.13} & \textbf{4.45} & \textbf{7.76} & \textbf{4.73} & \textbf{1.90} & \textbf{3.24} & \textbf{4.46} & \textbf{8.23}\\
         \cmidrule(lr){1-17}

         \Block{5-1}{KAUST} & CCM-CCT & 6.05 & 2.66 & 3.82 & 5.61 & 11.27 & 5.93 & 2.37 & 4.01 & 5.45 & 10.26 & 6.43 & 2.51 & 4.44 & 6.11 & 10.76\\
         & CST-MLP & 4.89 & 1.91 & 3.30 & 4.63 & 8.18 & 4.99 & 1.84 & 3.51 & 4.77 & 8.28 & 5.08 & 1.95 & 3.63 & 4.85 & 8.73\\
         & RP-MLP & 4.74 & 1.89 & 3.20 & 4.53 & 8.17 & 4.86 & 1.79 & 3.40 & 4.65 & 8.08 & 4.96 & 1.93 & 3.51 & 4.73 & 8.52\\
         & CCCNN & 4.64 & 1.83 & 3.17 & 4.36 & 8.04 & 4.73 & 1.71 & 3.34 & 4.49 & 7.98 & 4.87 & 1.78 & 3.46 & 4.57 & 8.23\\
         & C$^2$LUT (Ours) & \textbf{4.53} & \textbf{1.73} & \textbf{3.08} & \textbf{4.32} & \textbf{7.63} & \textbf{4.62} & \textbf{1.62} & \textbf{3.31} & \textbf{4.35} & \textbf{7.70} & \textbf{4.74} & \textbf{1.76} & \textbf{3.40} & \textbf{4.42} & \textbf{8.08}\\
         \cmidrule(lr){1-17}

         \Block{5-1}{MCC} & CCM-CCT & 6.72 & 3.95 & 3.73 & 6.01 & 14.29 & 7.17 & 4.26 & 3.83 & 6.56 & 15.26 & 7.62 & 4.45 & 4.31 & 6.92 & 16.19\\
         & CST-MLP & 5.87 & 3.09 & 3.24 & 5.77 & 11.36 & 6.14 & 3.02 & 3.52 & 5.91 & 11.57 & 6.36 & 3.30 & 3.49 & 6.13 & 12.37\\
         & RP-MLP & 5.65 & 3.10 & 3.01 & 5.32 & 11.10 & 5.99 & 3.04 & 3.42 & 5.60 & 11.19 & 6.22 & 3.32 & 3.32 & 5.81 & 12.42\\
         & CCCNN & 5.41 & \textbf{3.04} & 2.84 & 4.96 & 10.84 & 5.65 & \textbf{2.96} & 3.17 & 5.31 & \textbf{11.09} & 5.99 & \textbf{3.28} & 3.24 & 5.43 & 12.10\\
         & C$^2$LUT (Ours) & \textbf{5.04} & 3.12 & \textbf{2.47} & \textbf{4.46} & \textbf{10.79} & \textbf{5.32} & 3.10 & \textbf{2.74} & \textbf{4.80} & 11.49 & \textbf{5.56} & 3.46 & \textbf{2.60} & \textbf{4.96} & \textbf{11.73}\\
         \bottomrule
         
    \end{NiceTabular}
    \end{adjustbox}
\end{table*}

\subsection{Robustness to illuminant estimation errors}

In this section we evaluate the robustness of the compared methods to illuminant estimation errors, which arise in scenarios where the reference white target cannot be placed in the scene and the illuminant must be estimated. Building upon the procedure described in \Cref{sec:setup}, we additionally simulate illuminant estimation errors by sampling angular error values from an inverse Gaussian distribution~\cite{schrodinger1915theorie} fitted to the error statistics of the FC$^4$ illuminant estimation method~\cite{hu2017fc4}, and perturbing the true illuminant white point so that the angular error between the original and perturbed white points matches the sampled value. Under this setting, all models are trained from scratch using the perturbed illuminants as input.

\Cref{tab:results_illuminantestimation} reports $\Delta E_{00}$~\cite{sharma2005ciede2000} and AE for all methods across three cameras and three reflectance datasets. As expected, errors are higher than the ones observed in \Cref{subsec:colorcorrectionknownilluminants}, reflecting the additional difficulty introduced by illuminant estimation errors. C$^2$LUT consistently achieves the lowest mean error across all datasets, cameras and metrics, with improvements 
over the best competing method ranging from 3\% to 9\% in $\Delta E_{00}$ and from 2\% to 7\% in AE. The improvements are most pronounced on MCC~\cite{mccamy1976color}, where the limited number of reflectances makes the correction more sensitive to illuminant errors. Regarding variability, on the 41M~\cite{zhang2016metamer} and KAUST~\cite{li2021multispectral} datasets, C$^2$LUT achieves the lowest standard deviation in all configurations, confirming its robustness to varying illumination conditions. In addition, C$^2$LUT achieves the best 95th percentile in most configurations, with exceptions on MCC where the results are comparable to CCCNN~\cite{macdonald2021camera}, demonstrating that the proposed framework maintains its advantage even under the most challenging combinations of illuminant estimation errors and reflectance diversity.

\subsection{Robustness to exposure variations}
\begin{table*}[t]
    \centering
    \caption{Quantitative evaluation of color correction robustness to exposure variations on the 41M dataset. Input raw responses and corresponding reference XYZ values are scaled by a factor $\alpha \in {0.75, 0.5, 0.25}$ to simulate progressively decreasing exposure levels. All methods are evaluated across three camera sensors (Canon~EOS~40D, Sony~Alpha~6100, Huawei~Mate~20~Pro). Results are reported using angular error ($^\circ$), including mean, standard deviation, and percentile statistics (25th, 50th, 95th). Best results are highlighted in \textbf{bold}.}
    \label{tab:exposure_41M}
    \begin{adjustbox}{width=\textwidth}
    \begin{NiceTabular}{ccccccccccccccccc}
         \toprule
         \Block{3-1}{Exposure} & \Block{3-1}{Method} & \multicolumn{15}{c}{AE $\downarrow$}\\\cmidrule{3-17}
         & & \multicolumn{5}{c}{Canon} & \multicolumn{5}{c}{Sony} & \multicolumn{5}{c}{Huawei}\\\cmidrule(lr){3-7}\cmidrule(lr){8-12}\cmidrule(lr){13-17}
         & &  Mean & Std & 25th & 50th & 95th & Mean & Std & 25th & 50th & 95th & Mean & Std & 25th & 50th & 95th \\\midrule

         \Block{5-1}{0.75} & CCM-CCT & 4.51 & 2.05 & 2.67 & 4.46 & 8.11 & 4.63 & 2.19 & 2.65 & 4.47 & 8.68 & 5.14 & 2.38 & 2.98 & 5.07 & 9.10\\
         & CST-MLP & 3.16 & 1.49 & 1.95 & 2.95 & 5.56 & 3.30 & 1.58 & 2.01 & 3.06 & 5.88 & 3.35 & 1.71 & 1.97 & 3.14 & 6.23\\
         & RP-MLP & 3.01 & 1.48 & 1.82 & 2.82 & 5.31 & 3.15 & 1.57 & 1.86 & 2.91 & 5.71 & 3.23 & 1.71 & 1.83 & 3.10 & 6.28\\
         & CCCNN & 2.99 & 1.37 & 1.90 & 2.73 & 5.30 & 3.14 & 1.46 & 1.98 & 2.93 & 5.66 & 3.19 & 1.59 & 1.92 & 2.89 & 6.12\\
         & C$^2$LUT (Ours) & \textbf{2.74} & \textbf{1.29} & \textbf{1.71} & \textbf{2.52} & \textbf{4.98} & \textbf{2.87} & \textbf{1.40} & \textbf{1.73} & \textbf{2.67} & \textbf{5.21} & \textbf{2.93} & \textbf{1.54} & \textbf{1.71} & \textbf{2.62} & \textbf{5.51}\\
         \cmidrule(lr){1-17}

         \Block{5-1}{0.5} & CCM-CCT & 4.51 & 2.05 & 2.67 & 4.46 & 8.11 & 4.63 & 2.19 & 2.65 & 4.47 & 8.68 & 5.14 & 2.38 & 2.98 & 5.07 & 9.10\\
         & CST-MLP & 3.16 & 1.49 & 1.95 & 2.95 & 5.56 & 3.30 & 1.58 & 2.01 & 3.06 & 5.88 & 3.35 & 1.71 & 1.97 & 3.14 & 6.23\\
         & RP-MLP & 3.03 & 1.49 & 1.86 & 2.89 & 5.35 & 3.17 & 1.58 & 1.85 & 2.97 & 5.77 & 3.25 & 1.72 & 1.85 & 3.11 & 6.31\\
         & CCCNN & 3.13 & 1.43 & 2.00 & 2.89 & 5.54 & 3.28 & 1.52 & 2.04 & 3.05 & 6.09 & 3.33 & 1.61 & 2.06 & 2.98 & 6.20\\
         & C$^2$LUT (Ours) & \textbf{2.89} & \textbf{1.33} & \textbf{1.82} & \textbf{2.73} & \textbf{5.08} & \textbf{3.02} & \textbf{1.44} & \textbf{1.83} & \textbf{2.84} & \textbf{5.29} & \textbf{3.07} & \textbf{1.58} & \textbf{1.83} & \textbf{2.78} & \textbf{5.79}\\
         \cmidrule(lr){1-17}

         \Block{5-1}{0.25} & CCM-CCT & 4.51 & 2.05 & 2.67 & 4.46 & 8.11 & 4.63 & 2.19 & 2.65 & 4.47 & 8.68 & 5.14 & 2.38 & 2.98 & 5.07 & 9.10\\
         & CST-MLP & 3.16 & 1.49 & 1.95 & 2.95 & 5.56 & 3.30 & 1.58 & 2.01 & 3.06 & 5.88 & 3.35 & 1.71 & 1.97 & 3.14 & \textbf{6.23}\\
         & RP-MLP & \textbf{3.05} & 1.49 & \textbf{1.87} & \textbf{2.92} & \textbf{5.38} & \textbf{3.19} & 1.58 & \textbf{1.85} & \textbf{2.98} & \textbf{5.80} & \textbf{3.27} & 1.72 & \textbf{1.86} & \textbf{3.10} & 6.32\\
         & CCCNN & 3.60 & 1.63 & 2.24 & 3.36 & 6.34 & 3.78 & 1.75 & 2.38 & 3.51 & 7.20 & 3.83 & 1.71 & 2.51 & 3.44 & 6.75\\
         & C$^2$LUT (Ours) & 3.39 & \textbf{1.44} & 2.16 & 3.25 & 5.84 & 3.51 & \textbf{1.56} & 2.19 & 3.30 & 6.15 & 3.58 & \textbf{1.67} & 2.24 & 3.34 & 6.39\\
         
         \bottomrule
         
    \end{NiceTabular}
    \end{adjustbox}
\end{table*}

We evaluate the sensitivity of the compared methods to exposure variations. In a real camera, the sensor response is proportional to the amount of light reaching it, so multiplying both the raw pixel values and the corresponding reference XYZ by a scalar factor $\alpha$ provides a physically consistent simulation of a proportional change in exposure time~\cite{cogo2026leveraging}. We consider $\alpha \in \{0.75, 0.5, 0.25\}$, simulating progressively shorter exposure times with respect to the reference exposure and resulting in increasingly underexposed images.

\Cref{tab:exposure_41M} reports AE
for all methods across three cameras at each exposure level on the 41M dataset~\cite{zhang2016metamer}. Results on KAUST and MCC are reported in the supplementary material. Note that $\Delta E_{00}$ is not informative in this setting, as darker images tend to produce lower color differences regardless of correction accuracy. 
Notably, linear methods such as CCM-CCT~\cite{dng} and CST-MLP~\cite{tedla2025off} are by design invariant to exposure in terms of angular error, since multiplying the input by a scalar does not affect the direction of a 
linear transformation. Similarly, RP-MLP~\cite{finlayson2015color} benefits from the exposure robustness of the root-polynomial expansion. In contrast, due to their non-linear nature, CCCNN~\cite{macdonald2021camera} and C$^2$LUT are not intrinsically invariant to exposure changes.

Nevertheless, at exposure levels 0.75 and 0.5, C$^2$LUT achieves the best AE performance across all cameras, with improvements over the best competing method of up to 9\%. As the exposure decreases to 0.25, corresponding to a severely underexposed image, RP-MLP outperforms all other methods, demonstrating the advantage of its exposure-robust polynomial formulation under extreme conditions. Despite this, C$^2$LUT remains competitive with RP-MLP and consistently surpasses CCCNN across all exposure levels, suggesting that its structured LUT representation offers greater robustness to exposure changes than a direct pixel-wise MLP. 

\subsection{Generalization to real data}
\begin{figure}
    \centering
    \includegraphics[width=1\linewidth]{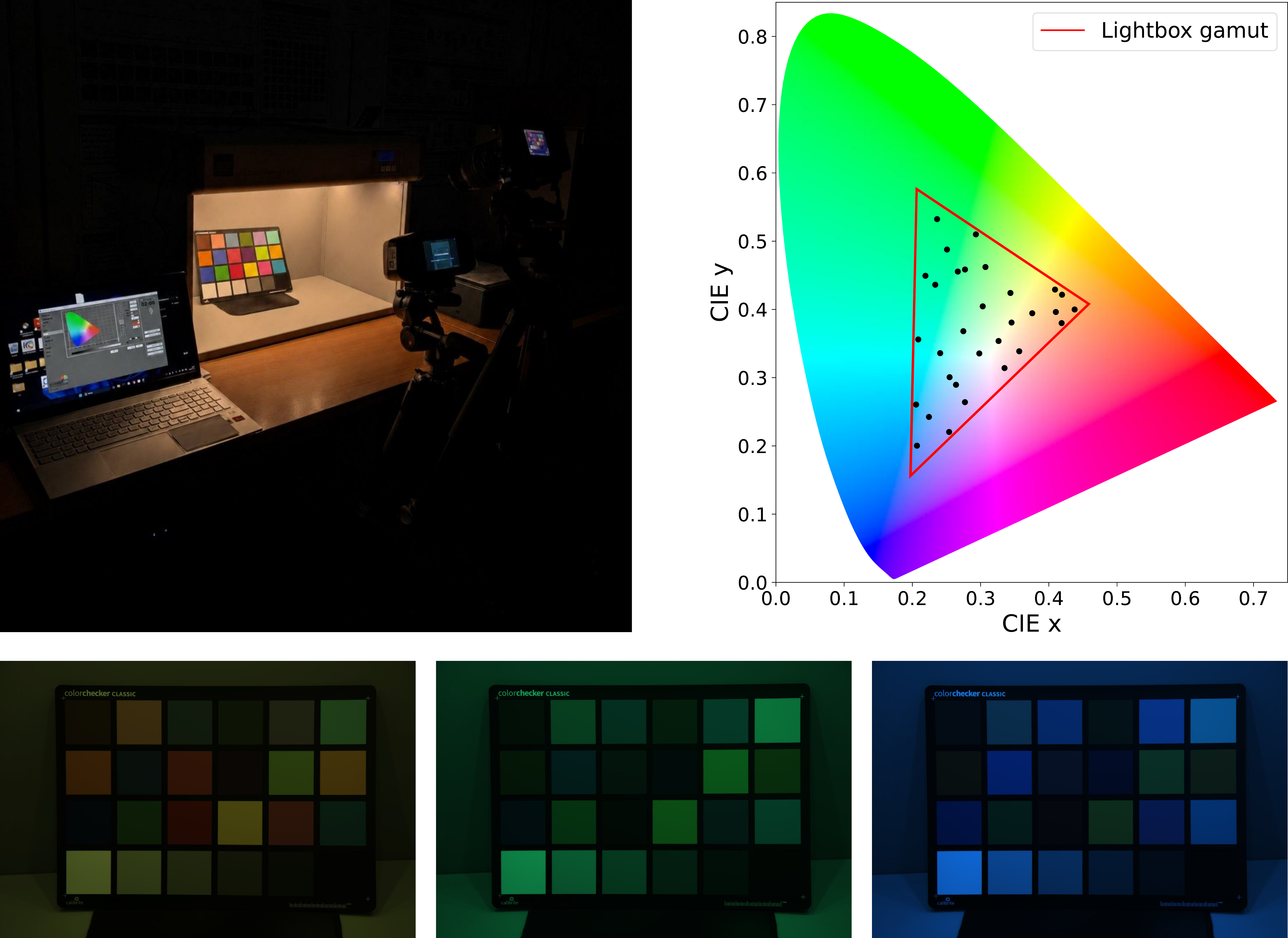}
    \caption{Overview of the real captures benchmark dataset. \textit{Top-left}: our acquisition setup. \textit{Top-right}: CIE xy-chromaticity distribution of the illuminants. The highlighted triangle delimits the chromaticity gamut reproducible by the lightbox. \textit{Bottom}: Examples from the dataset of the captured Macbeth Color Checker under multiple illuminants.
    \vspace{-4mm}
    }
    \label{fig:real_benckmark}
\end{figure}

\begin{table}[t]
    \centering
    \caption{Quantitative comparison of color correction methods on real-world images of a Macbeth Color Checker (MCC) captured with a Canon~EOS~40D under 31 LED illuminants reproduced by a programmable lightbox. Performance is reported in terms of CIE~$\Delta E_{00}$ and angular error ($^\circ$), including mean, standard deviation, and percentile statistics (25th, 50th, 95th). Best results are highlighted in \textbf{bold}.}
    \label{tab:real}
    \begin{NiceTabular}{cccccc}
         \toprule
         \Block{2-1}{Method} & \Block{1-5}{$\Delta E_{00} \downarrow$}\\\cmidrule(lr){2-6}
         & Mean & Std & 25th & 50th & 95th\\\midrule
        CCM-CCT & 5.10 & 1.91 & 3.65 & 4.44 & 8.86\\
         CST-MLP & 4.85 & 2.42 & 3.33 & 4.16 & 9.83\\
         RP-MLP & 4.65 & 2.43 & 3.14 & 3.96 & 9.86\\
         CCCNN & 4.23 & \textbf{1.53} & 3.33 & 3.74 & \textbf{7.03}\\
         C$^2$LUT (Ours) & \textbf{3.97} & 1.63 & \textbf{3.01} & \textbf{3.33} & 7.28\\
         \midrule
         & \Block{1-5}{Angular Error $(^\circ)\downarrow$}\\\cmidrule(lr){2-6}
         CCM-CCT & 2.89 & 1.65 & 1.70 & 2.26 & 6.09\\
         CST-MLP & 2.12 & 1.19 & 1.27 & 1.82 & 4.58\\
         RP-MLP & 2.04 & 1.22 & 1.22 & 1.68 & 4.68\\
         CCCNN & 2.01 & \textbf{0.84} & 1.44 & 1.74 & \textbf{3.36}\\
         C$^2$LUT (Ours) & \textbf{1.84} & 0.98 & \textbf{1.12} & \textbf{1.60} & 3.91\\
         \bottomrule
    \end{NiceTabular}
    \vspace{-3mm}
\end{table}

We validate the generalization of the proposed framework from synthetic to real data by capturing real images with a Canon~EOS~40D camera under different illuminants. We use a programmable LED lightbox (JUST-Normlicht LED Color Viewing Light) to illuminate an MCC~\cite{mccamy1976color} placed inside it, allowing precise control of the illuminant chromaticity by programming the desired CIE~xy coordinates. The adopted setup is illustrated in \Cref{fig:real_benckmark} (\textit{top-left}). We acquire images of the MCC under 31 LED illuminants, whose chromaticities are shown in \Cref{fig:real_benckmark} (\textit{top-right}). For each acquisition, the illuminant white point is extracted from the white patch of the MCC captured by the camera. The reference XYZ values for the MCC patches are computed from their reflectance spectra measured with a spectrophotometer, combined with the illuminant SPDs measured from the lightbox and the CIE~XYZ~\cite{CIE19} CMFs via \Cref{eq:xyz_target}. Additional visualizations and illuminant SPDs are provided in the supplementary material.

\Cref{tab:real} reports the results for all methods on the real acquisition dataset. Note that, despite being trained exclusively on synthetic data, all methods achieve results consistent with those obtained using the synthetic setup. This suggests that results on synthetic data are a reliable proxy for real-world acquisition conditions. C$^2$LUT achieves the best mean performance in both $\Delta E_{00}$~\cite{sharma2005ciede2000} and AE, 
demonstrating effective generalization from synthetic to real images, with improvements of 6.1\% in $\Delta E_{00}$ and 8.5\% in AE over the best competing method. C$^2$LUT also achieves the best 25th percentile and median in both metrics, confirming that the advantage reflects a systematic improvement. Although CCCNN~\cite{macdonald2021camera} achieves lower standard deviation and 95th percentile than C$^2$LUT in both metrics, C$^2$LUT still outperforms all remaining methods according to these statistics.

\subsection{Computational complexity}

\begin{table}
    \centering
    \caption{Comparison of computational complexity. GFLOPs refer to the complete inference pipeline, from raw input to XYZ. For C$^2$LUT, the memory footprint indicates the requirements to store both model parameters and reconstructed LUT.}
    \label{tab:complexity}
    \begin{NiceTabular}{cccccc}
         \toprule
         \Block{2-1}{Method} & \Block{2-1}{Memory} & \multicolumn{4}{c}{GFLOPs}\\\cmidrule{3-6}
         & & 40 MP & 24 MP & 10 MP\\\midrule
         CCM-CCT & 0.07 KB & 0.60 & 0.36 & 0.15\\
         CST-MLP & 5.80 KB & 1.24 & 0.74 & 0.31\\
         RP-MLP & 7.00 KB & 1.96 & 1.18 & 0.49\\
         CCCNN & 5.40 KB & 52.10 & 31.32 & 13.15\\
         C$^2$LUT (Ours) & 3.96 MB & 1.34 & 0.81 & 0.35\\
         \bottomrule
    \end{NiceTabular}
    \vspace{-4mm}
    
\end{table}

We compare the computational complexity of all methods in terms of memory requirements and inference cost, measured in GFLOPs. We evaluate it at three image resolutions corresponding to the native resolutions of the cameras used in our experiments: 40 MP (7296$\times$5472) for the Huawei~Mate~20~Pro, 24 MP (6000$\times$4000) for the Sony~Alpha~6100, and 10 MP (3888$\times$2592) for the Canon~EOS~40D. 

\Cref{tab:complexity} reports the results.
In terms of memory requirements, CCM-CCT~\cite{dng}, CST-MLP~\cite{tedla2025off}, RP-MLP~\cite{finlayson2015color} and CCCNN~\cite{macdonald2021camera} all require less than 8 KB to store their parameters, making them extremely lightweight. C$^2$LUT requires 726.96 KB for the model parameters, plus an additional 3.30 MB to store the constructed LUT at inference time. While this is larger than the other methods, it remains well within the memory constraints of modern camera pipelines. In terms of inference cost, C$^2$LUT is remarkably efficient despite its higher memory footprint. At 24 MP, C$^2$LUT requires only 0.81 GFLOPs, comparable to CST-MLP (0.74 GFLOPs) and significantly lower than RP-MLP (1.18 GFLOPs) and CCCNN (31.32 GFLOPs). At the highest resolution of 40 MP, C$^2$LUT requires 1.34 GFLOPs, while CCCNN reaches 52.10 GFLOPs, making it approximately $39\times$ more expensive. 

C$^2$LUT has low inference cost because the illuminant-dependent LUT is generated only once per scene. Subsequent pixel-wise color correction is then reduced to a standard 3D LUT lookup with trilinear interpolation, which is efficient and compatible with conventional ISP pipelines. As a result, the method is highly scalable to high-resolution images.

\vspace{-1mm}
\subsection{Ablation study}
\label{sec:ablations}

We analyze the impact of the illuminant representation~(IR) and the main hyperparameters of C$^2$LUT: the number of basis cores $B$, the Tucker rank $R$, and the LUT resolution $N$. \Cref{tab:ablation} reports mean values of $\Delta E_{00}$~\cite{sharma2005ciede2000} and AE on the 41M dataset~\cite{zhang2016metamer} for all three cameras, alongside the memory footprint of each configuration.
Our default configuration corresponds to $\text{IR}=rg$, $B=12$, $R=17$, and $N=65$.

Replacing illuminant chromaticity with CCT for IR leads to a substantial performance degradation, with $\Delta E_{00}$ increasing by approximately 25\% and AE by 30–38\%. This confirms that the chromaticity-based representation captures illuminant-dependent color variations considerably better than CCT, as CCT collapses the illuminant to a one-dimensional descriptor along the Planckian locus~\cite{tedla2025off}.

The number of basis cores $B$ has by far the largest impact on performance. Reducing $B$ from 12 to 1 leads to a degradation of approximately 49--51\% in $\Delta E_{00}$ and 52--55\% in AE across all cameras, confirming that multiple basis cores are essential for the framework to adapt the color correction to different illumination conditions. With a single core, the model cannot modulate the LUT based on the illuminant chromaticity, translating into a fixed color correction that is unable to account for illuminant variability.

The Tucker rank $R$ and the LUT resolution $N$ have a much smaller impact on performance. Reducing $R$ from 17 to 11 leads to a degradation of less than 2\% in both metrics, while reducing $N$ from 65 to 33 leads to a similarly small degradation of less than 2\%, but with a significant reduction in memory from 4.02 MB to 1.15 MB. These results suggest that $N=33$ and $R=11$ represent valid alternatives when memory is a constraint, at the cost of a marginal reduction in accuracy.

Finally, using full-resolution LUT cores without Tucker tensor compression produces a slight improvement of 1--3\% in both metrics compared to the default $R=17$, at the cost of a substantially larger memory footprint of 42.85 MB, which is approximately $10.7\times$ larger than the default configuration. In this configuration, each of the $B=12$ basis cores is itself a full-resolution LUT of shape $3 \times 65 \times 65 \times 65$.
        

\begin{table}
    \centering
    \caption{Hyperparameter analysis of C$^2$LUT on the 41M dataset. IR denotes the illuminant representation, $B$ is the number of basis cores, $R$ is the Tucker rank, and $N$ is the LUT resolution. Mean $\Delta E_{00}$ and mean angular error (AE) are reported for each camera. The first row corresponds to our default configuration. The last row corresponds to full-resolution LUT without Tucker tensor compression ($R=N$).}
    \label{tab:ablation}
    \begin{adjustbox}{width=\columnwidth}
        
    \begin{NiceTabular}{ccccccccccc}
    \toprule
    \Block{2-1}{IR} & \Block{2-1}{B} & \Block{2-1}{R} & \Block{2-1}{N} & \Block{2-1}{Memory} & \multicolumn{2}{c}{Canon} & \multicolumn{2}{c}{Sony} & \multicolumn{2}{c}{Huawei}\\\cmidrule(lr){6-7}\cmidrule(lr){8-9}\cmidrule(lr){10-11}
    & & & & & $\Delta E_{00} \downarrow$ & AE $\downarrow$ & $\Delta E_{00} \downarrow$ & AE $\downarrow$ & $\Delta E_{00} \downarrow$ & AE $\downarrow$\\\midrule
    rg & 12 & 17 & 65 & 4.02 MB & 4.32 & 2.64 & 4.54 & 2.78 & 4.72 & 2.85\\\midrule
    CCT & 12 & 17 & 65 & 4.02 MB & 5.44 & 3.48 & 5.66 & 3.61 & 5.88 & 3.93\\
    rg & 1 & 17 & 65 & 3.37 MB & 6.45 & 4.02 & 6.59 & 4.18 & 6.73 & 4.41\\
    rg & 12 & 11 & 65 & 3.50 MB & 4.40 & 2.68 & 4.59 & 2.80 & 4.77 & 2.86\\
    rg & 12 & 17 & 33 & 1.15 MB& 4.36 & 2.67 & 4.60 &2.83 & 4.79 & 2.90\\
    rg & 12 & 65 & 65 & 42.85 MB & 4.25 & 2.59 & 4.43 & 2.70 & 4.64 & 2.78\\
    \toprule
    \end{NiceTabular}
    \end{adjustbox}
    \vspace{-5mm}
\end{table}
\vspace{-1mm}
\section{Conclusion}

This paper presented C$^2$LUT, an illuminant-adaptive framework for in-camera color correction based on 3D~LUTs. By combining a chromaticity-aware illuminant representation with a non-linear LUT-based color transformation, the proposed C$^2$LUT addresses the limitations of conventional color correction methods under spectrally complex and highly chromatic lighting conditions. We parametrize the LUTs using Tucker tensor decomposition, significantly reducing computational requirements while preserving correction accuracy, enabling practical deployment within camera ISP pipelines. We introduce a large-scale dataset comprising 1,473 SPDs spanning a wide range of illuminant chromaticities and spectral profiles, together with a real-world benchmark acquired under controlled illumination conditions. 

Extensive experiments across multiple cameras, reflectance datasets, and illuminants demonstrate that the proposed C$^2$LUT consistently outperforms existing methods, with particularly large improvements under challenging LED lighting conditions. Further analyses confirm that these gains remain stable under illuminant estimation errors and moderate exposure variations. 
Finally, experiments on a real captured benchmark dataset suggest that the observed performance improvements transfer reliably to real-world images.

Future work will focus on extending C$^2$LUT toward cross-camera generalization, enabling the framework to generalize across different sensors without re-training or finetuning.

\printbibliography

\clearpage
\clearpage

\twocolumn[
\begin{center}
    {\Huge \papertitle\par}
    \vspace{0.7em}
    {\LARGE Supplementary Material\par}
    \vspace{1.5em}
\end{center}
]

 
\renewcommand{\thefigure}{S\arabic{figure}}
\renewcommand{\thetable}{S\arabic{table}}
\renewcommand{\theequation}{S\arabic{equation}}
\setcounter{figure}{0}
\setcounter{table}{0}
\setcounter{equation}{0}
 
\section{Additional experiments}

\subsection{Real illuminant spectral power distributions}
\begin{figure}[t]
    \centering
    \includegraphics[width=\columnwidth]{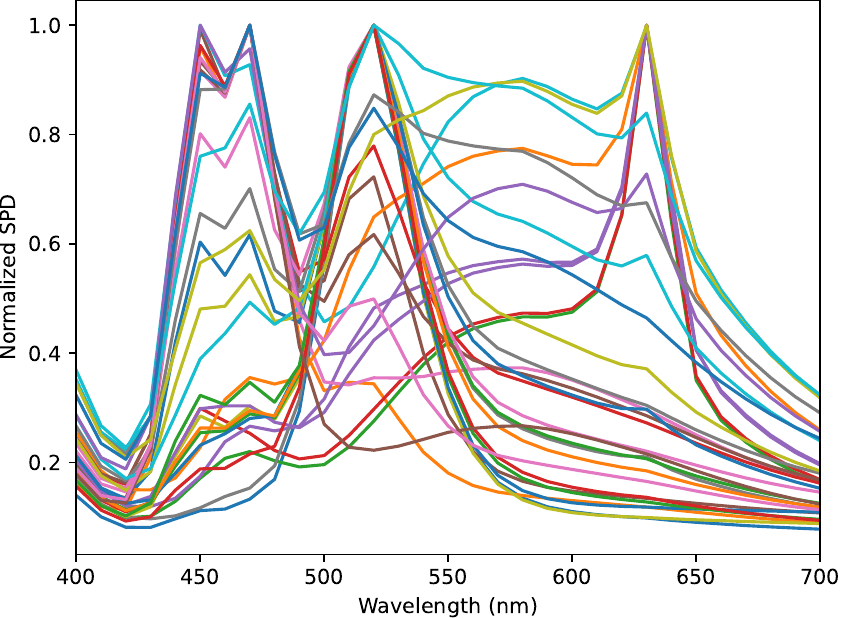}
    \caption{Visualization of the lightbox illuminant SPDs used for real data acquisition. SPDs are shown from 400 to 700 nm with a step of 10 nm.}
    \label{fig:spds_real}
\end{figure}

\Cref{fig:spds_real} shows the normalized spectral power distributions (SPDs) of the 31 illuminants used for the real acquisition of the Macbeth Color Checker (MCC)~\cite{mccamy1976color}, measured with the Specim IQ camera from the white patch of the color checker placed inside the scene. 

\subsection{Color correction results}

\begin{figure*}[t]
    \centering
    \setlength{\tabcolsep}{1pt}
    \begin{tabular}{ccccccccc}
    
    {\small Illuminant} &
    {\small RAW Input} &
    &
    {\small CCM-CCT} &
    {\small CST-MLP} &
    {\small RP-MLP} &
    {\small CCCNN} &
    {\small C$^2$LUT (Ours)} &
    {\small Reference} \\
    
    \includegraphics[width=0.15\textwidth]{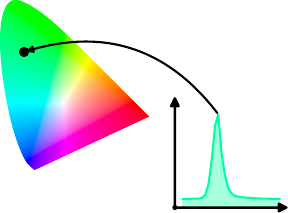} &
    \includegraphics[width=0.11\textwidth]{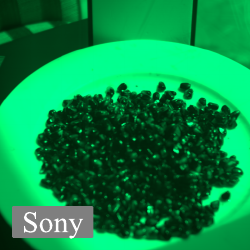} &
    \includegraphics[width=0.0089\textwidth]{Figures/KAUST/cbar.pdf} &
    \includegraphics[width=0.11\textwidth]{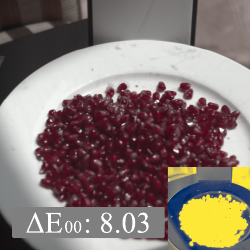} &
    \includegraphics[width=0.11\textwidth]{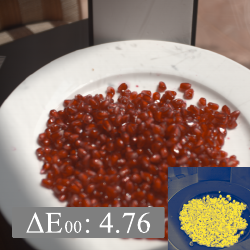} &
    \includegraphics[width=0.11\textwidth]{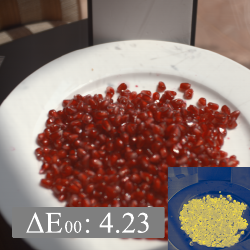} &
    \includegraphics[width=0.11\textwidth]{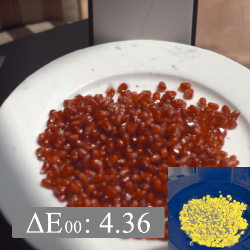} &
    \includegraphics[width=0.11\textwidth]{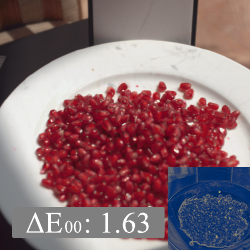} &
    \includegraphics[width=0.11\textwidth]{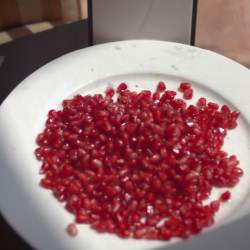}
    \\
    \includegraphics[width=0.15\textwidth]{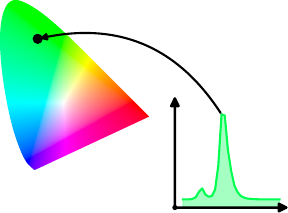} &
    \includegraphics[width=0.11\textwidth]{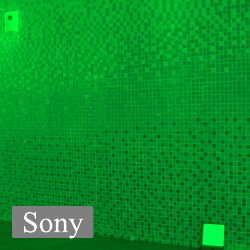} &
    \includegraphics[width=0.0089\textwidth]{Figures/KAUST/cbar.pdf} &
    \includegraphics[width=0.11\textwidth]{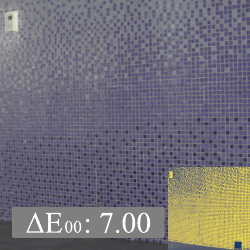} &
    \includegraphics[width=0.11\textwidth]{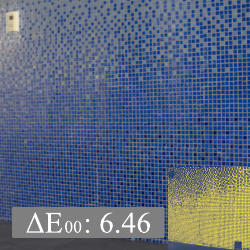} &
    \includegraphics[width=0.11\textwidth]{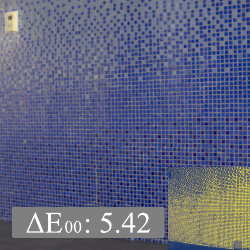} &
    \includegraphics[width=0.11\textwidth]{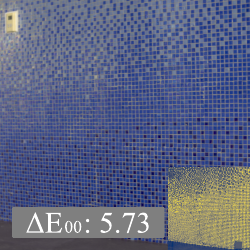} &
    \includegraphics[width=0.11\textwidth]{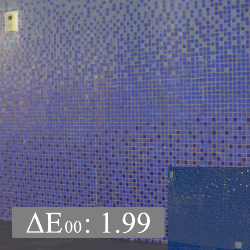} &
    \includegraphics[width=0.11\textwidth]{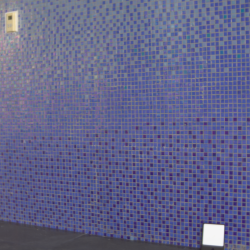}
    \\
    \includegraphics[width=0.15\textwidth]{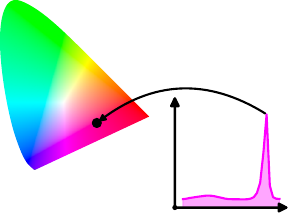} &
    \includegraphics[width=0.11\textwidth]{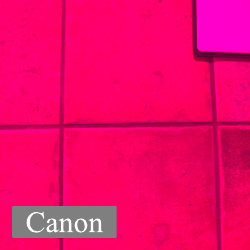} &
    \includegraphics[width=0.0089\textwidth]{Figures/KAUST/cbar.pdf} &
    \includegraphics[width=0.11\textwidth]{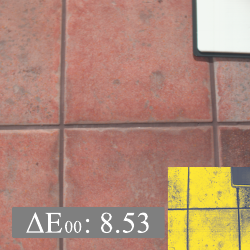} &
    \includegraphics[width=0.11\textwidth]{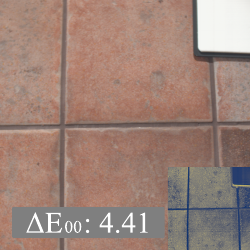} &
    \includegraphics[width=0.11\textwidth]{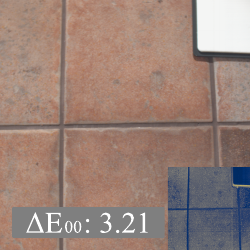} &
    \includegraphics[width=0.11\textwidth]{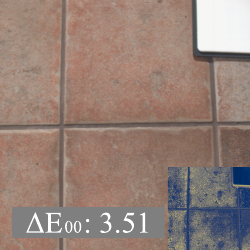} &
    \includegraphics[width=0.11\textwidth]{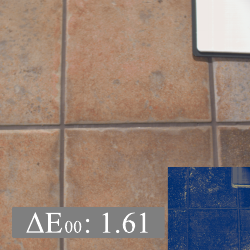} &
    \includegraphics[width=0.11\textwidth]{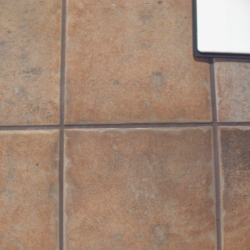}
    \\
    \includegraphics[width=0.15\textwidth]{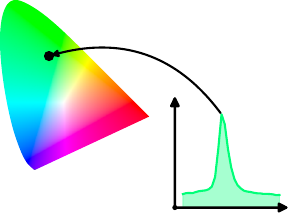} &
    \includegraphics[width=0.11\textwidth]{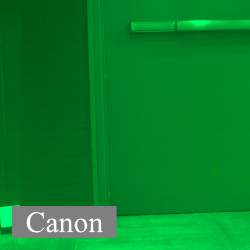} &
    \includegraphics[width=0.0089\textwidth]{Figures/KAUST/cbar.pdf} &
    \includegraphics[width=0.11\textwidth]{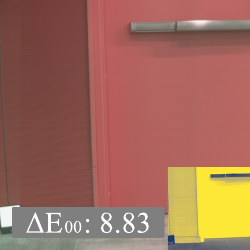} &
    \includegraphics[width=0.11\textwidth]{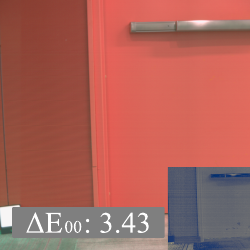} &
    \includegraphics[width=0.11\textwidth]{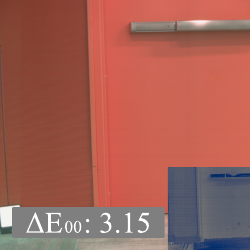} &
    \includegraphics[width=0.11\textwidth]{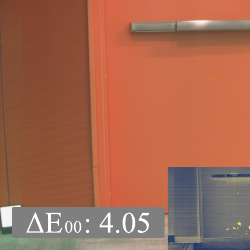} &
    \includegraphics[width=0.11\textwidth]{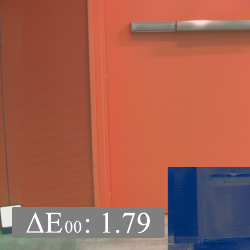} &
    \includegraphics[width=0.11\textwidth]{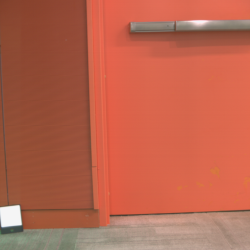}
    \\
    \includegraphics[width=0.15\textwidth]{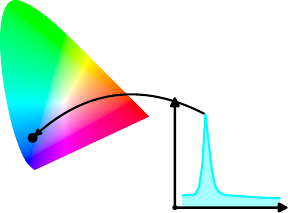} &
    \includegraphics[width=0.11\textwidth]{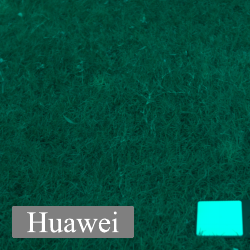} &
    \includegraphics[width=0.0089\textwidth]{Figures/KAUST/cbar.pdf} &
    \includegraphics[width=0.11\textwidth]{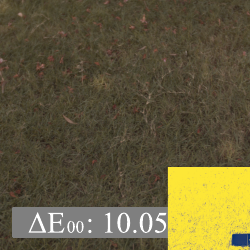} &
    \includegraphics[width=0.11\textwidth]{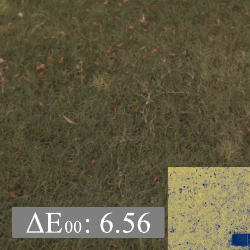} &
    \includegraphics[width=0.11\textwidth]{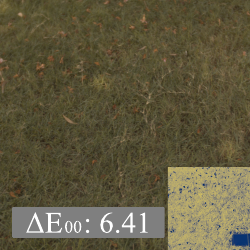} &
    \includegraphics[width=0.11\textwidth]{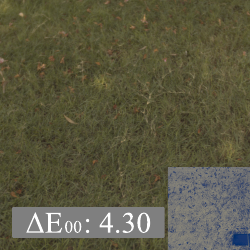} &
    \includegraphics[width=0.11\textwidth]{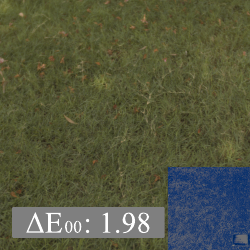} &
    \includegraphics[width=0.11\textwidth]{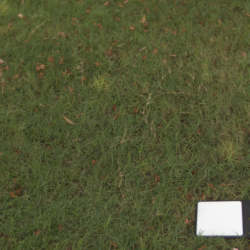}
    \\
    \includegraphics[width=0.15\textwidth]{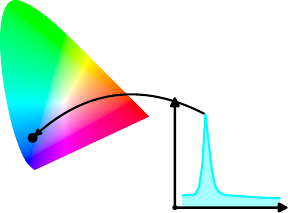} &
    \includegraphics[width=0.11\textwidth]{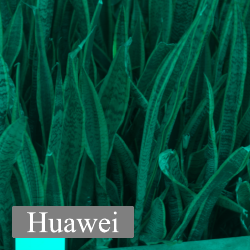} &
    \includegraphics[width=0.0089\textwidth]{Figures/KAUST/cbar.pdf} &
    \includegraphics[width=0.11\textwidth]{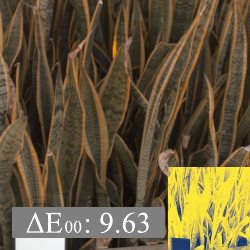} &
    \includegraphics[width=0.11\textwidth]{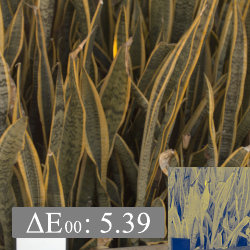} &
    \includegraphics[width=0.11\textwidth]{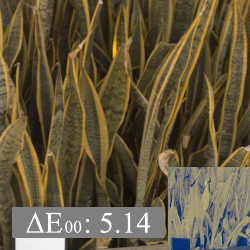} &
    \includegraphics[width=0.11\textwidth]{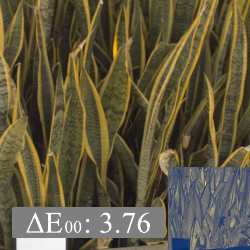} &
    \includegraphics[width=0.11\textwidth]{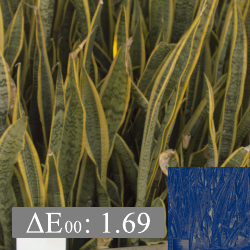} &
    \includegraphics[width=0.11\textwidth]{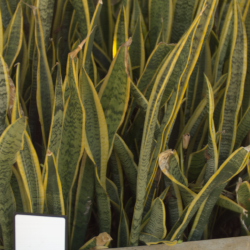}
    \end{tabular}
    \caption{Additional qualitative comparison of color correction results under challenging test illuminants. Results are shown for the compared methods on images rendered using three different camera SSFs. Each corrected image is associated with a per-pixel $\Delta E_{00}$ map in range [0-10], which helps visualizing the spatial distribution of color correction errors across the image.
    The proposed C$^2$LUT framework produces colors that are visually closer to the reference and achieves the lowest CIE~$\Delta E_{00}$ across all camera models.}
    \label{fig:suppl_qualitative_kaust}
\end{figure*}

\begin{figure*}[t]
    \centering
    \setlength{\tabcolsep}{1pt}
    \begin{tabular}{ccccccc}
    
    {\small Illuminant} &
    {\small RAW Input} &
    {\small CCM-CCT} &
    {\small CST-MLP} &
    {\small RP-MLP} &
    {\small CCCNN} &
    {\small C$^2$LUT (Ours)} \vspace{-4mm}\\

    \includegraphics[width=0.15\textwidth]{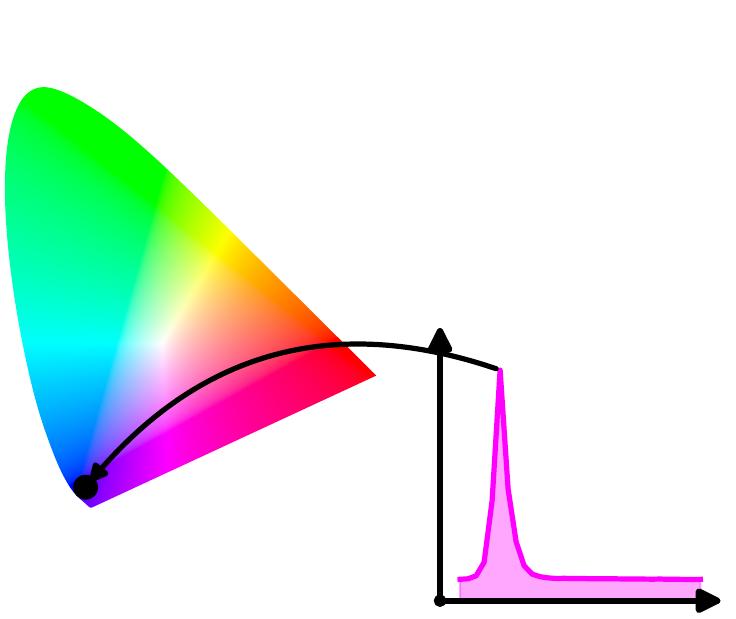} &
    \includegraphics[width=0.13\textwidth]{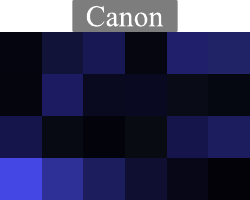} &
    \includegraphics[width=0.13\textwidth]{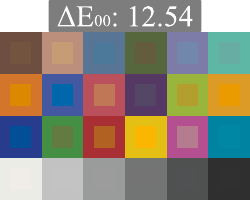} &
    \includegraphics[width=0.13\textwidth]{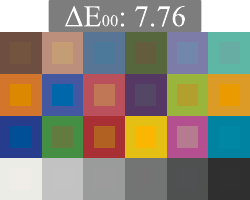} &
    \includegraphics[width=0.13\textwidth]{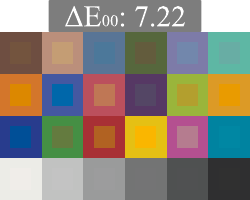} &
    \includegraphics[width=0.13\textwidth]{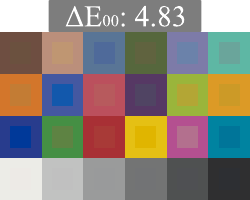} &
    \includegraphics[width=0.13\textwidth]{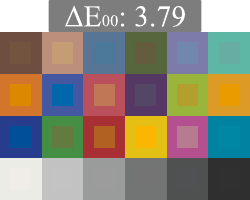}
    \\

    \includegraphics[width=0.15\textwidth]{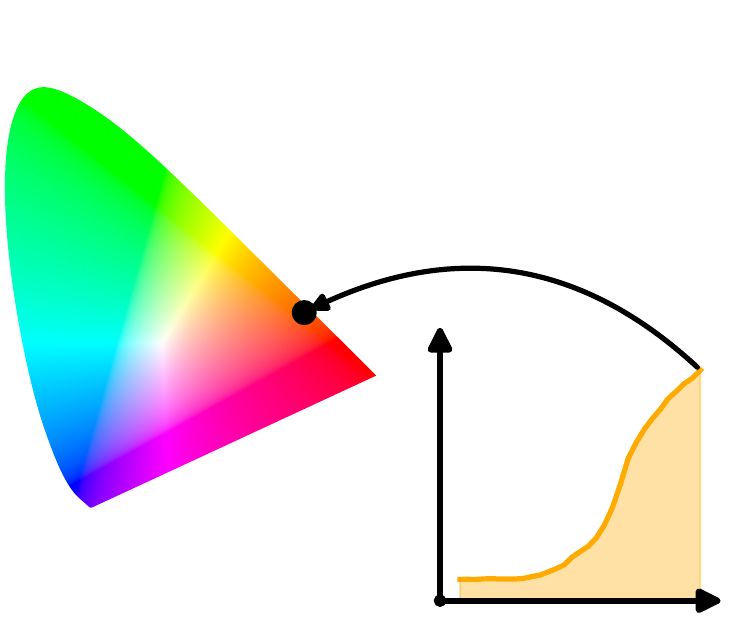} &
    \includegraphics[width=0.13\textwidth]{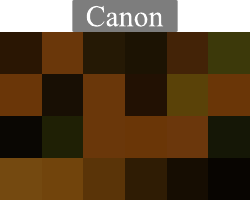} &
    \includegraphics[width=0.13\textwidth]{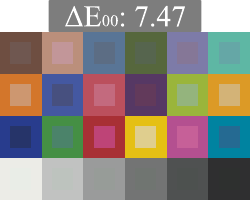} &
    \includegraphics[width=0.13\textwidth]{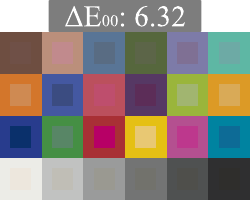} &
    \includegraphics[width=0.13\textwidth]{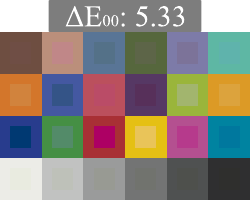} &
    \includegraphics[width=0.13\textwidth]{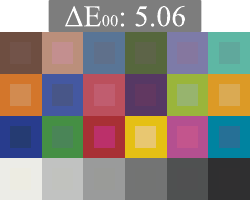} &
    \includegraphics[width=0.13\textwidth]{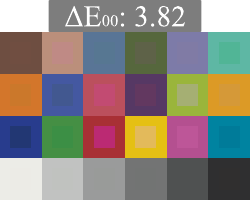}
    \\
    \includegraphics[width=0.15\textwidth]{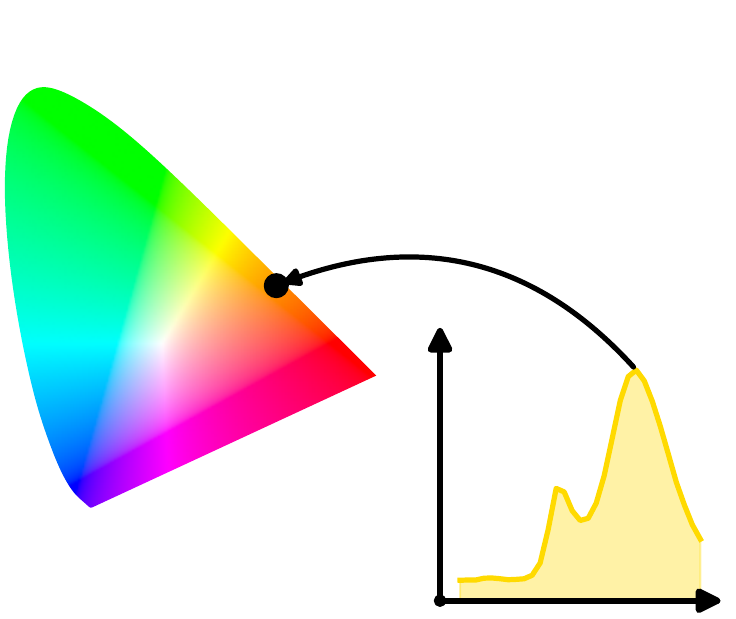} &
    \includegraphics[width=0.13\textwidth]{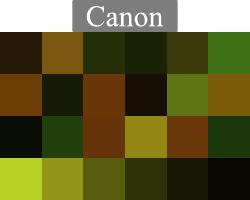} &
    \includegraphics[width=0.13\textwidth]{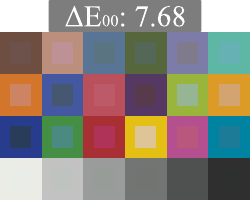} &
    \includegraphics[width=0.13\textwidth]{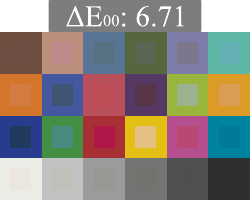} &
    \includegraphics[width=0.13\textwidth]{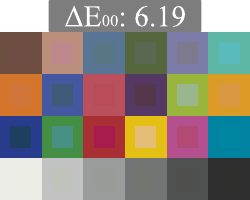} &
    \includegraphics[width=0.13\textwidth]{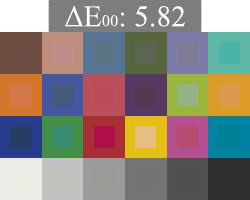} &
    \includegraphics[width=0.13\textwidth]{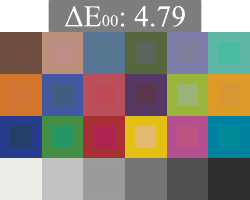}
    \\
    \includegraphics[width=0.15\textwidth]{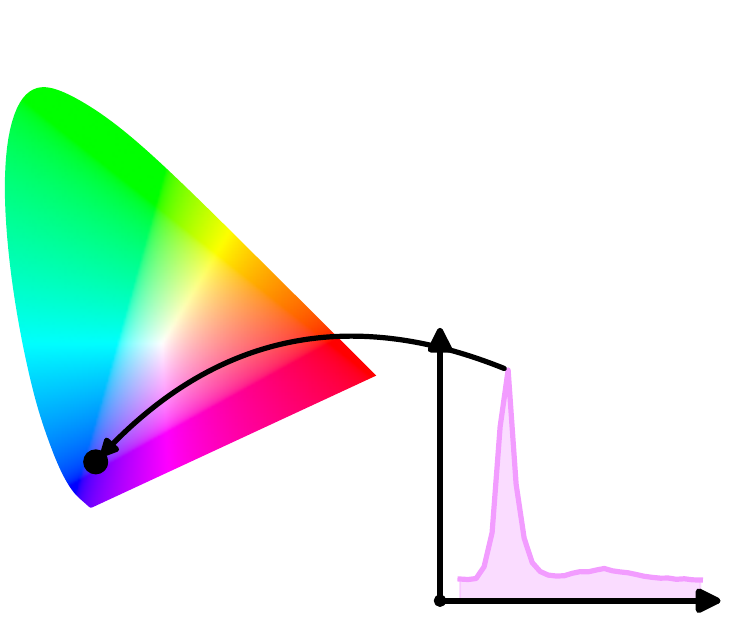} &
    \includegraphics[width=0.13\textwidth]{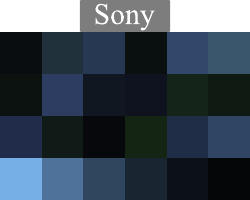} &
    \includegraphics[width=0.13\textwidth]{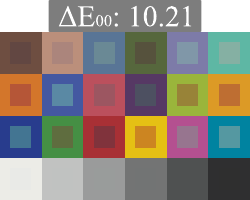} &
    \includegraphics[width=0.13\textwidth]{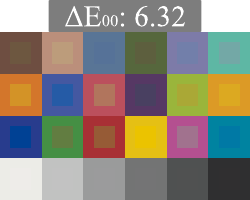} &
    \includegraphics[width=0.13\textwidth]{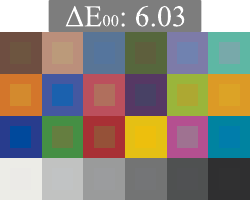} &
    \includegraphics[width=0.13\textwidth]{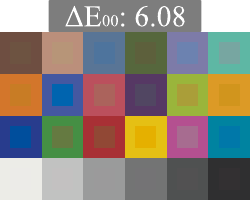} &
    \includegraphics[width=0.13\textwidth]{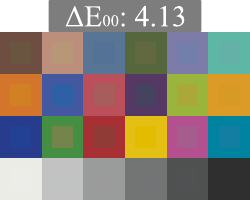}
    \\
    \includegraphics[width=0.15\textwidth]{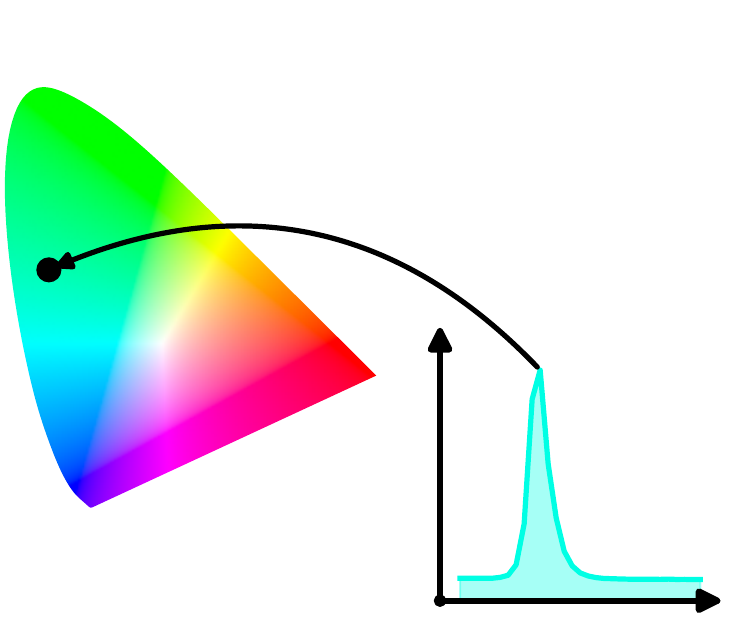} &
    \includegraphics[width=0.13\textwidth]{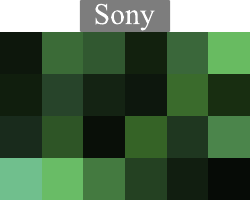} &
    \includegraphics[width=0.13\textwidth]{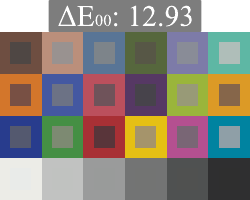} &
    \includegraphics[width=0.13\textwidth]{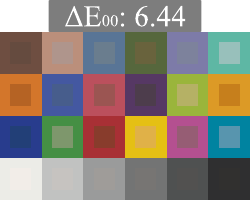} &
    \includegraphics[width=0.13\textwidth]{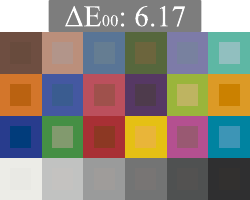} &
    \includegraphics[width=0.13\textwidth]{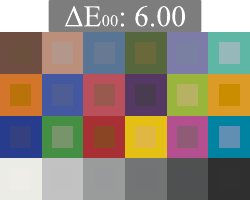} &
    \includegraphics[width=0.13\textwidth]{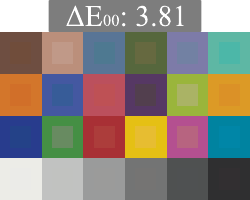}
    \\
    \includegraphics[width=0.15\textwidth]{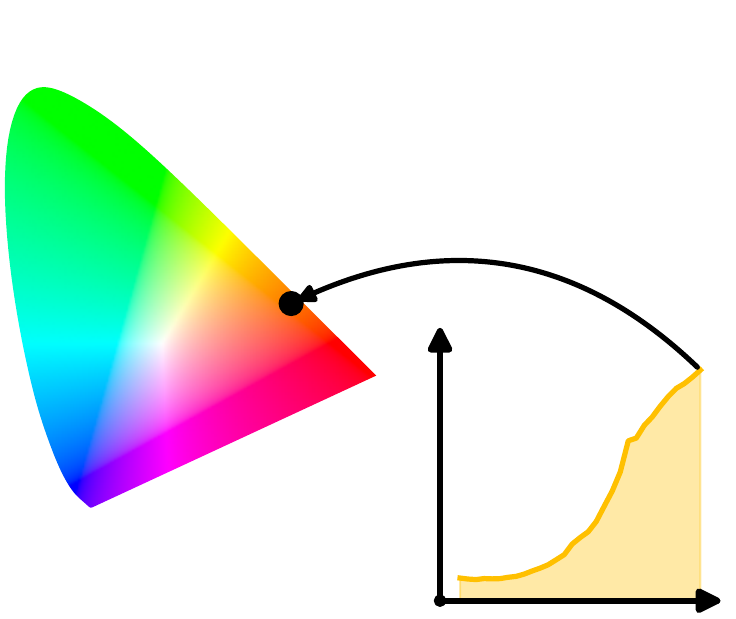} &
    \includegraphics[width=0.13\textwidth]{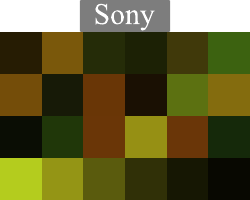} &
    \includegraphics[width=0.13\textwidth]{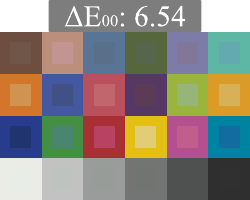} &
    \includegraphics[width=0.13\textwidth]{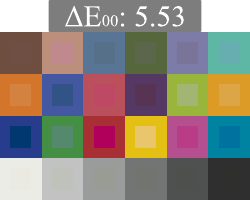} &
    \includegraphics[width=0.13\textwidth]{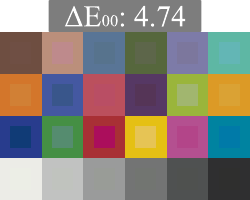} &
    \includegraphics[width=0.13\textwidth]{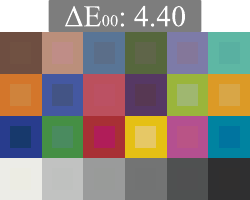} &
    \includegraphics[width=0.13\textwidth]{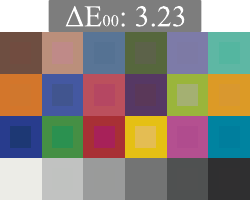}
    \\
    \includegraphics[width=0.15\textwidth]{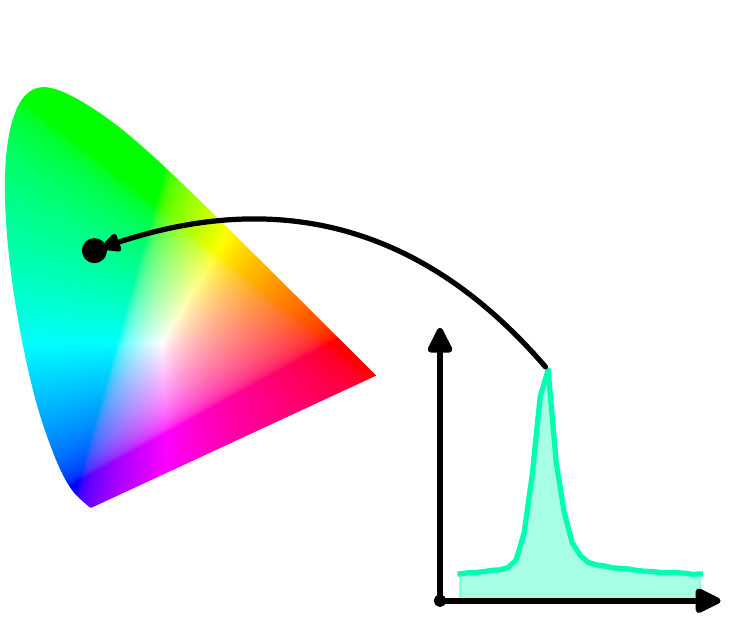} &
    \includegraphics[width=0.13\textwidth]{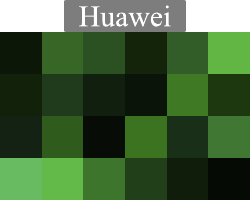} &
    \includegraphics[width=0.13\textwidth]{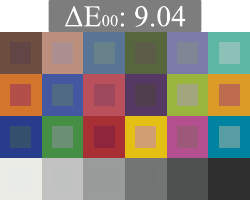} &
    \includegraphics[width=0.13\textwidth]{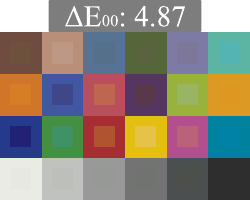} &
    \includegraphics[width=0.13\textwidth]{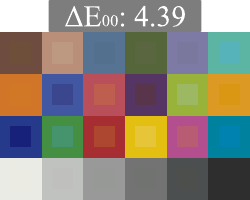} &
    \includegraphics[width=0.13\textwidth]{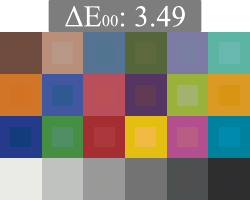} &
    \includegraphics[width=0.13\textwidth]{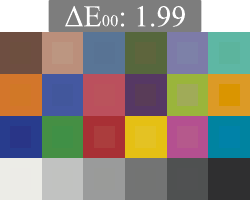}
    \\
    \includegraphics[width=0.15\textwidth]{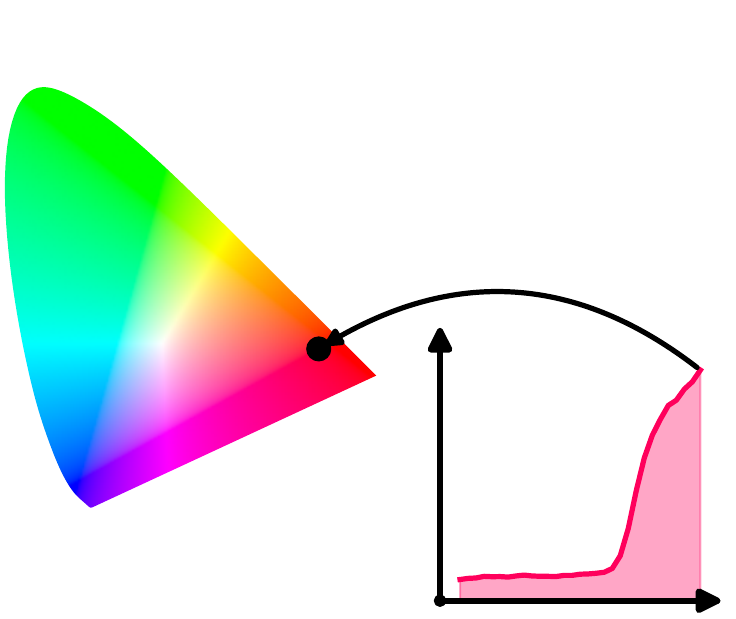} &
    \includegraphics[width=0.13\textwidth]{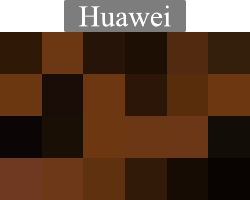} &
    \includegraphics[width=0.13\textwidth]{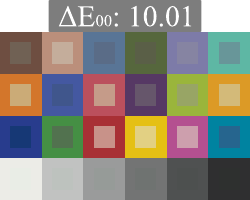} &
    \includegraphics[width=0.13\textwidth]{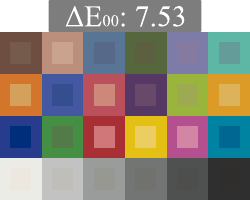} &
    \includegraphics[width=0.13\textwidth]{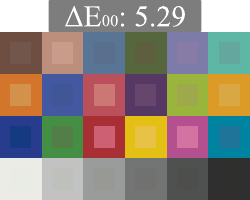} &
    \includegraphics[width=0.13\textwidth]{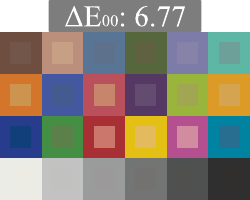} &
    \includegraphics[width=0.13\textwidth]{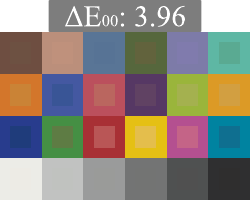}
    \\
    \includegraphics[width=0.15\textwidth]{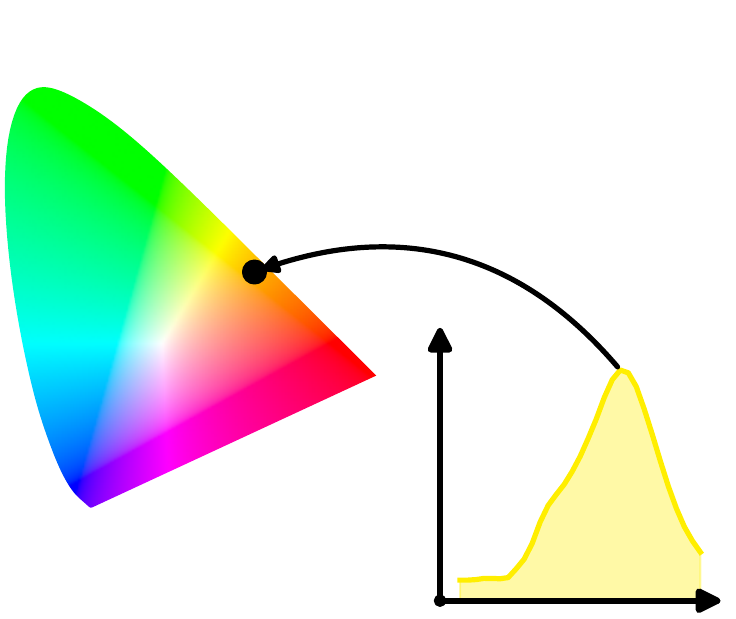} &
    \includegraphics[width=0.13\textwidth]{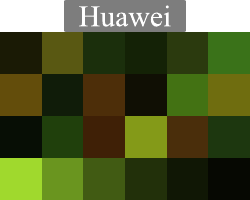} &
    \includegraphics[width=0.13\textwidth]{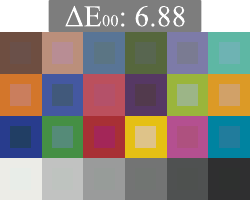} &
    \includegraphics[width=0.13\textwidth]{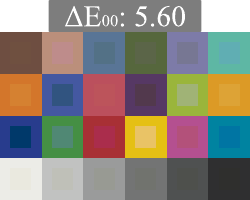} &
    \includegraphics[width=0.13\textwidth]{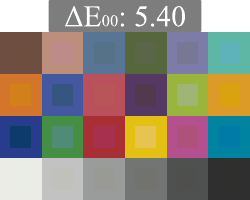} &
    \includegraphics[width=0.13\textwidth]{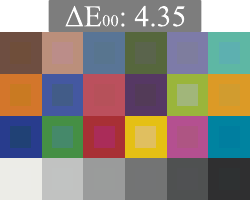} &
    \includegraphics[width=0.13\textwidth]{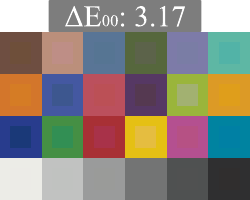}

    \end{tabular}
    \caption{Additional qualitative comparison of color correction methods on Macbeth Color Checker patches rendered under challenging test illuminants. Each corrected patch includes an inset square representing the reference color for direct comparison. For each camera, the proposed C$^2$LUT method produces colors that more closely match the target, resulting in the lowest CIE~$\Delta E_{00}$ among all compared approaches.
    \vspace{-2mm}
    }
    \label{fig:suppl_qualitative_mcc}
\end{figure*}

We show additional visualization of color correction results on KAUST~\cite{li2021multispectral} and MCC~\cite{mccamy1976color} in \Cref{fig:suppl_qualitative_kaust,fig:suppl_qualitative_mcc}, respectively. We compare our C$^2$LUT against CCM-CCT~\cite{dng}, CST-MLP~\cite{tedla2025off}, our extension of Finlayson et al.~\cite{finlayson2015color} (RP-MLP), and CCCNN~\cite{macdonald2021camera}.

Once again, the lower per-pixel $\Delta E_{00}$~\cite{sharma2005ciede2000} values in the error maps of \Cref{fig:suppl_qualitative_kaust}, the reduced visual discrepancy between the corrected patches and their inset reference colors in \Cref{fig:suppl_qualitative_mcc}, and the average $\Delta E_{00}$ scores for each image suggest that the proposed C$^2$LUT achieves more accurate visual color corrections compared to the other methods. 

\subsection{Performance across different illuminant properties}

We report the results on KAUST~\cite{li2021multispectral} and MCC~\cite{mccamy1976color} divided by illuminant properties: chromaticity distance from the canonical illuminant D65 in the CIE~xy space, and the area under their normalized SPD curve (AUC). For KAUST, they are reported in \Cref{tab:results_distancefromD65_KAUST,tab:results_AUC_KAUST}, respectively, while for MCC they are reported in \Cref{tab:results_distancefromD65_MCC,tab:results_AUC_MCC}, respectively. 

Across both datasets and both grouping criteria, the advantage of C$^2$LUT grows with the difficulty of the correction problem. For illuminants close to D65 and for broad-band illuminants, all methods perform similarly, and C$^2$LUT is competitive with RP-MLP and CCCNN. As the chromaticity distance from D65 increases, or as the illuminant spectrum narrows, C$^2$LUT consistently achieves the lowest $\Delta E_{00}$~\cite{sharma2005ciede2000} and angular error across all three cameras. This trend is particularly evident on MCC, where the gap between C$^2$LUT and the other methods widens substantially for the \quotes{mid} and \quotes{far} subsets and for \quotes{mid-band} and \quotes{narrow-band} illuminants.

\subsection{Robustness to exposure variations}

We evaluate robustness to exposure variations by scaling both the input raw responses and the reference XYZ values by a factor $\alpha \in \{0.75, 0.5, 0.25\}$, simulating progressively lower exposure levels. Results on KAUST~\cite{li2021multispectral} and MCC~\cite{mccamy1976color} are reported in \Cref{tab:exposure_KAUST,tab:exposure_MCC}, respectively.
At moderate exposure reduction ($\alpha = 0.75$ and $\alpha = 0.5$), C$^2$LUT achieves the lowest angular error across nearly all statistics and camera sensors on both datasets. At the most severe exposure reduction $\alpha = 0.25$, RP-MLP becomes competitive and obtains the lowest mean and median angular error on KAUST across all three cameras, and on MCC for the Canon and Sony sensors.

\begin{table*}[t]
    \centering
    \caption{Quantitative comparison of color correction methods on the KAUST dataset 
    , where test illuminants are divided into three subsets according to their CIE~xy-chromaticity distance from the reference D65 illuminant (Close, Mid, Far). Results are reported in terms of CIE~$\Delta E_{00}$ and angular error ($^\circ$), including mean, standard deviation, and percentile statistics (25th, 50th, 95th). The lower the better. All methods are evaluated across three camera sensors (Canon~EOS~40D, Sony~Alpha~6100, Huawei~Mate~20~Pro). Best results are highlighted in \textbf{bold}.}
    \label{tab:results_distancefromD65_KAUST}
    \begin{adjustbox}{width=\textwidth}
    \begin{NiceTabular}{ccccccccccccccccc}
         \toprule
         \Block{3-1}{Illuminant} & \Block{3-1}{Method} & \multicolumn{15}{c}{$\Delta E_{00} \downarrow$}\\\cmidrule{3-17}
         & & \multicolumn{5}{c}{Canon} & \multicolumn{5}{c}{Sony} & \multicolumn{5}{c}{Huawei}\\\cmidrule(lr){3-7}\cmidrule(lr){8-12}\cmidrule(lr){13-17}
         & &  Mean & Std & 25th & 50th & 95th & Mean & Std & 25th & 50th & 95th & Mean & Std & 25th & 50th & 95th \\\midrule

        \Block{5-1}{Close} & CCM-CCT & 3.94 & 1.78 & 2.89 & 3.58 & 7.21 & 3.75 & 1.65 & 2.91 & 3.36 & 6.87 & 4.10 & 2.00 & 3.04 & 3.64 & 7.86\\
         & CST-MLP & 3.12 & 1.84 & 1.91 & 2.80 & 6.26 & 2.97 & 1.84 & 1.99 & 2.61 & 5.82 & 3.07 & 1.96 & 1.89 & 2.66 & 6.36\\
         & RP-MLP & \textbf{2.91} & 1.82 & \textbf{1.85} & \textbf{2.43} & 6.24 & \textbf{2.95} & 1.81 & \textbf{1.93} & \textbf{2.50} & 5.64 & \textbf{3.01} & 1.89 & \textbf{1.84} & 2.66 & \textbf{6.33}\\
         & CCCNN & 3.10 & 1.58 & 2.14 & 2.62 & \textbf{5.49} & 3.00 & 1.65 & 1.96 & 2.56 & \textbf{5.42} & 3.09 & 1.84 & 2.12 & \textbf{2.58} & 6.49\\
         & C$^2$LUT (Ours) & 2.98 & \textbf{1.55} & 2.10 & 2.49 & 5.69 & 2.97 & \textbf{1.57} & 2.10 & 2.54 & 5.46 & 3.10 & \textbf{1.81} & 2.13 & 2.60 & 6.41\\
         \cmidrule(lr){1-17}

        \Block{5-1}{Mid} & CCM-CCT & 6.57 & 2.12 & 4.88 & 6.73 & 9.89 & 6.41 & 2.19 & 4.81 & 6.22 & 10.92 & 7.28 & 2.30 & 5.49 & 7.38 & 11.10\\
         & CST-MLP & 4.92 & 2.26 & 3.42 & 4.49 & 8.66 & 4.74 & 2.43 & 3.18 & 4.28 & 8.81 & 5.08 & 2.55 & 3.29 & 4.62 & \textbf{9.40}\\
         & RP-MLP & 4.67 & 2.30 & 3.05 & 3.96 & 8.63 & 4.63 & 2.47 & 3.03 & 4.01 & 8.44 & 4.86 & 2.50 & 3.15 & 4.24 & 9.56\\
         & CCCNN & 4.49 & 2.01 & 3.00 & 3.98 & \textbf{8.23} & 4.56 & 2.31 & \textbf{2.88} & 4.07 & 8.69 & 4.58 & 2.46 & 2.91 & 3.98 & 9.55\\
         & C$^2$LUT (Ours) & \textbf{4.27} & \textbf{1.97} & \textbf{2.76} & \textbf{3.70} & \textbf{8.23} & \textbf{4.32} & \textbf{2.03} & 2.89 & \textbf{3.86} & \textbf{8.20} & \textbf{4.47} & \textbf{2.28} & \textbf{2.83} & \textbf{3.75} & 9.62\\
         \cmidrule(lr){1-17}

         \Block{5-1}{Far} & CCM-CCT & 8.48 & 2.98 & 5.88 & 9.22 & 12.60 & 8.83 & 2.99 & 6.48 & 9.44 & 12.99 & 9.59 & 3.14 & 7.31 & 9.67 & 14.87\\
         & CST-MLP & 6.33 & 2.29 & 4.21 & 7.07 & 9.81 & 6.74 & 2.49 & 4.39 & 7.39 & 10.09 & 7.32 & \textbf{2.85} & 5.12 & 8.00 & 12.03\\
         & RP-MLP & 6.08 & 2.30 & 4.04 & 6.72 & 9.72 & 6.52 & 2.49 & 4.12 & 7.25 & 9.89 & 7.09 & 2.91 & 4.73 & 7.57 & 12.02\\
         & CCCNN & 6.21 & 2.22 & 4.22 & 6.74 & 9.59 & 6.72 & 2.37 & 4.47 & 7.26 & 9.99 & 7.18 & 2.89 & 4.65 & 6.84 & 11.82\\
         & C$^2$LUT (Ours) & \textbf{5.81} & \textbf{2.07} & \textbf{3.93} & \textbf{6.26} & \textbf{9.09} & \textbf{6.17} & \textbf{2.34} & \textbf{4.01} & \textbf{6.01} & \textbf{9.58} & \textbf{6.65} & \textbf{2.85} & \textbf{4.48} & \textbf{6.33} & \textbf{11.79}\\

         \midrule
         & & \multicolumn{15}{c}{Angular Error $(^\circ)\downarrow$}\\\cmidrule{3-17}
        
         \Block{5-1}{Close} & CCM-CCT & 2.79 & 1.31 & 2.11 & 2.47 & 5.35 & 2.68 & 1.27 & 2.00 & 2.36 & 4.93 & 2.92 & 1.47 & 2.11 & 2.59 & 5.96\\
         & CST-MLP & 2.27 & 1.30 & 1.48 & 1.97 & 5.05 & 2.13 & 1.29 & \textbf{1.39} & 1.91 & 4.60 & 2.06 & 1.28 & \textbf{1.22} & 1.86 & 5.06\\
         & RP-MLP & 2.12 & 1.28 & \textbf{1.40} & 1.85 & 4.75 & 2.09 & 1.31 & \textbf{1.39} & 1.84 & 4.42 & \textbf{2.03} & 1.27 & 1.28 & \textbf{1.82} & 4.95\\
         & CCCNN & 2.14 & 1.19 & 1.52 & 1.89 & 4.75 & 2.11 & 1.18 & \textbf{1.39} & 1.87 & 4.34 & 2.12 & \textbf{1.18} & 1.49 & 1.85 & 4.56\\
         & C$^2$LUT (Ours) & \textbf{2.09} & \textbf{1.17} & 1.46 & \textbf{1.81} & \textbf{4.58} & \textbf{2.06} & \textbf{1.15} & 1.43 & \textbf{1.80} & \textbf{4.10} & 2.04 & 1.19 & 1.38 & 1.83 & \textbf{4.40}\\
         \cmidrule(lr){1-17}

         \Block{5-1}{Mid} & CCM-CCT & 4.75 & 1.58 & 3.93 & 4.79 & 7.57 & 4.77 & 1.68 & 3.74 & 4.76 & 7.84 & 5.30 & 1.75 & 4.32 & 5.29 & 8.23\\
         & CST-MLP & 3.35 & 1.33 & 2.39 & 3.23 & 6.01 & 3.27 & 1.42 & 2.29 & 3.13 & 6.04 & 3.30 & 1.42 & 2.38 & 3.16 & 6.04\\
         & RP-MLP & 3.20 & 1.35 & 2.23 & 3.02 & 5.95 & 3.15 & 1.44 & 2.20 & 3.03 & 6.29 & 3.21 & 1.43 & 2.25 & 3.03 & 5.96\\
         & CCCNN & 3.03 & 1.30 & 2.10 & 2.74 & 6.14 & 3.07 & 1.40 & 2.23 & 2.85 & 5.75 & 2.99 & 1.34 & 2.13 & 2.84 & 5.82\\
         & C$^2$LUT (Ours) & \textbf{2.92} & \textbf{1.24} & \textbf{2.03} & \textbf{2.61} & \textbf{5.66} & \textbf{2.92} & \textbf{1.27} & \textbf{2.10} & \textbf{2.64} & \textbf{5.41} & \textbf{2.92} & \textbf{1.29} & \textbf{2.09} & \textbf{2.64} & \textbf{5.75}\\
         \cmidrule(lr){1-17}

         \Block{5-1}{Far} & CCM-CCT & 4.94 & 1.95 & 3.24 & 4.75 & 7.99 & 5.15 & 2.10 & 3.53 & 5.00 & 8.54 & 5.91 & 2.25 & 4.32 & 5.82 & 9.04\\
         & CST-MLP & 3.67 & 1.37 & 2.56 & 3.73 & 5.68 & 3.86 & 1.41 & 2.86 & 4.00 & 5.88 & 4.15 & 1.61 & 3.15 & 4.20 & 6.56\\
         & RP-MLP & 3.53 & 1.38 & 2.47 & 3.71 & 5.77 & 3.70 & 1.44 & \textbf{2.53} & 3.88 & 5.94 & 4.08 & 1.68 & 2.97 & 4.15 & \textbf{6.41}\\
         & CCCNN & 3.59 & 1.28 & 2.53 & 3.65 & \textbf{5.35} & 3.82 & \textbf{1.34} & 2.72 & 4.08 & \textbf{5.69} & 4.09 & \textbf{1.58} & 2.85 & 4.14 & 6.43\\
         & C$^2$LUT (Ours) & \textbf{3.39} & \textbf{1.23} & \textbf{2.35} & \textbf{3.55} & 5.61 & \textbf{3.57} & 1.35 & 2.57 & \textbf{3.82} & 5.97 & \textbf{3.85} & \textbf{1.58} & \textbf{2.78} & \textbf{3.80} & 6.48\\
         \bottomrule
         
    \end{NiceTabular}
    \end{adjustbox}
\end{table*}

\begin{table*}[t]
    \centering
    \caption{Quantitative comparison of color correction methods on the KAUST dataset
    , where test illuminants are categorized according to spectral bandwidth (broad-band, mid-band, narrow-band) based on the area under the normalized SPD. Results are reported in terms of CIE~$\Delta E_{00}$ and angular error ($^\circ$), including mean, standard deviation, and percentile statistics (25th, 50th, 95th). The lower the better. All methods are evaluated across three camera sensors (Canon~EOS~40D, Sony~Alpha~6100, Huawei~Mate~20~Pro). Best results are highlighted in \textbf{bold}.}
    \label{tab:results_AUC_KAUST}
    \begin{adjustbox}{width=\textwidth}
    \begin{NiceTabular}{ccccccccccccccccc}
         \toprule
         \Block{3-1}{Illuminant} & \Block{3-1}{Method} & \multicolumn{15}{c}{$\Delta E_{00} \downarrow$}\\\cmidrule{3-17}
         & & \multicolumn{5}{c}{Canon} & \multicolumn{5}{c}{Sony} & \multicolumn{5}{c}{Huawei}\\\cmidrule(lr){3-7}\cmidrule(lr){8-12}\cmidrule(lr){13-17}
         & &  Mean & Std & 25th & 50th & 95th & Mean & Std & 25th & 50th & 95th & Mean & Std & 25th & 50th & 95th \\\midrule

         \Block{5-1}{Broad-band} & CCM-CCT & 4.13 & 1.92 & 3.00 & 3.90 & 7.16 & 3.97 & 1.84 & 2.96 & 3.63 & 7.57 & 4.59 & 2.76 & 2.99 & 3.83 & 10.43\\
         & CST-MLP & 3.28 & 1.69 & 2.02 & 2.83 & 7.14 & 3.07 & 1.65 & 2.02 & 2.66 & 6.43 & 3.29 & 2.01 & \textbf{1.91} & 2.68 & 7.54\\
         & RP-MLP & \textbf{3.10} & 1.78 & \textbf{1.90} & \textbf{2.57} & 7.52 & \textbf{3.05} & 1.62 & \textbf{1.96} & \textbf{2.60} & \textbf{6.31} & 3.24 & 1.95 & \textbf{1.91} & \textbf{2.66} & 7.12\\
         & CCCNN & 3.22 & \textbf{1.49} & 2.17 & 2.73 & \textbf{6.42} & 3.13 & 1.58 & 2.04 & 2.72 & 6.34 & 3.26 & 1.88 & 2.16 & 2.69 & \textbf{6.87}\\
         & C$^2$LUT (Ours) & 3.16 & 1.64 & 2.13 & 2.67 & 6.68 & 3.09 & \textbf{1.55} & 2.10 & 2.65 & 6.39 & \textbf{3.23} & \textbf{1.79} & 2.13 & 2.68 & 7.29\\
         \cmidrule(lr){1-17}

\Block{5-1}{Mid-band} & CCM-CCT & 6.61 & 2.30 & 4.83 & 7.00 & 10.43 & 6.49 & 2.38 & 4.82 & 6.42 & 11.05 & 7.10 & 2.32 & 5.54 & 7.28 & 10.95\\
         & CST-MLP & 4.76 & 2.21 & 3.19 & 4.21 & 8.29 & 4.65 & 2.30 & 2.94 & 4.05 & 8.96 & 4.76 & 2.28 & 3.08 & 4.00 & 8.46\\
         & RP-MLP & 4.46 & 2.19 & 2.87 & 3.82 & 8.07 & 4.46 & 2.30 & 2.82 & 3.78 & 8.43 & 4.53 & 2.23 & 2.90 & 3.84 & 8.00\\
         & CCCNN & 4.41 & 1.94 & 2.88 & 3.96 & 7.54 & 4.45 & 2.12 & 2.86 & 3.91 & 8.26 & 4.34 & 2.10 & \textbf{2.79} & 3.76 & 8.41\\
         & C$^2$LUT (Ours) & \textbf{4.13} & \textbf{1.87} & \textbf{2.73} & \textbf{3.62} & \textbf{7.22} & \textbf{4.19} & \textbf{1.92} & \textbf{2.71} & \textbf{3.69} & \textbf{8.21} & \textbf{4.18} & \textbf{1.90} & 2.81 & \textbf{3.64} & \textbf{7.59}\\
         \cmidrule(lr){1-17}

         \Block{5-1}{Narrow-band} & CCM-CCT & 8.30 & 3.10 & 5.56 & 9.20 & 12.61 & 8.59 & 3.16 & 5.71 & 9.01 & 12.99 & 9.28 & 3.21 & 7.01 & 9.23 & 14.34\\
         & CST-MLP & 6.34 & 2.56 & 4.51 & 6.27 & 10.26 & 6.77 & 2.80 & 4.59 & 6.70 & 10.92 & 7.42 & 3.09 & 5.04 & 7.67 & 13.13\\
         & RP-MLP & 6.10 & 2.54 & 4.17 & 5.93 & 10.17 & 6.61 & 2.80 & 4.39 & 6.70 & 10.77 & 7.19 & 3.12 & 4.53 & 7.40 & 12.53\\
         & CCCNN & 6.20 & 2.46 & 4.29 & 6.08 & 10.03 & 6.73 & 2.69 & 4.38 & 6.65 & 10.69 & 7.28 & 3.16 & \textbf{4.36} & 6.80 & \textbf{12.43}\\
         & C$^2$LUT (Ours) & \textbf{5.77} & \textbf{2.22} & \textbf{4.03} & \textbf{5.65} & \textbf{9.33} & \textbf{6.21} & \textbf{2.50} & \textbf{4.10} & \textbf{5.96} & \textbf{10.49} & \textbf{6.82} & \textbf{3.05} & 4.37 & \textbf{6.33} & 12.59\\

         \midrule
         & & \multicolumn{15}{c}{Angular Error $(^\circ)\downarrow$}\\\cmidrule{3-17}
        
         \Block{5-1}{Broad-band} & CCM-CCT & 4.13 & 1.92 & 3.00 & 3.90 & 7.16 & 3.97 & 1.84 & 2.96 & 3.63 & 7.57 & 4.59 & 2.76 & 2.99 & 3.83 & 10.43\\
         & CST-MLP & 2.25 & 0.95 & 1.56 & 2.05 & 3.85 & 2.08 & 1.03 & 1.42 & 1.87 & 3.73 & 2.14 & 1.36 & \textbf{1.26} & 1.86 & \textbf{4.19}\\
         & RP-MLP & 2.11 & 0.96 & \textbf{1.43} & 1.88 & 3.84 & \textbf{2.02} & 0.98 & \textbf{1.35} & 1.82 & 3.74 & \textbf{2.09} & 1.33 & 1.28 & 1.81 & 4.23\\
         & CCCNN & 2.10 & \textbf{0.86} & 1.57 & 1.89 & \textbf{3.61} & 2.05 & \textbf{0.90} & 1.38 & 1.87 & 3.87 & 2.16 & 1.23 & 1.41 & 1.84 & 4.54\\
         & C$^2$LUT (Ours) & \textbf{2.09} & 0.89 & 1.50 & \textbf{1.87} & 3.90 & 2.04 & 0.94 & 1.41 & \textbf{1.77} & \textbf{3.72} & \textbf{2.09} & \textbf{1.19} & 1.33 & \textbf{1.75} & 4.22\\
         \cmidrule(lr){1-17}

         \Block{5-1}{Mid-band} & CCM-CCT & 4.56 & 1.61 & 3.55 & 4.66 & 7.17 & 4.66 & 1.77 & 3.18 & 4.75 & 7.89 & 5.11 & 1.78 & 3.73 & 5.17 & 8.13\\
         & CST-MLP & 3.14 & 1.19 & 2.17 & 3.05 & 5.14 & 3.12 & 1.28 & 2.18 & 2.89 & 5.46 & 3.12 & 1.32 & 2.15 & 2.91 & 5.33\\
         & RP-MLP & 2.98 & 1.15 & 2.08 & 2.82 & 4.78 & 2.98 & 1.25 & 2.07 & 2.79 & 5.07 & 3.02 & 1.32 & 2.09 & 2.82 & 5.15\\
         & CCCNN & 2.86 & 1.08 & 2.02 & 2.58 & 4.92 & 2.93 & 1.17 & 2.05 & 2.75 & 4.89 & 2.84 & 1.14 & \textbf{1.97} & 2.63 & 4.51\\
         & C$^2$LUT (Ours) & \textbf{2.72} & \textbf{0.99} & \textbf{1.97} & \textbf{2.43} & \textbf{4.39} & \textbf{2.74} & \textbf{1.04} & \textbf{2.03} & \textbf{2.46} & \textbf{4.48} & \textbf{2.71} & \textbf{1.07} & \textbf{1.97} & \textbf{2.50} & \textbf{4.42}\\
         \cmidrule(lr){1-17}

         \Block{5-1}{Narrow-band} & CCM-CCT & 5.00 & 1.99 & 3.58 & 4.86 & 8.04 & 5.23 & 2.09 & 3.76 & 5.19 & 8.57 & 5.71 & 2.16 & 4.23 & 5.70 & 8.79\\
         & CST-MLP & 3.87 & 1.62 & 2.65 & 3.76 & 6.50 & 4.06 & 1.61 & 2.94 & 4.06 & 6.84 & 4.25 & \textbf{1.62} & 3.14 & 4.27 & 6.89\\
         & RP-MLP & 3.74 & 1.66 & 2.56 & 3.67 & 6.54 & 3.95 & 1.68 & 2.69 & 4.01 & 6.61 & 4.18 & 1.70 & 3.05 & 4.21 & \textbf{6.74}\\
         & CCCNN & 3.78 & 1.54 & 2.54 & 3.76 & \textbf{6.40} & 4.02 & 1.58 & 2.81 & 4.15 & 6.92 & 4.20 & 1.64 & 2.84 & 4.30 & 6.91\\
         & C$^2$LUT (Ours) & \textbf{3.57} & \textbf{1.52} & \textbf{2.40} & \textbf{3.52} & 6.43 & \textbf{3.76} & \textbf{1.55} & \textbf{2.59} & \textbf{3.93} & \textbf{6.59} & \textbf{4.00} & 1.65 & \textbf{2.62} & \textbf{3.80} & 6.76\\
         \bottomrule
         
    \end{NiceTabular}
    \end{adjustbox}
\end{table*}
\begin{table*}[t]
    \centering
    \caption{Quantitative comparison of color correction methods on the MCC dataset 
    , where test illuminants are divided into three subsets according to their CIE~xy-chromaticity distance from the reference D65 illuminant (Close, Mid, Far). Results are reported in terms of CIE~$\Delta E_{00}$ and angular error ($^\circ$), including mean, standard deviation, and percentile statistics (25th, 50th, 95th). The lower the better. All methods are evaluated across three camera sensors (Canon~EOS~40D, Sony~Alpha~6100, Huawei~Mate~20~Pro). Best results are highlighted in \textbf{bold}.}
    \label{tab:results_distancefromD65_MCC}
    \begin{adjustbox}{width=\textwidth}
    \begin{NiceTabular}{ccccccccccccccccc}
         \toprule
         \Block{3-1}{Illuminant} & \Block{3-1}{Method} & \multicolumn{15}{c}{$\Delta E_{00} \downarrow$}\\\cmidrule{3-17}
         & & \multicolumn{5}{c}{Canon} & \multicolumn{5}{c}{Sony} & \multicolumn{5}{c}{Huawei}\\\cmidrule(lr){3-7}\cmidrule(lr){8-12}\cmidrule(lr){13-17}
         & &  Mean & Std & 25th & 50th & 95th & Mean & Std & 25th & 50th & 95th & Mean & Std & 25th & 50th & 95th \\\midrule

         \Block{5-1}{Close} & CCM-CCT & 4.55 & 2.19 & 3.10 & 4.28 & 9.40 & 4.50 & 2.02 & 3.13 & 4.14 & 9.29 & 4.89 & 2.22 & 3.45 & 4.48 & 9.73\\
         & CST-MLP & 3.57 & 2.16 & 2.23 & 3.15 & 7.85 & 3.57 & 2.16 & 2.38 & 3.15 & 7.89 & 3.64 & 2.19 & 2.29 & 3.15 & 7.88\\
         & RP-MLP & 3.13 & 2.12 & 1.79 & 2.59 & 7.37 & 3.26 & 2.12 & 1.98 & 2.70 & 7.35 & 3.42 & 2.10 & 2.08 & 2.93 & 7.44\\
         & CCCNN & 3.07 & \textbf{1.63} & 2.06 & 2.58 & 6.38 & 3.21 & \textbf{1.59} & 2.27 & 2.75 & 6.34 & 3.13 & \textbf{1.90} & 2.01 & 2.60 & 6.51\\
         & C$^2$LUT (Ours) & \textbf{2.65} & 1.70 & \textbf{1.58} & \textbf{2.09} & \textbf{6.27} & \textbf{2.68} & 1.68 & \textbf{1.68} & \textbf{2.16} & \textbf{6.27} & \textbf{2.79} & 1.93 & \textbf{1.77} & \textbf{2.18} & \textbf{6.05}\\
         \cmidrule(lr){1-17}

         \Block{5-1}{Mid} & CCM-CCT & 8.05 & 2.95 & 6.20 & 7.93 & 12.53 & 8.11 & 2.94 & 5.95 & 8.02 & 13.08 & 8.87 & 2.70 & 6.89 & 9.16 & 13.18\\
         & CST-MLP & 6.05 & 3.02 & 4.18 & 5.29 & 11.03 & 6.01 & 3.14 & 3.86 & 5.40 & 12.05 & 6.31 & 3.06 & 4.33 & 5.59 & 12.51\\
         & RP-MLP & 5.56 & 3.07 & 3.47 & 4.65 & 10.56 & 5.72 & 3.29 & 3.48 & 4.90 & 11.78 & 6.04 & 3.15 & 3.87 & 5.25 & 12.43\\
         & CCCNN & 4.86 & 2.35 & 3.30 & 4.24 & 10.11 & 5.30 & 2.70 & 3.58 & 4.48 & 11.32 & 5.29 & 2.66 & 3.39 & 4.74 & 10.53\\
         & C$^2$LUT (Ours) & \textbf{4.03} & \textbf{2.25} & \textbf{2.38} & \textbf{3.30} & \textbf{8.65} & \textbf{4.18} & \textbf{2.37} & \textbf{2.47} & \textbf{3.60} & \textbf{8.38} & \textbf{4.30} & \textbf{2.50} & \textbf{2.46} & \textbf{3.53} & \textbf{8.51}\\
         \cmidrule(lr){1-17}

         \Block{5-1}{Far} & CCM-CCT & 11.23 & 4.34 & 7.48 & 12.31 & 16.88 & 12.23 & 4.38 & 8.63 & 12.91 & 17.86 & 12.56 & 4.13 & 9.64 & 13.14 & 18.53\\
         & CST-MLP & 8.30 & 3.64 & 5.27 & 9.21 & 13.33 & 9.27 & 3.97 & 5.69 & 9.67 & 14.45 & 9.59 & 3.94 & 6.42 & 10.30 & 15.44\\
         & RP-MLP & 7.98 & 3.70 & 4.69 & 9.14 & 13.01 & 8.94 & 4.11 & 4.95 & 9.46 & 14.60 & 9.26 & 4.06 & 5.23 & 10.22 & 14.96\\
         & CCCNN & 7.98 & 3.47 & 4.63 & 8.56 & 12.56 & 9.38 & 3.71 & 5.66 & 9.71 & 14.56 & 9.38 & 3.81 & 6.10 & 9.94 & 15.17\\
         & C$^2$LUT (Ours) & \textbf{6.68} & \textbf{3.17} & \textbf{3.47} & \textbf{6.80} & \textbf{11.35} & \textbf{7.39} & \textbf{3.68} & \textbf{3.92} & \textbf{6.86} & \textbf{12.81} & \textbf{7.65} & \textbf{3.87} & \textbf{4.14} & \textbf{6.92} & \textbf{13.37}\\

         \midrule
         & & \multicolumn{15}{c}{Angular Error $(^\circ)\downarrow$}\\\cmidrule{3-17}
        
         \Block{5-1}{Close} & CCM-CCT & 3.16 & 1.62 & 2.06 & 2.87 & 5.70 & 3.19 & 1.62 & 2.04 & 2.70 & 6.04 & 3.49 & 1.78 & 2.36 & 3.10 & 6.38\\
         & CST-MLP & 2.34 & 1.53 & 1.33 & 1.93 & 5.52 & 2.39 & 1.54 & 1.44 & 2.08 & 5.41 & 2.39 & 1.55 & 1.46 & 2.05 & 5.51\\
         & RP-MLP & 2.10 & 1.47 & 1.13 & 1.68 & 5.11 & 2.24 & 1.48 & 1.27 & 1.81 & 5.09 & 2.29 & 1.50 & 1.42 & 1.91 & 5.38\\
         & CCCNN & 2.08 & 1.22 & 1.38 & 1.70 & 4.25 & 2.14 & \textbf{1.22} & 1.54 & 1.75 & 4.67 & 2.06 & \textbf{1.32} & 1.31 & 1.73 & 4.69\\
         & C$^2$LUT (Ours) & \textbf{1.78} & \textbf{1.17} & \textbf{1.06} & \textbf{1.48} & \textbf{4.10} & \textbf{1.88} & 1.23 & \textbf{1.19} & \textbf{1.59} & \textbf{4.39} & \textbf{1.84} & \textbf{1.32} & \textbf{1.12} & \textbf{1.50} & \textbf{4.07}\\
         \cmidrule(lr){1-17}

         \Block{5-1}{Mid} & CCM-CCT & 5.95 & 2.35 & 4.19 & 6.30 & 9.53 & 6.17 & 2.53 & 3.76 & 6.54 & 10.08 & 6.75 & 2.31 & 5.23 & 7.01 & 10.15\\
         & CST-MLP & 3.90 & 1.87 & 2.32 & 3.72 & 7.13 & 4.07 & 2.01 & 2.48 & 3.62 & 7.90 & 4.15 & 1.98 & 2.57 & 3.85 & 8.05\\
         & RP-MLP & 3.61 & 1.81 & 2.09 & 3.52 & 6.61 & 3.80 & 1.94 & 2.25 & 3.45 & 7.35 & 4.02 & 1.97 & 2.51 & 3.60 & 7.75\\
         & CCCNN & 3.20 & 1.52 & 2.00 & 2.84 & 6.26 & 3.52 & 1.72 & 2.16 & 3.06 & 6.68 & 3.45 & 1.67 & 2.09 & 3.28 & 6.51\\
         & C$^2$LUT (Ours) & \textbf{2.63} & \textbf{1.30} & \textbf{1.60} & \textbf{2.30} & \textbf{4.97} & \textbf{2.84} & \textbf{1.49} & \textbf{1.73} & \textbf{2.47} & \textbf{5.78} & \textbf{2.87} & \textbf{1.60} & \textbf{1.70} & \textbf{2.48} & \textbf{5.86}\\
         \cmidrule(lr){1-17}

         \Block{5-1}{Far} & CCM-CCT & 6.02 & 2.69 & 3.97 & 5.85 & 11.01 & 6.75 & 2.87 & 4.53 & 6.76 & 11.62 & 7.25 & 2.78 & 5.71 & 6.98 & 11.92\\
         & CST-MLP & 4.28 & 1.81 & 2.64 & 4.67 & 6.84 & 4.89 & 1.96 & 3.19 & 5.08 & 7.37 & 5.13 & 2.13 & 3.86 & 5.34 & 8.12\\
         & RP-MLP & 4.08 & 1.83 & 2.42 & 4.46 & 6.63 & 4.67 & 1.98 & 2.73 & 4.99 & 7.10 & 4.94 & 2.19 & 3.05 & 5.24 & 8.64\\
         & CCCNN & 4.14 & 1.68 & 2.52 & 4.35 & 6.58 & 4.70 & 1.75 & 3.08 & 4.92 & 7.24 & 4.83 & 1.87 & 3.37 & 5.08 & 7.33\\
         & C$^2$LUT (Ours) & \textbf{3.43} & \textbf{1.47} & \textbf{2.05} & \textbf{3.57} & \textbf{5.61} & \textbf{3.79} & \textbf{1.61} & \textbf{2.55} & \textbf{3.60} & \textbf{6.06} & \textbf{3.99} & \textbf{1.86} & \textbf{2.55} & \textbf{3.93} & \textbf{7.05}\\
         \bottomrule
         
    \end{NiceTabular}
    \end{adjustbox}
\end{table*}

\begin{table*}[t]
    \centering
    \caption{Quantitative comparison of color correction methods on the MCC dataset
    , where test illuminants are categorized according to spectral bandwidth (broad-band, mid-band, narrow-band) based on the area under the normalized SPD. Results are reported in terms of CIE~$\Delta E_{00}$ and angular error ($^\circ$), including mean, standard deviation, and percentile statistics (25th, 50th, 95th). The lower the better. All methods are evaluated across three camera sensors (Canon~EOS~40D, Sony~Alpha~6100, Huawei~Mate~20~Pro). Best results are highlighted in \textbf{bold}.}
    \label{tab:results_AUC_MCC}
    \begin{adjustbox}{width=\textwidth}
    \begin{NiceTabular}{ccccccccccccccccc}
         \toprule
         \Block{3-1}{Illuminant} & \Block{3-1}{Method} & \multicolumn{15}{c}{$\Delta E_{00} \downarrow$}\\\cmidrule{3-17}
         & & \multicolumn{5}{c}{Canon} & \multicolumn{5}{c}{Sony} & \multicolumn{5}{c}{Huawei}\\\cmidrule(lr){3-7}\cmidrule(lr){8-12}\cmidrule(lr){13-17}
         & &  Mean & Std & 25th & 50th & 95th & Mean & Std & 25th & 50th & 95th & Mean & Std & 25th & 50th & 95th \\\midrule

         \Block{5-1}{Broad-band} & CCM-CCT & 5.03 & 2.58 & 3.57 & 4.53 & 10.08 & 5.12 & 2.68 & 3.50 & 4.40 & 11.11 & 5.81 & 3.50 & 3.84 & 4.81 & 12.85\\
         & CST-MLP & 4.10 & 2.32 & 2.44 & 3.48 & 9.47 & 4.07 & 2.36 & 2.52 & 3.38 & 9.66 & 4.36 & 2.79 & 2.66 & 3.39 & \textbf{9.50}\\
         & RP-MLP & 3.65 & 2.40 & 2.04 & 2.92 & 9.55 & 3.74 & 2.33 & 2.12 & 3.00 & 9.65 & 4.09 & 2.70 & 2.32 & 3.35 & 9.92\\
         & CCCNN & 3.54 & \textbf{2.04} & 2.17 & 2.87 & 9.05 & 3.69 & 2.20 & 2.34 & 2.97 & 9.00 & 3.67 & 2.59 & 2.04 & 2.89 & 10.14\\
         & C$^2$LUT (Ours) & \textbf{3.03} & 2.05 & \textbf{1.76} & \textbf{2.39} & \textbf{7.83} & \textbf{3.04} & \textbf{2.07} & \textbf{1.82} & \textbf{2.37} & \textbf{7.80} & \textbf{3.22} & \textbf{2.33} & \textbf{1.83} & \textbf{2.39} & 10.14\\
         \cmidrule(lr){1-17}

\Block{5-1}{Mid-band} & CCM-CCT & 8.35 & 3.33 & 6.25 & 8.80 & 13.78 & 8.48 & 3.46 & 6.49 & 8.31 & 14.80 & 9.02 & 3.21 & 7.03 & 9.34 & 14.65\\
         & CST-MLP & 6.10 & 3.29 & 3.71 & 5.28 & 11.83 & 6.15 & 3.42 & 3.72 & 5.22 & 13.31 & 6.22 & 3.30 & 3.69 & 5.31 & 12.06\\
         & RP-MLP & 5.61 & 3.30 & 3.25 & 4.75 & 11.63 & 5.82 & 3.53 & 3.22 & 4.81 & 13.26 & 5.97 & 3.40 & 3.39 & 4.85 & 11.91\\
         & CCCNN & 5.16 & 2.86 & 3.04 & 4.24 & 10.63 & 5.63 & 3.11 & 3.37 & 4.69 & 12.25 & 5.42 & 2.89 & 3.13 & 4.80 & 11.37\\
         & C$^2$LUT (Ours) & \textbf{4.18} & \textbf{2.62} & \textbf{2.16} & \textbf{3.19} & \textbf{8.95} & \textbf{4.30} & \textbf{2.68} & \textbf{2.26} & \textbf{3.35} & \textbf{10.36} & \textbf{4.25} & \textbf{2.65} & \textbf{2.24} & \textbf{3.38} & \textbf{10.00}\\
         \cmidrule(lr){1-17}

         \Block{5-1}{Narrow-band} & CCM-CCT & 10.57 & 4.70 & 5.96 & 11.17 & 16.88 & 11.37 & 4.91 & 6.59 & 11.67 & 17.86 & 11.55 & 4.44 & 8.37 & 12.09 & 18.32\\
         & CST-MLP & 7.80 & 3.93 & 4.54 & 7.17 & 13.60 & 8.72 & 4.37 & 5.08 & 8.03 & 15.10 & 9.03 & 4.26 & 5.55 & 9.08 & 15.56\\
         & RP-MLP & 7.48 & 3.95 & 3.83 & 7.22 & 13.29 & 8.45 & 4.49 & 4.26 & 7.82 & 14.94 & 8.71 & 4.31 & 5.00 & 8.68 & 15.61\\
         & CCCNN & 7.28 & 3.69 & 4.00 & 6.35 & 12.56 & 8.67 & 4.10 & 5.33 & 8.43 & 14.56 & 8.78 & 4.18 & 5.24 & 8.67 & 15.17\\
         & C$^2$LUT (Ours) & \textbf{3.24} & \textbf{1.49} & \textbf{2.05} & \textbf{3.02} & \textbf{5.89} & \textbf{3.67} & \textbf{1.71} & \textbf{2.39} & \textbf{3.38} & \textbf{6.39} & \textbf{3.87} & \textbf{1.89} & \textbf{2.29} & \textbf{3.57} & \textbf{7.52}\\

         \midrule
         & & \multicolumn{15}{c}{Angular Error $(^\circ)\downarrow$}\\\cmidrule{3-17}
        
         \Block{5-1}{Broad-band} & CCM-CCT & 3.49 & 1.82 & 2.07 & 3.08 & 6.90 & 3.48 & 1.76 & 2.20 & 3.16 & 6.56 & 4.19 & 2.50 & 2.46 & 3.42 & 9.26\\
         & CST-MLP & 2.63 & 1.52 & 1.63 & 2.29 & 5.44 & 2.69 & 1.56 & 1.69 & 2.30 & 5.55 & 2.86 & 1.99 & 1.66 & 2.32 & 6.39\\
         & RP-MLP & 2.37 & 1.50 & 1.33 & 1.96 & 5.32 & 2.49 & 1.47 & 1.50 & 2.11 & 5.36 & 2.71 & 1.94 & 1.50 & 2.16 & 5.90\\
         & CCCNN & 2.30 & 1.32 & 1.48 & 1.92 & 4.55 & 2.36 & 1.36 & 1.60 & 1.83 & 4.98 & 2.39 & 1.70 & 1.46 & 1.90 & 5.62\\
         & C$^2$LUT (Ours) & \textbf{1.97} & \textbf{1.23} & \textbf{1.19} & \textbf{1.60} & \textbf{4.25} & \textbf{2.06} & \textbf{1.27} & \textbf{1.30} & \textbf{1.68} & \textbf{4.65} & \textbf{2.13} & \textbf{1.58} & \textbf{1.21} & \textbf{1.65} & \textbf{5.21}\\
         \cmidrule(lr){1-17}

         \Block{5-1}{Mid-band} & CCM-CCT & 6.00 & 2.63 & 3.75 & 6.47 & 10.03 & 6.28 & 2.78 & 3.73 & 6.68 & 10.61 & 6.69 & 2.66 & 4.50 & 7.04 & 10.81\\
         & CST-MLP & 3.89 & 2.01 & 2.11 & 3.83 & 7.21 & 4.06 & 2.10 & 2.27 & 3.68 & 8.13 & 4.04 & 2.12 & 2.18 & 3.82 & 7.83\\
         & RP-MLP & 3.59 & 1.95 & 1.91 & 3.47 & 6.71 & 3.79 & 2.04 & 2.04 & 3.45 & 7.64 & 3.91 & 2.10 & 1.99 & 3.64 & 7.58\\
         & CCCNN & 3.31 & 1.72 & 1.85 & 2.92 & 6.57 & 3.63 & 1.82 & 2.16 & 3.22 & 6.78 & 3.44 & 1.74 & 1.97 & 3.28 & 6.45\\
         & C$^2$LUT (Ours) & \textbf{2.65} & \textbf{1.44} & \textbf{1.51} & \textbf{2.12} & \textbf{5.32} & \textbf{2.82} & \textbf{1.55} & \textbf{1.62} & \textbf{2.42} & \textbf{5.78} & \textbf{2.73} & \textbf{1.59} & \textbf{1.50} & \textbf{2.25} & \textbf{5.46}\\
         \cmidrule(lr){1-17}

         \Block{5-1}{Narrow-band} & CCM-CCT & 5.68 & 2.61 & 3.61 & 5.65 & 10.77 & 6.41 & 2.89 & 3.94 & 6.55 & 11.72 & 6.64 & 2.68 & 4.74 & 6.36 & 11.66\\
         & CST-MLP & 4.04 & 1.92 & 2.37 & 3.88 & 6.87 & 4.66 & 2.16 & 2.96 & 4.39 & 7.61 & 4.81 & 2.10 & 2.76 & 4.91 & 8.19\\
         & RP-MLP & 3.86 & 1.91 & 2.12 & 3.64 & 6.76 & 4.48 & 2.17 & 2.59 & 4.12 & 7.30 & 4.65 & 2.12 & 2.54 & 4.85 & 8.39\\
         & CCCNN & 3.84 & 1.70 & 2.36 & 3.38 & 6.38 & 4.42 & 1.91 & 2.77 & 4.16 & 7.29 & 4.55 & 1.95 & 2.86 & 4.66 & 7.53\\
         & C$^2$LUT (Ours) & \textbf{3.24} & \textbf{1.49} & \textbf{2.05} & \textbf{3.02} & \textbf{5.89} & \textbf{3.67} & \textbf{1.71} & \textbf{2.39} & \textbf{3.38} & \textbf{6.39} & \textbf{3.87} & \textbf{1.89} & \textbf{2.29} & \textbf{3.57} & \textbf{7.52}\\
         \bottomrule
         
    \end{NiceTabular}
    \end{adjustbox}
\end{table*}
\begin{table*}[t]
    \centering
    \caption{Quantitative evaluation of color correction robustness to exposure variations on the KAUST dataset. Input raw responses and corresponding reference XYZ values are scaled by a factor $\alpha \in {0.75, 0.5, 0.25}$ to simulate progressively decreasing exposure levels. All methods are evaluated across three camera sensors (Canon~EOS~40D, Sony~Alpha~6100, Huawei~Mate~20~Pro). Results are reported using angular error ($^\circ$), including mean, standard deviation, and percentile statistics (25th, 50th, 95th). The lower the better. Best results are highlighted in \textbf{bold}.}
    \label{tab:exposure_KAUST}
    \begin{adjustbox}{width=\textwidth}
    \begin{NiceTabular}{ccccccccccccccccc}
         \toprule
         \Block{3-1}{Exposure} & \Block{3-1}{Method} & \multicolumn{15}{c}{AE $\downarrow$}\\\cmidrule{3-17}
         & & \multicolumn{5}{c}{Canon} & \multicolumn{5}{c}{Sony} & \multicolumn{5}{c}{Huawei}\\\cmidrule(lr){3-7}\cmidrule(lr){8-12}\cmidrule(lr){13-17}
         & &  Mean & Std & 25th & 50th & 95th & Mean & Std & 25th & 50th & 95th & Mean & Std & 25th & 50th & 95th \\\midrule
        
         \Block{5-1}{0.75} & CCM-CCT & 4.16 & 1.89 & 2.45 & 3.92 & 7.73 & 4.21 & 2.03 & 2.39 & 3.99 & 8.03 & 4.74 & 2.30 & 2.69 & 4.46 & 8.65\\
         & CST-MLP & 3.10 & 1.45 & 1.97 & 2.83 & 5.59 & 3.10 & 1.55 & 1.90 & 2.89 & 5.71 & 3.18 & 1.67 & 1.89 & 2.91 & 6.01\\
         & RP-MLP & 2.94 & 1.47 & 1.81 & 2.66 & 5.55 & 2.98 & 1.56 & 1.76 & 2.72 & 5.35 & 3.11 & 1.71 & 1.81 & 2.82 & 6.21\\
         & CCCNN & 2.98 & 1.45 & 1.89 & 2.56 & 5.49 & 3.07 & 1.54 & 1.88 & 2.72 & 5.68 & 3.15 & 1.62 & 1.92 & 2.71 & 6.28\\
         & C$^2$LUT (Ours) & \textbf{2.80} & \textbf{1.36} & \textbf{1.80} & \textbf{2.40} & \textbf{5.44} & \textbf{2.84} & \textbf{1.44} & \textbf{1.74} & \textbf{2.50} & \textbf{5.22} & \textbf{2.93} & \textbf{1.58} & \textbf{1.77} & \textbf{2.53} & \textbf{5.74}\\
\cmidrule(lr){1-17}

\Block{5-1}{0.5} & CCM-CCT & 4.16 & 1.89 & 2.45 & 3.92 & 7.73 & 4.21 & 2.03 & 2.39 & 3.99 & 8.03 & 4.74 & 2.30 & 2.69 & 4.46 & 8.65\\
         & CST-MLP & 3.10 & 1.45 & 1.97 & 2.83 & 5.59 & 3.10 & 1.55 & 1.90 & 2.89 & 5.71 & 3.18 & 1.67 & 1.89 & 2.91 & 6.01\\
         & RP-MLP & 2.95 & 1.48 & \textbf{1.82} & 2.68 & \textbf{5.56} & 2.99 & 1.58 & \textbf{1.75} & 2.74 & 5.39 & 3.12 & 1.72 & \textbf{1.80} & 2.87 & 6.29\\
         & CCCNN & 3.16 & 1.56 & 1.95 & 2.74 & 5.86 & 3.24 & 1.64 & 1.96 & 2.88 & 6.25 & 3.36 & 1.66 & 2.12 & 2.89 & 6.45\\
        & C$^2$LUT (Ours) & \textbf{2.92} & \textbf{1.40} & 1.87 & \textbf{2.53} & 5.60 & \textbf{2.97} & \textbf{1.47} & 1.80 & \textbf{2.63} & \textbf{5.37} & \textbf{3.07} & \textbf{1.60} & 1.84 & \textbf{2.67} & \textbf{5.95}\\

        \cmidrule(lr){1-17}

        \Block{5-1}{0.25} &CCM-CCT & 4.16 & 1.89 & 2.45 & 3.92 & 7.73 & 4.21 & 2.03 & 2.39 & 3.99 & 8.03 & 4.74 & 2.30 & 2.69 & 4.46 & 8.65\\
         & CST-MLP & 3.10 & \textbf{1.45} & 1.97 & 2.83 & 5.59 & 3.10 & \textbf{1.55} & 1.90 & 2.89 & 5.71 & 3.18 & 1.67 & 1.89 & 2.91 & \textbf{6.01}\\
         & RP-MLP & \textbf{2.96} & 1.48 & \textbf{1.82} & \textbf{2.69} & \textbf{5.57} & \textbf{3.00} & 1.58 & \textbf{1.76} & \textbf{2.74} & \textbf{5.42} & \textbf{3.13} & 1.73 & \textbf{1.80} & \textbf{2.88} & 6.32\\
         & CCCNN & 3.77 & 1.87 & 2.28 & 3.35 & 7.25 & 3.87 & 1.97 & 2.40 & 3.36 & 7.49 & 4.08 & 1.79 & 2.71 & 3.66 & 7.21\\
         & C$^2$LUT (Ours) & 3.39 & 1.49 & 2.19 & 3.08 & 5.88 & 3.46 & \textbf{1.55} & 2.15 & 3.22 & 6.07 & 3.65 & \textbf{1.62} & 2.33 & 3.28 & 6.48\\
         
         \bottomrule
         
    \end{NiceTabular}
    \end{adjustbox}
\end{table*}
\begin{table*}[t]
    \centering
    \caption{Quantitative evaluation of color correction robustness to exposure variations on the MCC dataset. Input raw responses and corresponding reference XYZ values are scaled by a factor $\alpha \in {0.75, 0.5, 0.25}$ to simulate progressively decreasing exposure levels. All methods are evaluated across three camera sensors (Canon~EOS~40D, Sony~Alpha~6100, Huawei~Mate~20~Pro). Results are reported using angular error ($^\circ$), including mean, standard deviation, and percentile statistics (25th, 50th, 95th). The lower the better. Best results are highlighted in \textbf{bold}.}
    \label{tab:exposure_MCC}
    \begin{adjustbox}{width=\textwidth}
    \begin{NiceTabular}{ccccccccccccccccc}
         \toprule
         \Block{3-1}{Exposure} & \Block{3-1}{Method} & \multicolumn{15}{c}{AE $\downarrow$}\\\cmidrule{3-17}
         & & \multicolumn{5}{c}{Canon} & \multicolumn{5}{c}{Sony} & \multicolumn{5}{c}{Huawei}\\\cmidrule(lr){3-7}\cmidrule(lr){8-12}\cmidrule(lr){13-17}
         & &  Mean & Std & 25th & 50th & 95th & Mean & Std & 25th & 50th & 95th & Mean & Std & 25th & 50th & 95th \\\midrule

\Block{5-1}{0.75} & CCM-CCT & 5.05 & 2.62 & 2.78 & 4.94 & 9.85 & 5.38 & 2.86 & 2.87 & 5.21 & 10.67 & 5.83 & 2.85 & 3.17 & 5.94 & 10.72\\
         & CST-MLP & 3.51 & 1.93 & 1.95 & 3.13 & 6.87 & 3.80 & 2.12 & 2.12 & 3.28 & 7.47 & 3.90 & 2.21 & 2.07 & 3.45 & 7.85\\
         & RP-MLP & 3.30 & 1.91 & 1.69 & 2.86 & 6.58 & 3.61 & 2.09 & 1.90 & 3.13 & 7.23 & 3.77 & 2.21 & 2.00 & 3.36 & 7.64\\
         & CCCNN & 3.23 & 1.72 & 1.81 & 2.69 & 6.34 & 3.52 & 1.92 & 1.91 & 2.98 & 7.13 & 3.52 & 1.98 & 1.91 & 2.91 & 7.18\\
         & C$^2$LUT (Ours) & \textbf{2.81} & \textbf{1.50} & \textbf{1.62} & \textbf{2.40} & \textbf{5.61} & \textbf{2.99} & \textbf{1.63} & \textbf{1.74} & \textbf{2.57} & \textbf{6.09} & \textbf{3.04} & \textbf{1.83} & \textbf{1.68} & \textbf{2.56} & \textbf{6.65}\\
\cmidrule(lr){1-17}

\Block{5-1}{0.5} & CCM-CCT & 5.05 & 2.62 & 2.78 & 4.94 & 9.85 & 5.38 & 2.86 & 2.87 & 5.21 & 10.67 & 5.83 & 2.85 & 3.17 & 5.94 & 10.72\\
         & CST-MLP & 3.51 & 1.93 & 1.95 & 3.13 & 6.87 & 3.80 & 2.12 & 2.12 & 3.28 & 7.47 & 3.90 & 2.21 & 2.07 & 3.45 & 7.85\\
         & RP-MLP & 3.33 & 1.91 & 1.72 & 2.87 & 6.60 & 3.64 & 2.10 & 1.91 & 3.13 & 7.26 & 3.80 & 2.23 & 2.02 & 3.36 & 7.68\\
         & CCCNN & 3.42 & 1.78 & 2.01 & 2.93 & 6.65 & 3.69 & 1.95 & 2.20 & 3.20 & 7.22 & 3.70 & 1.96 & 2.19 & 3.22 & 7.28\\
         & C$^2$LUT (Ours) & \textbf{3.05} & \textbf{1.57} & \textbf{1.80} & \textbf{2.64} & \textbf{5.94} & \textbf{3.25} & \textbf{1.71} & \textbf{1.87} & \textbf{2.76} & \textbf{6.53} & \textbf{3.24} & \textbf{1.90} & \textbf{1.72} & \textbf{2.71} & \textbf{6.77}\\
\cmidrule(lr){1-17}

\Block{5-1}{0.25} & CCM-CCT & 5.05 & 2.62 & 2.78 & 4.94 & 9.85 & 5.38 & 2.86 & 2.87 & 5.21 & 10.67 & 5.83 & 2.85 & 3.17 & 5.94 & 10.72\\
         & CST-MLP & 3.51 & 1.93 & 1.95 & 3.13 & 6.87 & 3.80 & 2.12 & 2.12 & 3.28 & 7.47 & 3.90 & 2.21 & 2.07 & 3.45 & 7.85\\
         & RP-MLP & \textbf{3.35} & 1.91 & \textbf{1.75} & \textbf{2.90} & 6.61 & 3.67 & 2.11 & \textbf{1.92} & \textbf{3.17} & 7.30 & 3.82 & 2.24 & \textbf{2.03} & 3.37 & 7.71\\
         & CCCNN & 3.86 & 1.94 & 2.35 & 3.40 & 7.26 & 4.17 & 2.07 & 2.56 & 3.69 & 7.78 & 4.11 & 2.06 & 2.55 & 3.55 & 7.97\\
         & C$^2$LUT (Ours) & 3.46 & \textbf{1.61} & 2.14 & 3.11 & \textbf{6.20} & \textbf{3.65} & \textbf{1.75} & 2.20 & 3.19 & \textbf{6.66} & \textbf{3.65} & \textbf{1.96} & 2.16 & \textbf{3.12} & \textbf{7.34}\\
         
         \bottomrule
         
    \end{NiceTabular}
    \end{adjustbox}
\end{table*}

\end{document}